\PassOptionsToPackage{table}{xcolor}
\documentclass[10pt]{article}
\ifdefined\QUATKANANON
  \usepackage{tmlr}
\else
  \usepackage[preprint]{tmlr}
\fi
\usepackage{amsmath,amssymb,amsthm}
\usepackage{booktabs}
\usepackage{colortbl}

\usepackage{array}
\usepackage{longtable}
\usepackage{caption}
\newcommand{\best}[1]{{\mathversion{bold}$#1$}}
\rowcolors{2}{black!4}{}
\usepackage{graphicx}
\usepackage{tikz}
\usetikzlibrary{arrows.meta, positioning, fit, calc}
\usepackage{sansmath}
\newcommand{\figfont}{\sffamily\sansmath\fontsize{6.5}{7.6}\selectfont}
\usepackage{multirow}
\usepackage[hidelinks,bookmarksnumbered,bookmarksopen,bookmarksopenlevel=1]{hyperref}
\usepackage{url}
\usepackage{microtype}

\ifdefined\usexr
  \IfFileExists{xr_hook.tex}{\input{xr_hook.tex}}{%
    \errmessage{Standalone build: xr_hook.tex not found. Build quatkan_tmlr.tex instead.}}
\fi

\usepackage{float}
\usepackage[section]{placeins}

\newtheoremstyle{quatkantheorem}%
  {0.8em}
  {0.5em}
  {\itshape}
  {}
  {\bfseries}
  {.}
  { }
  {\thmname{#1}\thmnumber{ #2}\thmnote{\ (#3)}}
\theoremstyle{quatkantheorem}
\newtheorem{lemma}{Lemma}

\newtheorem{corollary}{Corollary}
\newtheorem{criterion}{Criterion}
\theoremstyle{definition}
\newtheorem{definition}{Definition}
\theoremstyle{remark}
\newtheorem{remark}{Remark}

\makeatletter
\renewenvironment{proof}[1][\proofname]{%
  \par\vspace{0.6em}%
  \noindent\rule{\linewidth}{0.4pt}\par\nobreak\vspace{0.25em}%
  \pushQED{\qed}%
  \normalfont
  \begin{list}{}{%
    \setlength{\leftmargin}{1.5em}%
    \setlength{\rightmargin}{0pt}%
    \setlength{\listparindent}{0pt}%
    \setlength{\itemindent}{0pt}%
    \setlength{\labelwidth}{0pt}%
    \setlength{\labelsep}{0.5em}%
    \setlength{\topsep}{0pt}%
    \setlength{\partopsep}{0pt}%
    \setlength{\itemsep}{0pt}%
    \setlength{\parsep}{0.5\baselineskip}%
  }%
  \item[\itshape #1\@addpunct{.}]\ignorespaces
}{%
  \popQED\end{list}\@endpefalse
}
\makeatother

\makeatletter
\newif\ifquatkananon
\if@accepted\quatkananonfalse\else\quatkananontrue\fi
\makeatother

\ifquatkananon
  \hypersetup{pdftitle={LawKAN: Group-Law Layers for Composition},pdfauthor={},pdfsubject={},pdfkeywords={}}
  \newcommand{\eulerefcite}{\citep{anon2026eulerreg}}
  
\else
  \hypersetup{pdftitle={LawKAN: Group-Law Layers for Composition --- When Imposing a Known Law Pays Off},pdfauthor={Anna Mishchenko, Nikolai Gavrilin},pdfsubject={composition, group laws, Kolmogorov--Arnold networks},pdfkeywords={LawKAN, group law, KAN, robot calibration, SE(3), SU(2)}}
  \newcommand{\eulerefcite}{\citep{sun2026eulerkan}}
  
\fi

\newcommand{\qtilde}{\tilde{q}}
\newcommand{\so}{\mathfrak{so}}
\newcommand{\degunit}{^\circ}
\newcommand{\pverdict}{\ensuremath{\circ\mkern-6.2mu\vert}}

\newcommand{\appendixdivider}[1][Appendix]{%
  \clearpage
  \phantomsection
  \pdfbookmark[0]{#1}{appendix}%
  \begin{center}{\LARGE\bfseries\sffamily #1\par}\end{center}%
  \vspace{1.2em}%
}

\begin{document}

\title{LawKAN: Group-Law Layers for Composition\\ --- When Imposing a Known Law Pays Off}
\author{\name Anna Mishchenko, Nikolai Gavrilin \\
      \addr Independent researchers \email anna.v.mischenko@gmail.com}

\maketitle

\def\month{09}
\def\year{2026}

\begin{abstract}

Many tasks compose transformations by a known rule: chained rotations in robotics
and navigation. Learned models relearn it, compounding error; classical estimators
hardcode it and miss hardware deviations. We impose the known structure and learn
only the remainder: an exact non-commutative group law sits in the node of a
self-contained layer, its edges learn local increments, and a zero-initialized
correction $\psi$ learns unknown deviations of the law. A pre-training criterion predicts
exact representability. On fixed-angle $\mathrm{SU}(2)$ composition, a 54-parameter dense edge is
the most accurate dense model, at $27\times$ fewer parameters than the strongest fixed-matrix
baseline, and lies $0.05\degunit$ above the Bayes floor. On real robot arms, an isolated
306-parameter $\psi$ reaches 1.12--1.14\,mm versus 1.29--1.33\,mm for Gauss--Newton at $N\ge60$,
and the recipe transfers across three vendors and two protocols. The recipe matches classical fits
where the data obey the law and loses where the imposed law is not the task's.

\end{abstract}

\section{Introduction}
\label{sec:intro}

\looseness=-1
Composition tasks come with their rule known in advance, and the rule is a non-commutative
product. We ask how much of the model should \emph{be} that rule --- imposing the structure we know
and learning only the remainder.

\looseness=-1
The known structure operates at two levels: the \emph{group law} (the exact, non-commutative
symmetry of composition) and the \emph{form of the increment}, when supplied by the task (e.g., a
constant per-joint screw or a fixed analytic support). Consequently, only two things are learned: the values
inside the known form (the edge coefficients) and the deviation whose form is unknown. The natural
home for the law is the Kolmogorov--Arnold Network (KAN), which splits a layer into learnable edge
functions and a fixed additive node, a split exact for addition. The law to be embedded, however, is
multiplicative: composition is a fixed, non-commutative product. We ask what the split yields when
the node's law is a group law (a rule for composing two elements into a third), and under what
conditions embedding a known law in the node pays off.

\begin{figure}[!htbp]
\centering
\caption{The recipe and its rotation case. (a) Only the edge is learned (the
coefficients of the step): where the exponential
map covers the whole group, the edge emits an infinitesimal step that the node turns into a group
element; otherwise the edge emits the group element directly (e.g.\ $GL^{+}(2,\mathbb{R})$,
dashed). The node applies the exact law and carries no parameters itself, but may be modified by
an optional zero-initialized learned correction $\psi$ (grey; Section~\ref{sec:almost-theory}). (b) Rotations: Fourier edges over the axis
phases give a rotation vector, the exponential map lifts it to a unit quaternion, and the node
composes by the Hamilton product (for other groups, only the edge's target and the node's realization
differ, Section~\ref{sec:crossgroup}); an optional $\psi$ modifies the product there as well. A
parameter-annotated wiring of case (b) is Figure~\ref{fig:wiring}.}
\label{fig:recipe}
\begin{tikzpicture}[
  font=\figfont,
  box/.style={draw, rounded corners=2pt, align=center, inner sep=3pt, minimum height=8mm},
  learned/.style={box, fill=black!7},
  exact/.style={box, fill=blue!7},
  io/.style={box, fill=white},
  ar/.style={-{Latex[length=1.6mm]}, semithick},
  node distance=5mm
]
\node[io, text width=20mm] (x) {chart $x$};
\node[learned, right=of x, text width=23mm] (e) {edge (learned)};
\node[io, right=of e, text width=13mm] (l) {Lie alg.};
\node[exact, right=of l, text width=25mm] (n) {node: exp $+$ group law (exact)\\ \textcolor{black!60}{$+$ $\psi$ (optional, learned, $0$ at init)}};
\node[io, right=of n, text width=17mm] (g) {group element};
\draw[ar] (x)--(e); \draw[ar] (e)--(l); \draw[ar] (l)--(n); \draw[ar] (n)--(g);
\coordinate (eup) at ([yshift=5.5mm]e.north);
\coordinate (nup) at ([yshift=5.5mm]n.north);
\draw[ar, dashed] (e.north) -- (eup)
  -- node[above, font=\figfont\itshape, inner sep=1pt] {group elem.\ ($GL^{+}$)} (nup)
  -- (n.north);
\node[left=1mm of x, text=black!60] {(a)};
\node[io, below=5mm of x, text width=22mm] (p) {axis phases $(\alpha,\beta)$};
\node[learned, below=5mm of e, text width=25mm] (fe) {Fourier edge, $K{=}1$};
\node[io, below=5mm of l, text width=14mm] (v) {$v\in\mathbb{R}^3$};
\node[exact, below=5mm of n, text width=27mm] (q) {unit quaternion $q=\exp(v)\in S^3$};
\node[io, below=5mm of g, text width=22mm] (pr) {$q_2q_1$, probe readout\\ \textcolor{black!60}{$+$ optional $\psi$ (learned)}};
\draw[ar] (p)--(fe); \draw[ar] (fe)--(v); \draw[ar] (v)--(q); \draw[ar] (q)--(pr);
\node[left=1mm of p, text=black!60] {(b)};
\end{tikzpicture}

\end{figure}
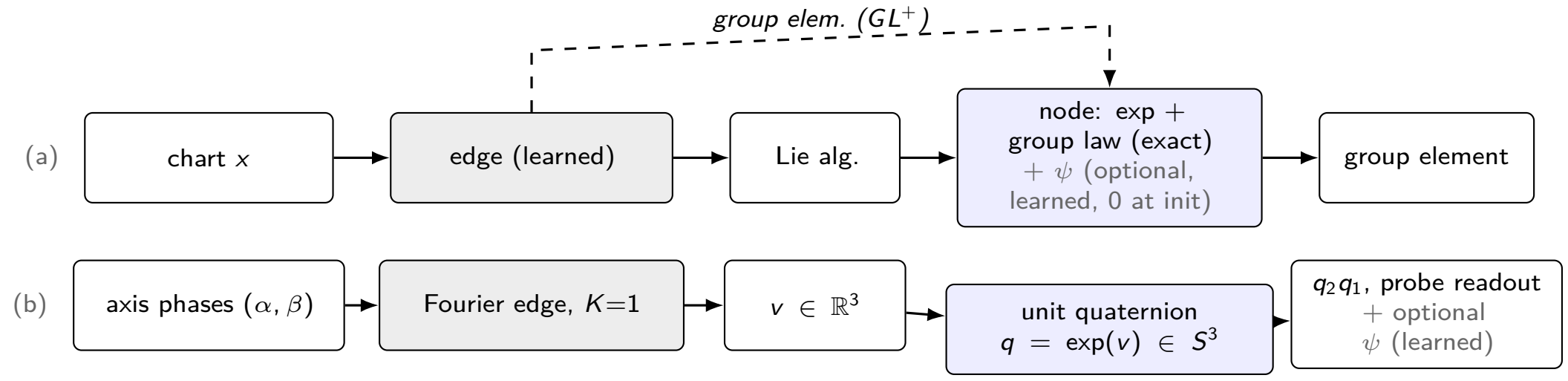

\looseness=-1
Our answer keeps the layer's edge/node split and changes its node. The node applies the known group
law exactly; a zero-initialized learned correction $\psi$ modifies it only where the hardware deviates
from the law, and at $\psi{=}0$ the node is exact. Each edge learns only the local step --- the
coefficients of a map from the input chart (a coordinate system on the group) into the group's
increment. Where the exponential covers the whole
group (the compact case and $\mathrm{SE}(3)$) the increment is such a step; where it does not
($GL^{+}(2,\mathbb{R})$) the edge emits the group element itself, and the node composes by the law
(Figure~\ref{fig:recipe}). Only the edge's coefficients are learned; with the law fixed, the outcome
is a reading of the data (Section~\ref{sec:map}), and the isolated correction of
Section~\ref{sec:almost-theory} learns the unknown part of a deviation. We call the family
\textbf{LawKAN}\footnote{\label{footnote:lawkan}%
\textbf{Law}KAN $=$ the composition \emph{law} $+$ \emph{KAN}. The name commits only to the KAN
edge/node split: learned coefficients on the edges, a fixed node --- here a non-commutative group law, not an
additive activation. It does not commit to a spline: the basis functions are fixed in advance, and only the
coefficients are learned. The Kolmogorov--Arnold superposition is not lost but relocated inside the edge: the edge is a sum
of fixed feature terms, while the node executes the law.}
with one instance per group: \textbf{LawKAN-SU(2)} for rotations, \textbf{LawKAN-GL$^{+}$} and
\textbf{LawKAN-SE(3)} (Section~\ref{sec:crossgroup}). The governing principle --- the law composes,
the edge spans --- is made precise in Section~\ref{sec:method} as two levels of known structure and a
representability criterion (Proposition~\ref{prop:span}); the rest measures the consequences.%
\footnote{Code and reproduction artifact: \ifquatkananon
\url{https://anonymous.4open.science/r/lawkan-6A4A/} (anonymous mirror during review)\else
\url{https://github.com/occamolab/lawkan}\fi.}

\paragraph{Contributions.}
\begin{itemize}
\item \emph{Known structure, imposed.} The composition is a \emph{group law} --- an exact,
non-commutative symmetry --- placed in the node, with the edge spanning the step: composition is
representable exactly when the edge carries the right coordinates for it
(Proposition~\ref{prop:span}), and better edge features then improve results; placing an exact group law in a KAN node is, to our knowledge, new.
\item \textbf{Learning the unknown deviation.} Where the deviation's form is not known,
a zero-initialized correction in the node learns it (trained in isolation) and doubles as a
diagnostic probe: it deviates from zero only when the data demand it, and its shape localizes the deviation.
\item \textbf{One recipe, across the group laws.} The same construction serves the
compact $\mathrm{SU}(2)$ (rotations), the semidirect $\mathrm{SE}(3)$ (rigid motions) and the
non-compact $GL^{+}(2,\mathbb{R})$ (positive-determinant $2{\times}2$ matrices); adapting it changes
only the group's chart, the edge's target and the node's operation.
\item \textbf{A measured map.} The exact composer wins on composition-structured tasks with a
small fraction of the parameters and saves data on a real arm at small training sets
($1.5$--$2\times$ lower error with $60$--$200\times$ fewer parameters), while converging far
faster than the matched MLP. It matches classical fits (least-squares or closed-form methods
that hard-code the same law) without tuning on real sensors and robots, and the learned correction
adds the deformed-law \emph{case} where the hardware deviates. The full outcome map is
Sections~\ref{sec:applied}--\ref{sec:limitations} and Table~\ref{tab:applicability}.
\end{itemize}

\paragraph{Headline results.} On real arm calibration, an isolated $306$-parameter correction to the
law matches or overtakes the same-class Gauss--Newton fit at every training size ($1.12$ vs
$1.29$--$1.33\,$mm for GN; exact node $1.37$--$1.40\,$mm at $N\ge60$) and transfers across three
robot vendors and two protocols. On the remaining real probes (consumer IMUs, a gyroscope, three
arms) the exact node matches the classical constant-fit reference without per-model tuning; where
the imposed law is not the task's, it loses. The full map is Section~\ref{sec:applied} and
Table~\ref{tab:applicability}.

\paragraph{Where it is useful.} The practical case is the one the applications meet: a task with a
\emph{law-side deformation} and \emph{scarce data}. There the exact edge fixes the law, so the
correction starts from the right model and learns only the deformation. On our arm task the two-slot
correction reaches $1.51\,$mm at $N{=}20$ and $1.24\,$mm at $N{=}40$, against $2.39$ and $2.03\,$mm
for the matched MLP --- $1.5$--$2\times$ lower error with $60$--$200\times$ fewer parameters --- and
our convergence check (Section~\ref{sec:convergence}) shows it reaching its plateau within a few hundred epochs,
while the matched MLP needs $500$--$2000$ and then overfits.

\section{Related work}
\label{sec:related}

\textbf{Rotation representations for learning.} Continuous rotation heads --- 6D
(Gram--Schmidt; \citealp{zhou2019continuity}) and SVD-orthogonalized 9D
\citep{levinson2020svd} --- are the output standard and our strongest baselines; continuity fixes
the codomain, not the algebra of composition, which such heads must still learn, and regression on
manifold-valued regression is studied separately \citep{bregier2021deep}.

\textbf{Building known structure into the network.} Equivariant architectures
\citep{cohen2016group,thomas2018tfn,fuchs2020se3,batzner2022nequip,deng2021vn} fix a
\emph{symmetry} (for a robotics survey see \citealp{seo2025se3survey}); Lie-group
reparameterizations \citep{falorsi2019} respect the manifold; Riemannian
optimization \citep{absil2008,boumal2023} and RPMG \citep{chen2022rpmg} fix the gradient side, and
decompositions of non-compact, non-abelian groups into charts \citep{mironenco2024lie} supply the
coordinate choices we use. The general principle --- impose the known operator, learn the remainder
--- is known-operator learning \citep{maier2022known}. For composition, these methods never place the
composition law itself as the node's operation: each fixes a structure but learns the remainder
wholesale, without separating the \emph{known form} --- which can be imposed exactly --- from the part
whose form is unknown. LawKAN takes this approach (Section~\ref{sec:method}). Equivariant models transport features under a global rotation rather than
compose scalar phases, so we measure the standard readouts as opponents instead.

\textbf{KANs and quaternion networks.} KANs \citep{liu2024kan} and MultKAN \citep{liu2024kan20} put
learnable functions on edges with an additive (or scalar-multiplicative) node; EKAN
\citep{hu2025ekan}, geometric \citep{alesiani2025geometric} and permutation \citep{elbaz2025fskan}
KAN extensions add symmetries, and polynomial or wavelet edges replace splines
\citep{sidharth2024chebykan,bozorgasl2024wavkan}, while range-aware Euler-angle
regression~\eulerefcite{} keeps the
additive node where bounded joint ranges make the target near-additive --- the complementary case
to composition, where the node must carry the law. Concretely, LawKAN inherits the KAN
edge/node split --- a fixed node plus learned edge coefficients --- but the node composes by a
non-commutative group law rather than adding, and the edge basis functions are fixed in advance; only the
coefficients are learned. Our use of the KAN label stays
modest: the node is the group's law, not an activation (footnote~\ref{footnote:lawkan}); a
controlled comparison likewise finds KANs ahead mainly on symbolic tasks \citep{yu2024kanmlp}. The
additive node cannot serve here: composing non-commuting increments through a finite
Baker--Campbell--Hausdorff expansion does not terminate, so a truncation leaves a positive floor
for generic inputs (Appendix~\ref{app:bch}), while the exact group law is a finite operation (the
Hamilton product costs $28$ scalar operations).
Quaternion-valued networks
\citep{parcollet2019qrnn,parcollet2020survey,zhang2020qpu,pavllo2019quaternet} put quaternion
arithmetic in the weights, not composition as a primitive.

\textbf{Learned calibration and inertial odometry.} Learned gyroscope calibration and inertial
odometry \citep{liu2022lgcnet,stolero2024rapid,zhang2025ckanio} are points on the
accuracy--complexity curve; TinyGC-Net \citep{cui2024tinygc} shrinks a calibrator to $27$
parameters for MCU deployment; configuration-dependent kinematics \citep{lu2025configuration}
interpolates local PoE models with a Fourier correction over the configuration --- a
configuration-indexed model, not a pair-level deformation of the law. Classical strapdown integration \citep{titterton2004strapdown} and
product-of-exponentials calibration are the constant-edge limits of our recipe
(Section~\ref{sec:method}).

\textbf{Positioning.} These lines build a known structure --- a symmetry or a model class --- into
the architecture, while the composition stays fixed arithmetic. We build the composition law itself
into the node (a non-commutative group law; to our knowledge the first KAN whose node carries one),
and, because real hardware deviates from the law, we let a zero-initialized correction learn the part of
the deviation whose form is unknown --- imposed where the form is known, learned where it is not. The
measured envelope (win/tie/lose) is in Section~\ref{sec:applied}.

\section{Method}
\label{sec:method}

\noindent\textbf{Object of study.} We study the case the paper's premise is about: a
\emph{composition}, where elements of a group $G$ are combined in a fixed order by $\circ$ (for
$\mathrm{SU}(2)$, the Hamilton product) and the law
itself is the structure we know and impose. This is the class of tasks --- orientation, poses,
calibration --- in which embedding the known law in the node and learning only the remainder applies.
The inputs are chart coordinates $x_1,\dots,x_n$ in the domain $U$ of a chart $\chi:U\to G$; the answer
is $R=\chi(x_n)\circ\cdots\circ\chi(x_1)\in G$, read out through a fixed map $\pi$ (for
$\mathrm{SU}(2)$, the direction of a rotated probe vector). The law $\circ$ is exactly the structure a
generic learner must otherwise relearn. We develop the construction on three representatives chosen to
exercise the design branches --- compact vs non-compact, global vs non-global exponential, pure
rotation vs mixed rotation--translation: the compact $\mathrm{SU}(2)$, the non-compact
$GL^{+}(2,\mathbb{R})$, and the mixed $\mathrm{SE}(3)$ (Section~\ref{sec:crossgroup}).

\looseness=-1
\noindent\textbf{The recipe.} Two pieces of structure are known in a composition task: the \emph{law}
of composition (the node) and the \emph{form of the increment} (what the task tells us about its
increments). Everything else is learned --- the \emph{values} (coefficients, always unknown) and the
\emph{deviation} (what the hardware adds, of unknown form). The recipe imposes the law exactly in the
node, realizes the edge in that known form, and leaves the values and the deviation to learning. A generic
network is the degenerate case in which nothing is imposed; a classical calibration estimator fixes
the law to the exact kinematics and the increment's form to a constant-screw. This paper walks the
middle; \textbf{LawKAN} is the recipe's instance when the known structure is a composition law placed
in the node.

\subsection{The outcome map}
\label{sec:map}

\looseness=-1
The outcome is not a claim we make but a reading of the data: whether it gives \textbf{parity},
a \textbf{win}, a \textbf{loss}, or \textbf{no gain} is fixed by whether the data \emph{obey} the
imposed law or \emph{deviate} from it, and in what form. Two objects can differ from the truth: the
\emph{edge}, which produces the increment from the chart, and the \emph{law}, which the node applies.
Reading each off the data gives the outcome; the two criteria of Section~\ref{sec:principle} decide it
before training. We measure the outcome against two reference points:

\begin{itemize}\itemsep2pt
\item the \emph{classical estimator} --- hard-codes the law and fits only its parameters. The
  recipe \emph{contains} it, by the span criterion of Section~\ref{sec:principle}: where the law holds and
  the edge spans the chart's logarithm, the trained model \emph{is} that estimator. Comparing with it
  is therefore a theorem, not an experiment.
\item a \emph{generic network} --- given no law, it relearns the composition from scratch. The recipe
  is a constrained instance of it (the law is fixed), so the empirical question is what that
  structure buys against it.
\end{itemize}

\looseness=-1
With these yardsticks we read the two objects in turn, edge first --- the edge is the primary learned
object, and the correction is trained only after the edge is frozen (Section~\ref{sec:almost-theory}):

\looseness=-1
\noindent\textbf{Edge (the increment).}
\begin{itemize}\itemsep2pt
\item \emph{the increment's form is right} (the deviation lies in the estimator's class) $\to$
  \textbf{parity}: the model \emph{is} the classical estimator, its coefficients learned from data
  (the fully known case is the ideal limit).
\item \emph{an unknown increment form that falls inside the edge's span} $\to$ \textbf{win}: the
  form is richer than the estimator's class.
\item \emph{an increment form outside the span} $\to$ \textbf{loss}: a floor no training removes
  (Proposition~\ref{prop:span}).
\end{itemize}

\looseness=-1
\noindent\textbf{Law (the composition).}
\begin{itemize}\itemsep2pt
\item \emph{no effect at the node} (the law matches) $\to$ the edge reading stands.
\item \emph{a deformation inside $\psi$'s span} $\to$ \textbf{win}: the classical estimator cannot
  carry a product-level deformation (Corollary~\ref{cor:law}).
\item \emph{a deformation outside $\psi$'s span} $\to$ \textbf{loss} (the projection remains).
\item \emph{the wrong law} (a different group or rule) $\to$ \textbf{loss}.
\end{itemize}

\looseness=-1
Common to either object: if the deviation is \emph{not identifiable} (it depends on an input not
given) or is \emph{noise}, the outcome is \textbf{no gain} --- the signal is not in the data, and no
structure recovers it (at most, add the missing input).

\looseness=-1
One further boundary lives not in the data but in the fit: even a win or a parity can be undone by an
unregularized correction trained on a small, seed-unstable set. We report it as a boundary of the
trained model, not of the data (Section~\ref{sec:almost-theory}).

\subsection{The construction, on the $\mathrm{SU}(2)$ example}
\label{sec:construct}

\looseness=-1
The construction is general --- any group $G$ with a chart whose analytic support can be written (the
criterion below asks nothing more) --- and we give it concretely on the compact $\mathrm{SU}(2)$, the
running example, noting at each step what a different group changes (Section~\ref{sec:generic}). For
$\mathrm{SU}(2)$ two pieces are known: the \emph{law} --- the Hamilton product the node composes ---
and the \emph{form of the increment} --- a fixed-amount rotation about a unit axis, carried by the
edge's analytic support over the axis phases $(\alpha,\beta)$. Transfer to the other groups is
measured on synthetic composition tasks (Section~\ref{sec:crossgroup}) and, where the law is real, in the
applications (Section~\ref{sec:applied}). The whole layer is:

\begin{center}
\small
\setlength{\fboxsep}{7pt}%
\fbox{\begin{minipage}{0.94\columnwidth}
\noindent\textbf{LawKAN-SU(2) layer (one composition).}
\textbf{Input:} axis phase pairs $(\alpha_i,\beta_i)$, in a fixed order; chain length $n$. The
increment is $\xi_i=\tfrac{\theta_i}{2}\,a(\alpha_i,\beta_i)$ --- rotation amount $\theta_i$, unit axis
$a(\alpha_i,\beta_i)$ --- and $\mathbb{H}$ matches $\mathrm{SU}(2)$ exactly.\\[3pt]
\begin{enumerate}\itemsep2pt\topsep2pt\parsep0pt
\item \textbf{Features (fixed).} $F_i=\{1,\cos\alpha_i,\sin\alpha_i\}\otimes\{1,\cos\beta_i,\sin\beta_i\}$ (the tensor product of the two feature sets, $9$ scalars per factor) --- the known increment form.
\item \textbf{Edge (coefficients learned, form fixed).} $\xi_i \leftarrow W_i F_i\in\mathfrak{so}(3)$, yielding the rotation vector.
\item \textbf{Node (exact, no parameters).} $q_i \leftarrow \exp(\xi_i)\in S^3$; $Q \leftarrow q_n\cdots q_1$, the Hamilton product.
\item \textbf{Readout (fixed).} $\pi(Q)$: the direction of $Q\hat e_z$.
\item \textbf{Optional correction (learned, zero at init).} $Q \leftarrow Q\exp\big(\psi(f(A),f(B))\big)$, on the physical composition (Section~\ref{sec:almost-theory}).
\end{enumerate}
\noindent\textbf{Parameters:} the $n$ edge matrices $W_i\in\mathbb{R}^{3\times9}$ ($54$ for $n{=}2$) and,
optionally, the $\psi$ head; the node and readout carry none.\\[2pt]
\noindent\textbf{Known-values limit:} with the amounts $\{\theta_i\}$ and axes known exactly, every
edge map is written down and nothing is learned --- the layer is a transcription of the task.
\end{minipage}}
\end{center}

\looseness=-1
With hand-set edge weights the machine reproduces the known rule exactly --- the algorithm written
down, not approximated; we state the lemma for $\mathrm{SU}(2)$ and do not claim validity for any
other instance (it verifies the constructive direction of Proposition~\ref{prop:span} at $K{=}1$; the criterion
itself is general, Section~\ref{sec:generic}). The classical estimators of the applications --- strapdown
integration, nominal product-of-exponentials kinematics --- are the zero-correction points of this
layer: where nothing is unknown, the layer is the classical algorithm.

\begin{center}
\small
\setlength{\fboxsep}{7pt}%
\fbox{\begin{minipage}{0.94\columnwidth}
\begin{lemma}[the layer is exact, $\mathrm{SU}(2)$]
\label{lem:kan-exact}
For every fixed $\{\theta_i\}$ there exist edge weights, depending only on $\{\theta_i\}$
and not on the input, such that the network computes the composition exactly:
$\widehat Q = Q$, hence $\pi(\widehat Q)=\pi(Q)$, for all inputs $(\alpha_i,\beta_i)$.
\end{lemma}
\begin{proof}[Proof sketch]
At $K{=}1$ the features
$\{1,\cos\alpha,\sin\alpha\}\otimes\{1,\cos\beta,\sin\beta\}$ contain the three axis
components as single features ($a_x=\cos\alpha\sin\beta$, $a_y=\sin\alpha\sin\beta$,
$a_z=\cos\beta$); weights $\theta_i/2$ on those, zero elsewhere, give
$\xi_i=(\theta_i/2)\,a(\alpha_i,\beta_i)$;
$\exp(\xi_i)=(\cos\tfrac{\theta_i}{2},\,a(\alpha_i,\beta_i)\sin\tfrac{\theta_i}{2})$
is the exact per-factor rotation; the node composes by the Hamilton product.
Full proof: Appendix~\ref{app:exact}.
\end{proof}
\end{minipage}}
\end{center}

\looseness=-1
The layer is fixed by $(G,\chi,K,n)$ and carries no width to tune: the only tensors are the edge
matrices $W_i$ (and, optionally, the $\psi$ head); the node and readout carry none.

\subsection{The two criteria}
\label{sec:principle}

If the layer is exact, what is there to learn? Two things: the \emph{increment's values} (the edge
coefficients) and the \emph{deviation} --- the departure of the data from the known, at the increment
or at the law. Two criteria decide the outcome \emph{before training}, and we use both throughout the
paper: the \emph{span criterion} (Proposition~\ref{prop:span}) for the increment, and the
\emph{law-match criterion} (Corollary~\ref{cor:law}) for the law. Both are proved in
Appendix~\ref{app:spanproof}.

\looseness=-1
\noindent\textbf{(i) The increment's values (coefficients).} The law and the increment's form are
right; the numbers are not --- a rotation amount, a joint's miscalibration, a sensor's increment map.
The learned object is the edge's coefficient vector $W$, over the known increment form. Whether the
edge can carry the unknown exactly --- and what floor remains if it cannot --- is decided before
training by the \emph{span criterion}:

\begin{criterion}[span]
\label{prop:span}
With a linear edge and the exact node, the architecture represents the composition exactly iff
the row space of $F$ contains a branch of $\log\chi$ (one consistent choice of the multi-valued
logarithm), up to the antipodal symmetry of the double cover --- where $q$ and $-q$ give the same rotation
(Remark~\ref{rem:doublecover}); if a needed component is absent, a positive error floor
\emph{independent of training} remains.
\end{criterion}

\looseness=-1
The criterion cuts both ways: a form too poor leaves the floor above, and an \emph{over-complete} form
is not harmless either --- features beyond $\log\chi$ are spare capacity that fits noise --- so the
\emph{minimal sufficient} form is preferred (pruning to the analytic edge beats the dense one). The constructive direction places $\log\chi$ in the edge weights and the converse
is in Appendix~\ref{app:spanproof}; in words, the node is never the bottleneck when its law is the
task's --- the chart and the feature set are.\footnote{For \emph{variable} angles, exposing the amount
$\theta$ as an input and extending the edge to the tensor product
$F'=\{1,\theta\}\otimes\{1,\cos\alpha,\sin\alpha\}\otimes\{1,\cos\beta,\sin\beta\}$ ($18$ features)
keeps exactness.}

\looseness=-1
\noindent\textbf{(ii) The deviation: of the increment or of the law.} The data may depart from the
known --- at the increment (a form the edge does not carry, absorbed by a richer edge form;
Proposition~\ref{prop:span}) or at the law (a product-level change the edge cannot cancel). By the
\emph{law-match criterion}, the correction $\psi$ absorbs the law side exactly when the deformation
lies in $\psi$'s span:

\begin{criterion}[law-match]
\label{cor:law}
With the edge as the only learner ($\psi\equiv0$), a node law differing from the chart's on a reachable
pair cannot be represented exactly. If the true law is written as $A\star B=(AB)\exp(\eta(A,B))$,
the edge-only floor is, to first order, $\Vert\eta\Vert$, independent of training; $\psi$ removes it exactly
when $\eta$ lies in $\psi$'s feature span, leaving its projection otherwise (Appendix~\ref{app:probes}).
\end{criterion}

The two results bound the design space: the edge must carry the right coordinates, the node the right
law. What is learned is always a deviation from the known, never the composition itself
(Section~\ref{sec:scope}).

\subsection{The law correction $\psi$}
\label{sec:almost-theory}

\looseness=-1
We stay in the $\mathrm{SU}(2)$ construction of Section~\ref{sec:construct} and now let the law deviate too.
When the data depart from the Hamilton product, we keep it exact and add a small learned \emph{law
correction},
\begin{equation}
R(A,B)=(AB)\exp\!\big(\psi(f(A),f(B))\big),
\end{equation}
with $A,B\in S^3$ (juxtaposition the Hamilton product), $f$ the pair features of the two factors, and $\psi$ a
small head whose output lies in $\mathfrak{so}(3)$, so $R$ is again a unit quaternion. The correction is
right-multiplied onto the \emph{physical} composition: a joint or a step. It is not applied at every
intermediate product along a long chain, where it would compound. Every head's last layer is \emph{zero},
so the relaxed model \emph{is} the exact model at initialization. This is exactly the \emph{law-match
criterion} (Section~\ref{sec:principle}): a product-level deformation $\eta$ is removed when it lies in
$\psi$'s feature span, and its projection left otherwise. The parameterization makes no additivity
assumption --- any perturbation of the Hamilton product is again a unit quaternion, right-multiplied as
$\exp(\eta)$.

\looseness=-1
What is known is imposed here too. When the form of the \emph{law's} deformation is known --- a
gravity-torque term, the sign of a backlash, a fixed per-joint offset --- it goes into an
\emph{analytic} $\psi$ with only its coefficients fitted (the known-deformation probes,
Appendix~\ref{app:probes}).

\looseness=-1
The learned heads are for the case where the law's deformation has no known
form. A learned head maps the pair features $f=f(A,B)$ into the algebra and, in general, combines two
parts --- a global and a local one --- as their \emph{superposition}
$\psi=\psi_{\mathrm g}+\psi_{\mathrm k}$ (the two corrections add in the algebra), so no per-task
choice is needed in the common case:
\begin{itemize}\itemsep2pt
\item a \emph{global} readout --- a smooth linear map $\psi_{\mathrm g}(f)=Wf$ with weights $W$ (a
  small MLP decoupled from the arm's physical links), which extrapolates a smooth deformation such
  as thermal drift or compliance;
\item a \emph{local} kernel --- an expansion
  $\psi_{\mathrm k}(f)=\sum_k c_k\exp(-\Vert f-f_k\Vert^2/2\sigma_k^2)$ over data centers $f_k$, with learned
  coefficients $c_k$ and widths $\sigma_k$ (the locality scale of each center); it is large where the
  data speak and decays off them,
  matching localized mechanisms such as backlash or friction, and conservative off the data.
\end{itemize}
A single head wins only in the
boundary regimes measured later: the kernel on a seed-unstable set (Section~\ref{sec:socket}); the global
where the deformation is smooth (Section~\ref{sec:abb}). The scopes are collected in Appendix~\ref{app:probes}
(Table~\ref{tab:slot}).

\looseness=-1
The experiments force two caveats. First, zero initialization is only a starting point. Training can
move $\psi$ away from the exact law, and an unregularized head can \emph{raise} the held-out error (the
two-slot on Kinova, Section~\ref{sec:socket}). Conservatism is a property of the map off the data, not a
guarantee for the trained model. Second, the correction doubles as a \emph{probe}. Under isolation
--- freeze the exact edge and train $\psi$ alone --- a nonzero correction is a candidate signal
against the law and localizes it. Joint training, by contrast, is non-identifiable: the edge absorbs part of the correction
and drifts into a compensated solution that does not transfer. Isolation removes the drift
(Appendix~\ref{app:probes}).

\subsubsection{What ``the data obey the law'' means}
\label{sec:obey}

\looseness=-1
The phrase is shorthand for a property of the classical estimator's residual --- the estimator being
the constant-edge limit of our layer with $\psi\equiv0$. Fit it and examine what it leaves. Three
cases: (i) the residual is at the noise floor --- the data obey the law; the exact node is already
the estimator, and no correction has a signal to find (Section~\ref{sec:parity}). (ii) The residual
is structured but lies inside the estimator's class --- a form gap on the edge side, to be closed by
a richer edge, not by a law correction (Section~\ref{sec:socket}, the socket FR3 unit). (iii) The
residual is structured and pair-dependent --- a deformation of the law, the case of this section.
The two criteria of Section~\ref{sec:principle} decide which case a dataset is before any training.

\looseness=-1
Physically, a law-side deformation makes the composition of two configurations differ from the
group's product. Three physical candidates were probed. Gravity load is ruled out as the dominant
term on the FR3: a physics-shaped correction is at parity with the exact model and worse than the
agnostic $\psi$ at $N\ge60$ (X19). Backlash appears in a synthetic probe with matched approaches:
the sign of the last joint motion predicts the residual direction (alignment $1.000$), and a
constrained correction in isolation removes $98\%$ of the hysteresis floor (X20); on the real arm
the signature does not rise above the $0.44\,$mm residuals, so that verdict stays inconclusive
(X21). The smooth-drift regime is represented by the range-only data, where a global head suffices
(Section~\ref{sec:abb}); a thermal origin is a hypothesis read off the head comparison, not a
verified mechanism. None of these forms can be assumed a priori on a new platform, which is why the
default correction is learned.

\looseness=-1
The correction doubles as the detector. Train $\psi$ in isolation on the frozen exact edge:
at $\psi=0$ the model is the exact-law model, so a nonzero correction is a candidate signal of a
law-side deformation, and its input dependence is diagnostic --- zeroing a channel group of the
pair features degrades the held-out error on both sides (X16; quantified in
Section~\ref{sec:almost}). On synthetic exact-law data the isolated correction gives no held-out
gain and overfits at small $N$ (S1): the verdict is read at held-out error, not from the norm of
$\psi$.

\subsection{Generalization to three group laws}
\label{sec:generic}

\looseness=-1
The construction above is the $\mathrm{SU}(2)$ instance. The recipe itself is general, so we now
separate what carries over to any group law from what has to be re-chosen for each.

The theory is general: the span criterion and its converse (the chart-spanning argument,
Appendix~\ref{app:crossgroup}) hold for \emph{any} group $G$ and chart satisfying hypotheses (i)--(iv)
of Appendix~\ref{app:spanproof}. The criterion that gates the edge therefore works before training for
any law a task supplies.

The law must be associative, not merely invertible. Associativity is what makes the fixed-order
product well-defined, and the tasks' laws are groups, whose inverses power the converse of the span
criterion (Appendix~\ref{app:crossgroup}). We therefore do not use a non-associative algebra such as
the octonions $\mathbb{O}$. The recipe needs only an associative law and a chart whose analytic
support can be written; the group case adds the inverses that power the converse of the span
criterion. Laws beyond groups --- monoid and semigroup products, such as strings or typed chains ---
therefore carry over unchanged, and we leave their study to future work (Section~\ref{sec:future}).

\looseness=-1
What is group-generic is the recipe's core: the node executes the law, the edges carry the increments,
the span criterion gates the edge, and the zero-initialized correction handles deviations. Exactly
three things are group-specific --- the \emph{chart}, the \emph{edge target}, and the \emph{node
realization} --- and the three representatives we test exercise the design-relevant branches:

\begin{itemize}\itemsep2pt
\item \emph{$\mathrm{SU}(2)$} --- compact, global exponential, periodic chart. Chart: axis phases
  $(\alpha,\beta)$; edge: $F_i=\{1,\cos\alpha,\sin\alpha\}\otimes\{1,\cos\beta,\sin\beta\}
  \to\mathfrak{so}(3)$; node: the Hamilton product.
\item \emph{$GL^{+}(2,\mathbb{R})$} --- non-compact, no global exponential. Chart: Iwasawa
  $(\theta,s_1,s_2,t)$; edge: the six analytic features $\to$ the group element itself; node: the
  matrix product.
\item \emph{$\mathrm{SE}(3)$} --- mixed, surjective exponential, semi-direct product. Chart: axis,
  angle, translation $p$; edge: $\{a,p,a{\times}p,a(a{\cdot}p)\}\to\mathfrak{se}(3)$; node: $\exp$
  then the rigid-motion product.
\end{itemize}

\looseness=-1
For each branch the constructive direction is realized with written weights over the chart's analytic
features: Lemma~\ref{lem:kan-exact} for $\mathrm{SU}(2)$, and the constructed weights of
Appendix~\ref{app:support} for $GL^{+}(2,\mathbb{R})$ (residual $1.1\times10^{-7}$) and
$\mathrm{SE}(3)$ ($1.6\times10^{-7}$). All three instances are constructively exact, the two
non-compact ones up to floating-point residual.

\subsection{Model summary}
\label{sec:summary}

\looseness=-1
The model has two parts. The \emph{node} is the composition operator: the \emph{exact} node applies the known group
law, and a \emph{relaxed} node adds the learned correction $\psi$. The
\emph{edge} is the learned feature map into the node's target (the algebra when the exponential is
global, the group element itself otherwise; Section~\ref{sec:generic}), and its type follows the task. We
write a model as \textbf{LawKAN-$\langle$group$\rangle$, edge: $\langle$edge$\rangle$, node:
$\langle$node$\rangle$} (e.g.\ LawKAN-SU(2), edge: Fourier ($K{=}1$), node: exact), where the group
names the instance (the law and chart), not the architecture. Tables~\ref{tab:edges}
and~\ref{tab:nodes} (Appendix~\ref{app:notation}) list the edge types and the node
variants studied here.

\section{Synthetic tasks: transfer, exactness, and limits}
\label{sec:crossgroup}

The synthetic program has two parts. One composition task per group asks whether the construction
transfers (the exact node, different charts and edges); the home case then asks what each part
contributes --- the exact node's representability, the relaxed node's correction under a deviating law
--- and where the advantage ends. Existence is settled in Section~\ref{sec:method}; what follows
measures the trained edges and the necessity side (blind ablations, floors). Each subsection states the result and delegates the construction, the mechanism and the
per-seed tables to the appendices.

\subsection{Across groups}
One composition task per group tests that the recipe transfers; Section~\ref{sec:generic} lists the
per-group construction. $\mathrm{SU}(2)$ is the home case --- exact already at $K{=}1$
(Lemma~\ref{lem:kan-exact}) and the one the sensor tasks use --- developed in Section~\ref{sec:minimal}.

On the non-compact $GL^{+}(2,\mathbb{R})$, where the exponential is not global, the constructed weights
are exact to $1.1\times10^{-7}$ (Appendix~\ref{app:support}) and the trained edge reaches $0.0024$
relative Frobenius, beating a $4996$-parameter MLP ($0.53$); a generic edge over raw coordinates
plateaus ($0.82$): the limit is the chart and the feature span, not the node. On the mixed
$\mathrm{SE}(3)$ (surjective exponential) the constructed weights are exact to $1.6\times10^{-7}$ and
the trained edge reaches $4.8\times10^{-4}$, orders below a dense linear edge ($0.39$) and the strongest
MLP ($0.33$, $h{=}64$; paired $p{=}3\mathrm{e}{-8}$); a blind feature-group test on held-out seeds confirms that both nonlinear
feature groups are necessary and that the floor scales linearly in the translation noise.
Table~\ref{tab:instances} summarizes the three, and Figure~\ref{fig:crossgroup} the transfer.

\begin{table}[t]
\centering
\small
\caption{The three instances: one exact node, different charts and learned edges
($GL^{+}(2,\mathbb{R})$: the edge emits the group element directly, since the exponential is not
global). Errors: $\mathrm{SU}(2)$ exact; $GL^{+}(2,\mathbb{R})$ relative Frobenius;
$\mathrm{SE}(3)$ per-coordinate probe. Details in Section~\ref{app:crossgroup}.}
\label{tab:instances}
\begin{tabular}{@{}>{\raggedright\arraybackslash}p{0.155\columnwidth}>{\raggedright\arraybackslash}p{0.18\columnwidth}>{\raggedright\arraybackslash}p{0.32\columnwidth}>{\raggedright\arraybackslash}p{0.245\columnwidth}@{}}
\toprule
Instance & Chart (input coords) & Learned edge $\to$ exact node & Result \\
\midrule
LawKAN-SU(2) & axis phases $(\alpha,\beta)$ & Fourier features $\to\so(3)$, exp $+$ Hamilton &
exact (Lemma~\ref{lem:kan-exact}) \\
LawKAN-GL$^{+}$ & Iwasawa $(\theta,s_1,s_2,t)$ & 6 product features (group elem.), matrix product &
$0.0024$ (dense $0.82$) \\
LawKAN-SE(3) & axis, angle, translation $p$ &
$\{a,\ p,\ a{\times}p,\ a(a{\cdot}p)\}\to\mathfrak{se}(3)$, exp $+$ rigid product &
$4.8\mathrm{e}{-4}$ (MLP $0.33$) \\
\bottomrule
\end{tabular}
\end{table}

The cross-group evidence is bounded: one synthetic composition task per group with
task-specific metrics, so error magnitudes are not directly comparable across different structures
--- a characterization rather than a strict taxonomy; the general statement is the criterion, and the transfer of the construction is what the
three branches test. Further laws ($\mathrm{SU}(N)$, symplectic, $\mathrm{sim}(3)$, nilpotent, Lorentz,
affine) are future work (Section~\ref{sec:future}).

\subsection{The compact case in depth}
\label{sec:minimal}

\looseness=-1
We study one case, \textbf{LawKAN-SU(2)} on rotations, in depth: it is the home case --- the compact
instance, and the one the sensor tasks use (Section~\ref{sec:parity}) --- and it admits a measured Bayes
floor, which makes every effect attributable.
The construction is in Appendix~\ref{app:construction}, mechanism details in Appendix~\ref{app:mech},
and per-seed and expanded tables in Appendix~\ref{app:tables}.
\subsubsection{The benchmark and the opponent matrix}
\label{sec:main}

\begin{figure}[!t]
\centering
\caption{Opponent matrix as a Pareto front (log--log): geodesic error vs parameter count (E1;
hollow markers: the same opponents at their best learning rate, E38); rotation amounts fixed, not
fed as input.
LawKAN-SU(2) ($54$ params) leads the front with a single point --- its count is fixed by the
construction (no width knob); the node/edge capacity curves are Figure~\ref{fig:nodecapacity}.}
\label{fig:pareto}
\includegraphics[width=0.60\columnwidth]{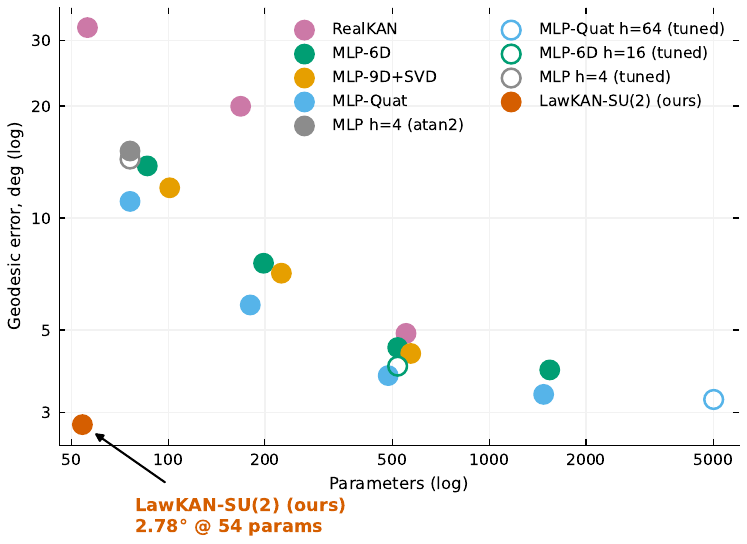}
\end{figure}

The benchmark composes two of these rotations, with fixed amounts (0.7 and 1.1 rad, not fed
as input) about random axes, and the network predicts the probe direction from the noisy axes.
Training follows the shared protocol of Appendix~\ref{app:repro}; the standard rotation heads
(6D, SVD-9D, normalized quaternion) share a cos/sin-featured MLP backbone, so only the head differs.

\begin{table}[t]
\centering
\small
\caption{Opponent matrix: geodesic error (deg), mean $\pm$ std, input/output noise
$\sigma{=}0.03$ ($h$: hidden width); the rotation amounts are fixed, not fed as input. The
full $21$-row table is Table~\ref{tab:matrix-full} (Appendix~\ref{app:compact}); the lower block
gives the same opponents at their best learning rate (E38).}
\label{tab:matrix}
\begin{tabular}{lrr}
\toprule
Model & Params & Error (deg) \\
\midrule
\textbf{LawKAN-SU(2), edge: Fourier ($K{=}1$), node: exact} & \textbf{54} & \textbf{2.78 $\pm$ 0.02} (5 seeds) \\
\midrule
\multicolumn{3}{@{}l}{\emph{Opponents}} \\
MLP + quaternion head $h{=}32$ & 1476 & 3.36 $\pm$ 0.06 (5 seeds) \\
MLP + quaternion head $h{=}32$ + RPMG \citep{chen2022rpmg} & 1476 & 3.42 $\pm$ 0.15 (5 seeds) \\
MLP + 6D $h{=}32$ \citep{zhou2019continuity} & 1542 & 3.91 $\pm$ 0.11 \\
MLP + 9D-SVD $h{=}16$ \citep{levinson2020svd} & 569 & 4.32 $\pm$ 0.46 \\
RealKAN + 6D head $h{=}16$ & 550 & 4.90 $\pm$ 0.26 \\
MLP $h{=}4$ (atan2 head) & 76 & 15.14 $\pm$ 1.44 (5 seeds) \\
\midrule
\multicolumn{3}{@{}l}{\emph{Tuned learning rate (E38):}} \\
MLP + quaternion head $h{=}64$, tuned & 4996 & 3.25 $\pm$ 0.02 (5 seeds) \\
MLP + 6D $h{=}16$, tuned \citep{zhou2019continuity} & 518 & 4.00 $\pm$ 0.54 (5 seeds) \\
MLP $h{=}4$ (atan2), tuned & 76 & 14.41 $\pm$ 1.05 (5 seeds) \\
\bottomrule
\end{tabular}
\end{table}

\looseness=-1
LawKAN-SU(2) is the most accurate model, with $27\times$ fewer parameters than the strongest
baseline in the fixed matrix ($92\times$ fewer than the strongest tuned baseline), and it comes
within $0.05\degunit$ of the measured Monte-Carlo Bayes floor --- the error of the best possible
predictor under this input noise ($+0.05\degunit$; strongest fixed-matrix baseline $+0.63\degunit$,
strongest tuned baseline $+0.52\degunit$). The learned edge matches the exact
construction it embeds rather than beating it: the exact composition under the analytic features is its
optimization instance, and the floor is its noise-limited value. The unstructured architectures must learn
the composition and lag behind. Neither a Lie-algebra baseline, nor baseline tuning, nor training on
the reported metric closes the gap, and the edge is roughly $32\times$ more sample-efficient than the
fixed-protocol 6D baseline (Figure~\ref{fig:sample}); against the strongest tuned baseline the factor
is about $7\times$ (E39). Tuning strengthens these baselines without changing the
ordering: the best cell of the registered grid (E32; MLP-Quat $h{=}64$, lr $3\cdot10^{-3}$) reaches
$3.25\pm0.02\degunit$ at $4996$ parameters, and the 6D head reaches $4.00\pm0.54\degunit$ at
lr $10^{-2}$ (E9/E38); both remain above the floor, and LawKAN's own sweep (E38) confirms
$10^{-2}$ as its best rate ($2.78$ vs $2.82$ at $3\cdot10^{-3}$). Figure~\ref{fig:pareto} shows the benchmark
(E1\footnote{Alphanumeric tags such as E1 or X26c are experiment IDs indexing the block map of Appendix~\ref{app:blockmap}.}) as a Pareto front;
the full $18$-configuration matrix and the protocol argument are in Appendix~\ref{app:compact}
(Table~\ref{tab:matrix-full}). The edge-side capacity variants are in
Figure~\ref{fig:nodecapacity}.

\subsubsection{Node versus edge: the mechanism}
\label{sec:nodeedge}

\looseness=-1
\paragraph{Which features, not how many.} On the benchmark of Section~\ref{sec:main}, the $K{=}1$
construction leaves only the edge's \emph{feature set} to choose. Pruning the $54$-parameter dense edge
to the three analytic features ($18$ parameters) is \emph{more} accurate, while a random equal-size mask
is far worse ($32.9\degunit$): the identity of the features, not their number, and automatic magnitude
pruning does not recover it. The exp-map node also avoids the singular $q/\Vert q\Vert$ regimes
(Appendices~\ref{app:construction}, \ref{app:mech}). When the rotation amount is supplied as an input,
the same edge stays at the floor for variable angles ($0.14$--$0.35\degunit$;
Table~\ref{tab:varangles}) where the fixed-angle edge leaves a $24$--$35\degunit$ floor.

Pre-registered isolation separates the two ingredients under the \emph{same} node. The spanning edges
(Fourier, MLP $h{\ge}8$) reach the best rows of Table~\ref{tab:isolation}
(Appendix~\ref{app:main-notes}); both no-node controls remain above them; and every matched edge below
$278$ parameters fails. The span criterion (Section~\ref{sec:principle}) is tested by the two linear-side
rows --- the $30$-parameter linear edge fails and the $18$-parameter analytic edge spans --- while the
$278$-parameter MLP threshold is a capacity observation, not a span test. The node is necessary but
not sufficient, and the analytic edge is the cheapest spanning point.
The full-$\mathrm{SO}(3)$ readout keeps the ordering (Table~\ref{tab:e30}); details and element
ablations are in Appendix~\ref{app:main-notes}. This is the \emph{exact} node's contribution --- the
representability side. Where the law itself deviates, the \emph{relaxed} node is what wins, and
Section~\ref{sec:deformed} measures that case.

\subsubsection{Boundaries: where the advantage ends}
\label{sec:stress}

\looseness=-1
All stress tests use the dense edge; the boundaries:
\begin{itemize}\itemsep2pt
\item \emph{noise}: linear degradation, dominant over the curve; the Fourier edge and the group law are
global out of band (Figure~\ref{fig:noise});
\item \emph{depth}: the advantage widens along chains up to $n{=}8$, where every architecture leaves
the regime ($n{=}6$ seed-bimodal; Appendix~\ref{app:tables});
\item \emph{hidden angles}: the margin narrows to the Bayes floor --- structure--task alignment,
not a universal parameterization win (Table~\ref{tab:applicability}).
\end{itemize}

\paragraph{Mismatch: the null task.}
\label{sec:null}
Where the imposed law is not the task's, the node loses: the Hamilton node reaches
$23.87\pm0.41\degunit$ against $12.64\pm3.00\degunit$ for an MLP ($h{=}4$). This is our violation of
the recipe's own premise, not an unknown structure (E7; Appendix~\ref{app:probes}).

\subsubsection{The deformed law}
\label{sec:deformed}

\looseness=-1
On synthetic data generated by an \emph{exact} serial-chain law the isolated correction gives no
held-out gain and overfits at small $N$ (S1). On data deformed by a known pair-dependent function it
recovers the noise floor ($1.40$ vs $11.5\,$mm at $N{=}160$), while the exact model stays at its
floor (S1; Figure~\ref{fig:lawdiag}(a), Appendix~\ref{app:probes}).

\section{Applications}
\label{sec:applied}

\looseness=-1
The two criteria of Section~\ref{sec:principle} predict three outcomes, and we test each on real data with known
ground truth: where a same-class classical estimator exists and the law holds, the model \emph{is}
that estimator (parity, with data efficiency); where the hardware deviates from the law and the deviation
falls in the correction's span, the correction wins; and where the imposed structure is not the
task's, it loses. A group-law subcomputation cannot be read off an end-to-end system, so we evaluate
the recipe in \emph{isolated} subcomputations whose ground truth is known.

\looseness=-1
Across the applications our models use the same optimizer (Adam with a cosine schedule) and the task's
own objective. Where a matched comparison is possible we fit the reference with the same optimizer and
loss (row~3 vs row~6 of Table~\ref{tab:armcal}; the socket reference), so those comparisons isolate the
architecture, not the optimizer; we also report the classical optimum on its own solver (GN), so
optimizer and loss effects can be told apart from structure. Our models use the recipe's constant
learning rate, without a per-seed tune; the learned opponents are given their best configuration from
the registered four-configuration grid (X22) --- a deliberately conservative asymmetry for our side.
Unless stated otherwise, $p$-values are per-comparison (two-sided paired $t$-test over shared
seeds; Appendix~\ref{app:seeds}); where a family of tests is reported, the correction is stated
(Benjamini--Hochberg for the socket table), and the remaining cells are exploratory.

\subsection{The deformed arm: the correction as remedy and probe}
\label{sec:almost}
\label{sec:armcal}

\noindent\textbf{Why this task.} The test needs a real device whose kinematics deviate from the ideal,
with ground-truth poses at several data sizes --- which separates a data-scarcity win from an
asymptotic one. Our test bed is the \emph{Franka Research 3} (FR3) robot, evaluated on $114$ real poses
\citep{philnad2024robotkinecal} (joint encoders plus camera-recovered flange poses). Performance is
scored by held-out end-point position error at $N{=}20/40/60/80$. The classical bar and the learned opponent are
the two poles of Section~\ref{sec:applied}. Section~\ref{sec:obey} makes ``obey the law'' operational:
it is a property of the classical estimator's residual, not an assumption about the hardware.

\noindent\textbf{Design: why these choices.} The classical pole is the product-of-exponentials least-squares
fit (PoE) and its Gauss--Newton optimum (GN, the classical estimator). The learned opponent is a residual MLP, tuned
to $18$k--$69$k parameters ($59$--$225\times$ the two-slot model's $306$). What is known about the task
then fixes the recipe:
\begin{itemize}\itemsep2pt
\item \emph{The group law.} The task is rigid-body motion, so the known structure is the exact
  $\mathrm{SE}(3)$ \emph{group law}, imposed in the node.
\item \emph{The form of the mismatch.} We impose the known per-joint form --- the classical estimator's
  corrected per-joint screw --- on the edge; whatever the hardware adds beyond it is left to $\psi$.
\item \emph{What is not known.} The mismatch's \emph{form} is unknown --- any
  pair-dependent deformation --- so we learn it with a small two-slot $\psi$ (a global linear trend
  with a size-dependent weight-decay dose, plus a local RBF kernel; $306$ parameters) on the frozen
  exact edge, trained in isolation; the two-slot is the default because no single head dominates
  across settings (Section~\ref{sec:almost-theory}).
\end{itemize}

\noindent\textbf{Experiments.} The headline: on the FR3 arm the isolated two-slot correction is below both the
exact node and the classical GN optimum at every $N$ ($1.51$ vs $1.56\,$mm at $N{=}20$; $1.12$ vs
$1.29$--$1.33\,$mm for GN at $N{\ge}60$; the exact node is $1.37$--$1.40\,$mm). A paired seed bootstrap ($10{,}000$ resamples) of row~7 minus
row~1 gives $-0.05\,[-0.18,+0.07]$ at $N{=}20$ and $-0.15\,[-0.25,-0.08]$,
$-0.14\,[-0.24,-0.06]$, $-0.21\,[-0.26,-0.16]$ at $N{=}40/60/80$.

\noindent The controls below show that this is a law correction, not capacity
and not an edge effect.

\begin{figure}[!ht]
\centering
\caption{(a) The correction wins: held-out error against $N$ for the exact node, the classical GN
optimum, the isolated two-slot correction, and the residual MLP; the two-slot is lowest at every $N$.
(b) The measured head comparison on the FR3 arm: the global (Fourier) head overfits at $N{=}20$; the two-slot
head is best in the data-starved regime, and at large $N$ the global and two-slot heads coincide within
noise --- no single head dominates. (c) On the FR3 arm, which inputs carry the learned correction (X16, Fourier head; the two-slot head
gives the same qualitative split, X16b): held-out error with one input channel group zeroed --- both sides of the pair matter; the inset plots
the correction norm $\Vert\psi\Vert$ against the B-side rotation angle $\theta_B$.}
\label{fig:learnedlaw}
\includegraphics[width=0.98\columnwidth]{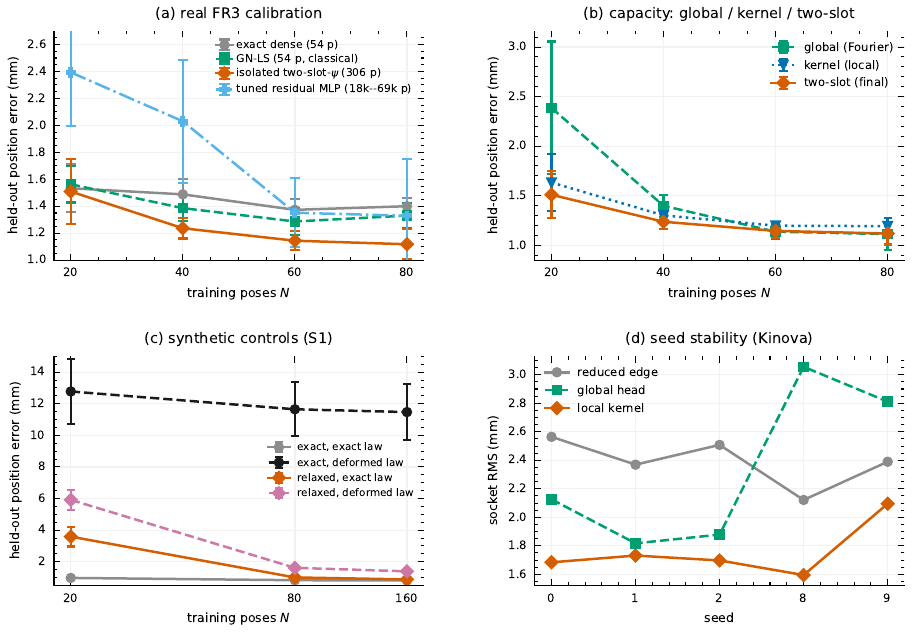}
\end{figure}

\begin{itemize}\itemsep2pt
\item \emph{Does the learned edge with the rigid law reproduce the classical estimator?} By the span
criterion the exact node \emph{is} the classical estimator of its class, and on real data its outputs
match the reference per-joint screw to machine precision (${\le}10^{-11}$) at every $N$
(Table~\ref{tab:armcal}, rows~3 and~6); the gap to the strongest optimum (row~1) is optimizer and
loss, not class. The next question is whether an isolated $\psi$ adds beyond this.
\item \emph{Is there an extra deformation the law does not carry?} The law-match criterion
  (Section~\ref{sec:principle}) says an isolated $\psi$ can win \emph{only} if the data carry a product-level
  deformation inside $\psi$'s span --- one the per-joint class cannot express. On the FR3 arm it wins: the
  two-slot correction (Table~\ref{tab:armcal}, row~7) is below the exact node (row~6) at every $N$
  (significantly for $N{\ge}40$: paired $p=0.028/0.007/0.003$ at $N{=}40/60/80$; $N{=}20$ $p=0.84$) and below the classical GN optimum (row~1) as
  well ($1.51$ vs $1.56\,$mm at $N{=}20$, within noise; $1.12$ vs $1.29$--$1.33\,$mm for GN at $N{\ge}60$, exact node $1.37$--$1.40$).
  The FR3 arm's deviation therefore has a component outside the per-joint class: the correction removes a
  floor the per-joint edge cannot. It beats the exact node under the same Adam optimizer, which we do
  not tune per task, and it also beats the best GN fit, where the optimizer is tuned.
\item \emph{Is the gain just capacity?} A strong learned opponent --- the validation-selected residual
  MLP (Table~\ref{tab:armcal}, row~4) --- is at parity with the exact node (row~6) for $N{\ge}60$, and even its test-oracle best
  (row~5) stays above the two-slot model (row~7). Extra parameters do not help, so the gain is not
  capacity.
\item \emph{Is the deformation real, and where does it sit?} The correction is pair-dependent: both
  sides of the pair carry weight --- setting either side's angle or translation channels to zero raises the
  error from $1.30$ to $3.75$--$4.71\,$mm (X16, Fourier head; Figure~\ref{fig:learnedlaw}a); the
  final two-slot head reproduces the qualitative split ($1.25$ to $3.35$--$4.38\,$mm; X16b). With the
  per-joint edge fixed, only $12\%$ of its effect field can be absorbed by the edge ($88\%$ orthogonal; G3,
  and the final two-slot head reproduces $88\%$; G3b), and training the edge jointly
  (Table~\ref{tab:armcal}, row~8) stays within $0.07\,$mm of isolation (row~7) --- a law-level
  correction rather than compensation of this edge (a richer edge could absorb more; that control is
  not run here). A node-only learned law (Fourier $\psi$, no per-joint edge; row~9) overfits at small $N$, so the
  learned per-joint edge matters.
\end{itemize}

\begin{table}[H]
\centering
\footnotesize
\setlength{\tabcolsep}{2pt}
\caption{Real-robot calibration (FR3 arm): end-point position error (mm; mean$\pm$std over $5$ seeds,
each a random split). Nominal $10.65\pm1.90\,$mm; the base-only fit saturates near $4.4\,$mm.
Superscripts give each row's optimizer and loss (the matched pair shares them, rows~3 and~6): $^{\rm A}$ Adam $+$
bounded $2(1-\cos)$ pose loss; $^{\rm G}$ GN on the position residual (strongest); $^{\rm P}$ PoE
least squares (weaker). The note marks the independent-implementation sanity check (agrees with the
exact node to machine precision, X42), the isolated versus joint (control) correction, and the
analytic-edge ablation; the oracle row picks the best configuration on the test set itself (it uses test information, so its error is an optimistic best-case bound, not a deployable model), reported for $N{\ge}60$. The opponent's width and learning rate are selected per seed on
validation data; the $18$k--$69$k range is that per-seed selection, so the capacity comparison is
against the best per-seed opponent.}
\label{tab:armcal}
\begin{tabular}{@{}rllrlrrrr@{}}
\toprule
\# & kind & model & params & note & $N{=}20$ & $N{=}40$ & $N{=}60$ & $N{=}80$ \\
\midrule
1 & reference & per-joint screw$^{\rm G}$ & $54$ & strongest & $1.56\pm0.14$ & $1.39\pm0.10$ & $1.29\pm0.10$ & $1.33\pm0.09$ \\
2 & reference & PoE$^{\rm P}$ & $34$ & weaker class & $1.72\pm0.22$ & $1.58\pm0.14$ & $1.47\pm0.08$ & $1.52\pm0.07$ \\
3 & reference & per-joint screw$^{\rm A}$ & $54$ & sanity check & $1.53\pm0.18$ & $1.49\pm0.11$ & $1.37\pm0.08$ & $1.40\pm0.06$ \\
4 & opponent & residual MLP$^{\rm A}$ & $18$k--$69$k & selected & $2.39\pm0.40$ & $2.03\pm0.46$ & $1.35\pm0.26$ & $1.33\pm0.42$ \\
5 & opponent & residual MLP, oracle$^{\rm A}$ & $18$k--$69$k & optimistic & --- & --- & $1.27\pm0.25$ & $1.26\pm0.37$ \\
6 & LawKAN & exact node$^{\rm A}$ & $54$ & --- & $1.53\pm0.18$ & $1.49\pm0.11$ & $1.37\pm0.08$ & $1.40\pm0.06$ \\
7 & LawKAN & two-slot$^{\rm A}$ & $306$ & isolated & $1.51\pm0.24$ & $1.24\pm0.08$ & $1.14\pm0.07$ & $1.12\pm0.11$ \\
8 & LawKAN & two-slot$^{\rm A}$ & $306$ & joint (control) & $1.52\pm0.31$ & $1.19\pm0.04$ & $1.08\pm0.06$ & $1.09\pm0.17$ \\
9 & LawKAN & node-only $\psi$ (Fourier) & $198$ & no per-joint edge & $2.55\pm0.70$ & $1.53\pm0.16$ & $1.20\pm0.09$ & $1.18\pm0.14$ \\
\bottomrule
\end{tabular}
\end{table}

\subsection{Convergence check: the model, not the budget}
\label{sec:convergence}

\noindent\textbf{The question.} On the FR3 arm calibration task the learned opponent --- the residual
MLP, trained under the same optimizer, loss and splits --- trails LawKAN-SE(3), node: relaxed
(two-slot) (Table~\ref{tab:armcal}, rows~4 and~7). Is that a limit of the MLP's model class, or is it
undertrained? A checkpoint grid answers this directly.

\begin{figure}[H]
\centering
\caption{Convergence check (X46 A1) on the FR3 arm calibration task of Table~\ref{tab:armcal} (not
the FR3 unit of the socket protocol, Table~\ref{tab:crossplat}): held-out error vs training epochs
(checkpoint grid $10$--$4000$, $5$ seeds) for LawKAN-SE(3), node: relaxed (two-slot), and the MLP opponent at
$N{=}20/60/80$. LawKAN reaches its plateau by $250$--$1000$ epochs; the MLP is best at
$500$--$2000$ and degrades after. The grid was run on this task only. The MLP opponent runs at its per-seed
selected width and learning rate ($h\in\{128,256\}$, $18$k--$69$k parameters, $10^{-3}$ or $10^{-2}$,
Table~\ref{tab:armcal}); the two-slot at $5\cdot10^{-3}$, so the early transient is the optimizer, not
the model.}
\label{fig:convergence}
\includegraphics[width=0.82\columnwidth]{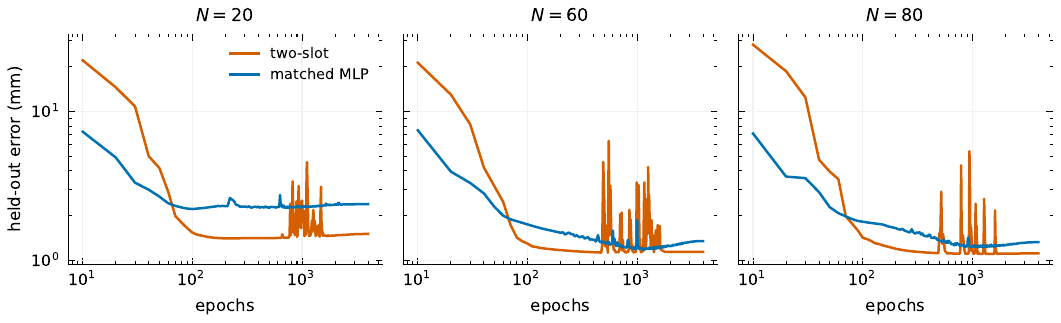}
\end{figure}

\noindent\textbf{What the grid shows.}
\begin{itemize}\itemsep2pt
\item \emph{LawKAN plateaus early.} It reaches its held-out error by $250$--$1000$ epochs and
  does not improve afterwards; extended $8000$-epoch runs are not better.
\item \emph{The MLP opponent peaks and then degrades.} It is best at $500$--$2000$ epochs and rises afterwards:
  at $N{=}80$, $1.24\,$mm at $1000$ vs $1.33\,$mm at $4000$; at $N{=}60$, $1.25\,$mm at $2000$ vs
  $1.35\,$mm at $4000$. More training makes it worse, not better.
\item \emph{On this task, the gap is the model.} Both models are past their own optimum, so the
  difference comes from the parameterisation, not from the training budget.
\end{itemize}

\noindent\textbf{Training dynamics.} Two different effects produce the remaining spikes.

\emph{Startup transient.} At lr $5\cdot10^{-3}$ LawKAN overshoots in the first tens of epochs
(up to ${\sim}24\,$mm) and recovers. Gradient clipping does not remove it; lr $10^{-3}$ cuts the peak
fivefold at the same final error (X52).

\emph{Residual chart spike.} At some seeds a smaller spike remains. Its cause is an \emph{operand chart
singularity}: the pair features log the accumulated pose, and when its rotation crosses $\pi$ the
axis--angle chart branches (the quaternion component $w_A$ changes sign), although the learned
correction $\psi$ itself stays small ($\lVert\omega_\psi\rVert\approx0.02\,$rad; X53-print-2). Three checks separate
mechanism from symptom:
\begin{itemize}\itemsep2pt
\item \emph{Features.} The quaternion-feature variant of LawKAN halves the peak but does not remove it, and the
  final error rises ($26\to13\,$mm; $1.09\to1.29\,$mm; X53-fix).
\item \emph{Metric.} A kernel defined by the group metric instead of input coordinates does not
  recover the final error either ($1.27$ and $1.39\,$mm; X60).
\item \emph{Trajectory.} Averaging the weights over training (Polyak/EMA) removes the spike in every
  configuration we ran (raw maxima of $6$--$40\,$mm fall to ${\le}1.6\,$mm) and leaves the final
  error within $0.02\,$mm (X59, X61).
\end{itemize}

\noindent\textbf{What follows.} The spike is a property of the training trajectory, not of the learned
solution: average the weights before deployment. A separate matched-protocol check at $N{=}60$
($200$ epochs, six seeds; X61, not a Table~\ref{tab:armcal} row) finds no cost for the branch-safe
variant (quaternion features, global slot only): mean $-0.062\,$mm, $90\%$ CI
$(-0.114,-0.011)$; the interval misses the $\pm0.11\,$mm margin by $0.004\,$mm, so
equivalence at the $\pm10\%$ margin is not established. We therefore keep the feature kernel as the
default and recommend the branch-safe variant for tasks that cross $\pi$.

\subsection{Transfer across vendors: the socket protocol}
\label{sec:socket}

\noindent\textbf{Why this task.} The deformed arm is a single robot; a general recipe should transfer
across vendors on a common protocol, where each unit's true deviation is unknown and the ground truth
is an independent metrology. The MUKCa socket-consistency protocol gives one task and metric on four
units (KUKA iiwa14, Panda, Kinova Gen3 lite, the FR3 unit): held-out root-mean-square (RMS) scatter (mm).

\noindent\textbf{Design.} The scheme is the one of Section~\ref{sec:almost}: the exact
$\mathrm{SE}(3)$ group law in the node, the full per-joint screw (the \emph{dense screw}) as the
classical reference, and the reduced edge (tool plus joint biases) with $\psi$ as the isolated
correction. The residual deformation is unknown; where it lies inside the full constant-screw class
that the reduced base does not contain, no correction can span it --- a base-\emph{form} boundary,
not an accuracy limit. Unlike configuration-dependent PoE interpolation \citep{lu2025configuration},
which fits local models plus a smooth, configuration-indexed correction, one exact law is imposed
here and the correction is a pair-level deformation of the composition, trained in isolation.

The reference and our models share the optimizer: on this protocol both the classical per-joint
screw and our edges are fitted with the same Adam schedule, so the comparison isolates the
architecture (the node's law and $\psi$). The classical Gauss--Newton fit differs in both optimizer
and loss; where we report it (the arm table, Table~\ref{tab:armcal}, row~1), the table separates the
two and we do not attribute the difference to the optimizer alone.

\noindent\textbf{Experiments.}
\begin{itemize}\itemsep2pt
\item \emph{Does the correction transfer across vendors?} On units whose deviation is pair-dependent
  and whose reduced base suffices, $\psi$ should win. Measured (Table~\ref{tab:crossplat}): it does on
  KUKA iiwa14 ($0.166$ vs the reduced edge $0.55\,$mm, $p<0.001$; and vs the dense screw $0.26\,$mm,
  $p{=}0.023$) and Panda ($0.330$ vs $0.93\,$mm, $p<0.001$; vs dense $0.49\,$mm, $p{=}0.007$); on the
  same KUKA in two further tool sessions it wins both ($0.23/0.38$ vs $0.61/0.75$), so the
  deformation is a property of the robot, not the setup.
\item \emph{Where does it stop, and is that a limit?} On units whose mismatch lies inside the
  constant-screw class, we predict a boundary. Measured: on the FR3 unit the full per-joint screw
  already spans the mismatch ($0.29$) and $\psi$ cannot beat the dense screw. Both boundaries are
  base-\emph{form} mismatches --- a base whose form nests the classical estimator would reduce them to
  parity, not loss (a design direction we do not measure here).
\item \emph{Is the collapse the correction's fault?} On the Kinova platform, the reduced edge underfits a large base
  mismatch, forcing $\psi$ to absorb a large residual; consequently, the socket objective is
  susceptible to a degenerate workspace collapse. Measured: the global slot is seed-unstable ($2/10$ seeds, Figure~\ref{fig:kinovaseeds}),
  while the kernel head removes it ($1.908$, matches the dense screw $1.98$, $p{=}0.10$) and stays
  below the reduced edge on every seed --- a small-data instability of the global slot, not of the
  correction.
\end{itemize}

\begin{table}[!htbp]
\centering
\footnotesize
\setlength{\tabcolsep}{3pt}
\caption{MUKCa socket consistency, held-out RMS scatter (mm); $10$ seeds for KUKA/Panda/Kinova, $5$ for the FR3 unit. All rows use Adam and the same socket loss. \emph{reference} = the full \emph{per-joint screw} (the classical estimator of the base's class, fitted with the \emph{same optimizer} as our models); \emph{LawKAN} = the reduced-edge base (tool $+$ joint biases) with $\psi{=}0$, global and \emph{sequential} two-slot heads (kernel first, then the global slot on the residual, dose $0$; a joint two-slot differs within noise on KUKA/Panda, is slightly better on Kinova ($p{=}0.061$) and worse on the FR3 unit ($p{=}0.017$), X47), and the per-joint-screw base with its $\psi{=}0$, global, kernel and two-slot heads (the per-joint-screw $\psi{=}0$ column reproduces the dense-screw reference). Bold marks the best cell per row, $\ddag$ an attribution boundary. $p$ = the declared head of each base: the reduced base's sequential two-slot, tested against its own edge ($p_{\rm edge}$, the correction's contribution) and against the dense screw ($p_{\rm dense}$, parity); and the per-joint base's best form (sequential two-slot on KUKA, global on Panda, kernel on Kinova, $\psi{=}0$ on the FR3 unit), tested against the dense screw; for the FR3 unit the declared form is the dense screw itself, so its entry is a parity by construction, not a tested $p$-value. Benjamini--Hochberg over the
eleven numeric tests in this table (taking $p{<}0.001$ as $0.001$): all reported
$p\le0.04$ survive at $q=0.05$; the Kinova edge ($p{=}0.483$, $q{=}0.48$) and the Kinova
per-joint dense entry ($p{=}0.101$, $q{=}0.11$) do not.}
\label{tab:crossplat}
\begin{tabular}{@{}ll r rrrrr rrrrr@{}}
\toprule
\multicolumn{13}{@{}l}{\emph{Socket consistency (MUKCa; $10$ seeds for KUKA/Panda/Kinova, $5$ for the FR3 unit)}} \\
\midrule
 & & reference & \multicolumn{5}{c}{LawKAN (reduced-edge base)} & \multicolumn{5}{c}{LawKAN (per-joint-screw base)} \\
\cmidrule(lr){3-3}\cmidrule(lr){4-8}\cmidrule(lr){9-13}
Platform & $N$ & dense screw & $\psi{=}0$ & global $\psi$ & two-slot $\psi$ & $p_{\rm edge}$ & $p_{\rm dense}$ & $\psi{=}0$ & global $\psi$ & kernel $\psi$ & two-slot $\psi$ & $p_{\rm dense}$ \\
\midrule
KUKA iiwa14 & $42$ & $0.26$ & $0.55$ & $0.187$ & \best{0.166} & $<\!0.001$ & $0.023$ & $0.26$ & $0.198$ & $0.218$ & $0.190$ & $0.011$ \\
Panda & $42$ & $0.49$ & $0.93$ & $0.365$ & \best{0.330} & $<\!0.001$ & $0.007$ & $0.49$ & $0.387$ & $0.466$ & $0.390$ & $0.0015$ \\
Kinova$^\ddag$ & $39$ & $1.98$ & $2.47$ & $2.269$ & $2.325$ & $0.483$ & $0.040$ & $1.98$ & $2.386$ & \best{1.908} & $2.397$ & $0.101$ \\
FR3 unit$^\ddag$ & $46$ & \best{0.29} & $1.49$ & $0.450$ & $0.441$ & $<\!0.001$ & $0.028$ & $0.29$ & $0.310$ & $0.300$ & $0.310$ & \emph{parity} \\
\bottomrule
\end{tabular}
\end{table}

\subsection{Range-only arm calibration}
\label{sec:abb}

\noindent\textbf{Why this task.} Range-only calibration tests whether the recipe holds when
the observation is a single distance per pose instead of the full pose. The ABB RobotCali draw-wire
protocol \citep{robotcali2021dataset,li2021overview} gives held-out mean $|\Delta L|$ (mm) at
$N{=}80/160/320$. The dataset has no explicit license; we use it under the authors' stated permission
for calibration and verification and do not redistribute the raw file.

\noindent\textbf{Design.} The underlying motion is rigid-body, so the exact $\mathrm{SE}(3)$ group
law is imposed in the node; the classical reference is the exact screw model and its Gauss--Newton
fit, the strongest classical form of the class; and the residual deformation is smooth, so the
isolated \emph{global} $\psi$ should suffice (the scheme of Section~\ref{sec:almost}).

\noindent\textbf{Experiments.}
\begin{itemize}\itemsep2pt
\item \emph{Does it beat the exact screw and reach the Gauss--Newton fit?} With the deviation in span
  and a smooth deformation, it should. Measured (Table~\ref{tab:rangeonly}): the global $\psi$ is
  below the exact screw at every $N$ and below the Gauss--Newton fit at $N\ge160$
  ($p{=}0.013/0.0004$); at $N{=}80$ it is numerically below GN as well, but the gap is not
  significant ($p{=}0.74$).
\item \emph{Does a richer head help?} The deformation is smooth, so a global head should suffice;
  we keep it and do not add a local kernel here.
\end{itemize}

\begin{table}[!htbp]
\centering
\footnotesize
\setlength{\tabcolsep}{3pt}
\caption{ABB RobotCali draw-wire (range-only), held-out mean $|\Delta L|$ (mm), $10$ seeds. \emph{reference} = Gauss--Newton on the range residual, the classical optimum; $\psi{=}0$ = the same class without the correction on our Adam $+$ bounded range loss; the $\psi$ head is the learned correction on our optimizer (Adam); $p$ = the global $\psi$ vs the reference. (A fit of the neural $\psi$ on the GN solver is unstable across runs, so we report $\psi$ on Adam only.)}
\label{tab:rangeonly}
\begin{tabular}{@{}lrrrc@{}}
\toprule
\multicolumn{5}{@{}l}{\emph{Range-only (ABB RobotCali, draw-wire; $10$ seeds)}} \\
\midrule
 & reference & \multicolumn{2}{c}{LawKAN} & \\
\cmidrule(lr){2-2}\cmidrule(lr){3-4}
$N$ & GN & $\psi{=}0$ (Adam) & global $\psi$ (Adam) & $p$ \\
\midrule
$80$  & $1.41$ & $1.53$ & \best{1.37} & $0.74$ \\
$160$ & $1.20$ & $1.38$ & \best{1.11} & $0.013$ \\
$320$ & $1.14$ & $1.33$ & \best{1.06} & $0.0004$ \\
\bottomrule
\end{tabular}
\end{table}

\subsection{Parity: the model is the classical estimator}
\label{sec:parity}

This section checks that the parity result reproduces on tasks of a different type: a
rotation-only sensor (Section~\ref{sec:imu-app}) and a low-dimensional calibration (Section~\ref{sec:gyro}). In
both the law holds exactly and a same-class classical estimator exists, so parity is the prediction:
the exact node should \emph{be} that estimator, with no distortion of the law and only estimation,
data-efficiency and gauge differences.

\subsubsection{IMU orientation}
\label{sec:imu-app}

\noindent\textbf{Why this task.} A rotation-only sensor is where a same-class classical estimator
exists --- the training-free strapdown integrator already carries the law --- so parity is the
prediction. Orientation increments over $0.07$--$5\,$s windows on the real BROAD sensor (geodesic
error against motion capture), a trial-level split.

\noindent\textbf{Design: why these choices.} $\mathrm{SU}(2)$ is the known group law; the exact node
uses one shared edge ($27$ parameters, the full $K{=}1$ group features, pruned to the
$18$-parameter analytic edge, and $36$ with a bias feature at long horizons), against the training-free
strapdown integrator (which already carries the law) and structure-free (flat) MLPs. The $36$-parameter variant is the
$27$-parameter edge plus the $3{\times}3$ bias-to-direction map ($9$ more); the flat MLPs are the two sizes ($h{=}8$,
$h{=}32$) carried verbatim from the synthetic opponent battery, where they were the strongest flat
opponents; three seeds ($0$--$2$) split the $16$ trials at trial level; we report equivalence under two one-sided tests (TOST) at a
$10\%$ margin as \emph{consistent with parity} --- with three seeds the interval is wide, so it
bounds the difference rather than establishing power. The theory then predicts parity.

\noindent\textbf{Experiments.} Parity (Table~\ref{tab:imu}): the exact node keeps the rotation
manifold where flat MLPs leave it --- the advantage is data efficiency.

\subsubsection{Gyroscope calibration}
\label{sec:gyro}

\noindent\textbf{Why this task.} A second parity case, on a low-dimensional calibration whose
classical fit has the same form. Calibration of the raw gyroscope ($\omega_{\mathrm{true}} =
M(\omega_{\mathrm{raw}}-b)$), held-out.

\noindent\textbf{Design: why these choices.} The known form is the linear $12$-parameter $(M,b)$
calibration, learned by the edge and compared against the classical Levenberg--Marquardt (LM) fit
of the same form. The theory predicts parity.

\noindent\textbf{Experiments.} Parity (Table~\ref{tab:gyrocal}): the learned calibration matches the LM fit to
$\le0.009\degunit$ on every window, and the TinyGC-style subnet to $\le0.002\degunit$; both fits are
deterministic (identical across seeds), so this is agreement between optimizers rather than a
statistical equivalence.

\subsection{The applied envelope}
\label{sec:envelope}

\noindent\textbf{The applied envelope.} Table~\ref{tab:applicability} collects the measured envelope:
parity where an analytic estimator exists and the law is exact; a win where the hardware deviates
from the law and the deviation lies in $\psi$'s span; and a loss where the imposed law is not the
task's. Which head to apply follows from the deformation and the setting
(Table~\ref{tab:whichhead}, Appendix~\ref{app:applied}): the two-slot is the default, the local
kernel is the collapse-safe choice for a small, seed-unstable set, and a global head suffices where
the deformation is smooth.

\subsection{Using the span criterion}
\label{sec:span-use}

\looseness=-1
It is a five-step check:
\begin{enumerate}\itemsep1pt
\item fix the group and the chart --- what composes, and in which coordinates the increment is
  expressed;
\item write the target's chart logarithm, the algebra element the edge must produce (or the group
  element itself when the exponential is not global, as for $GL^{+}(2,\mathbb{R})$);
\item test the edge's span: does its feature map cover those coordinates (or the chart element)? If yes, the exact node is
  representable (Proposition~\ref{prop:span}); if not, a training-independent floor remains;
\item for a deviating law, write the deviation $\eta$ and test whether $\psi$'s features span it (the
  law-match criterion); the projection left is the floor;
\item verify empirically: parity if the span is full, a floor if it is not.
\end{enumerate}
A companion script, \path{tools/span_check.py}, implements steps 2--5 as a linear probe from the
features to the logarithm (with a runnable \texttt{--demo}).
The applied tasks are instances: the per-joint screw spans the arm's constant-screw class (parity with
the classical fit); the reduced socket base does not span the dense screw class (a form gap,
Section~\ref{sec:socket}); the global $\psi$ spans the smooth range-only deviation (a win, Section~\ref{sec:abb});
and the IMU and gyroscope steps span exactly (parity, Section~\ref{sec:parity}).

\looseness=-1
\noindent\textbf{What it is good for.} \emph{Before training}, it screens an edge: a short span is a
training-independent floor, so a wider head or a longer schedule will not close it --- fix the
features instead. \emph{Diagnosing a plateau}, it separates a span limit from an optimization one,
and \emph{designing the correction} it states in advance whether $\psi$ can remove a measured
deformation and which feature set is the cheapest spanning one (the $18$-parameter analytic edge
over the $54$-parameter dense edge, Section~\ref{sec:minimal}). The probe needs only features and
targets, so it is a cheap first check on a new task.
\section{Limitations}
\label{sec:limitations}

\noindent\textbf{Scope.} The recipe is an architectural component, not a standalone, universal
alternative:
\begin{itemize}\itemsep2pt
\item the pruned edge assumes fixed known angles, and chain scaling holds to $n{=}4$;
\item the dense composer does not beat strapdown on real data (the pruned one does, modestly);
\item on $GL^{+}(2,\mathbb{R})$ a \emph{raw} (unstructured) dense edge does not beat an MLP, while
  the analytic-support edge solves the same task ($0.0024$ vs $0.53$, Section~\ref{sec:crossgroup}) --- the
  boundary is the edge's form, not the method;
\item without a composition law the model loses: the exact node presumes the law, and the correction
  belongs on the physical composition (a joint or a step).
\end{itemize}

\noindent\textbf{Measured failure modes.} The correction is not universal glue; its failures fall
into the three kinds that the recipe's own framing separates:
\begin{itemize}\itemsep2pt
\item \emph{Form boundaries} (different class): on the FR3 unit the mismatch lies inside the
  constant-screw class --- a form the classical estimator fixes and our reduced base does not contain
  --- so $\psi$ is the wrong tool (an attribution boundary, Section~\ref{sec:socket}); a base whose form
  nests the classical class would reduce this to parity.
\item \emph{Estimation gaps} (same class): on UR10 and UR5 the same-class Gauss--Newton fit is ahead
  at large $N$ --- a matter of optimizer and data, not of representable form (Appendix~\ref{app:probes}).
\item \emph{Identifiability}: on UR10 the freely-tuned edge drifts (test error $0.40\to11.7\,$mm with
  the correction off) and isolation removes it; on Kinova the global slot collapses on $2/10$ seeds
  until a \emph{local} head is used; at $N{=}20$ the correction overfits until a validation gate is
  added (Section~\ref{sec:armcal}, Section~\ref{sec:socket}, Appendix~\ref{app:probes}).
\end{itemize}

\noindent\textbf{External validity.} The evidence is narrow: one synthetic composition benchmark,
$39$ consumer-MEMS BROAD trials, six real calibration datasets across three tasks (FR3 arm; socket
consistency on KUKA iiwa14, Panda, Kinova Gen3 lite and the FR3 unit; ABB draw-wire range), and
$3$--$10$ seeds per platform. Tactical-grade IMUs, other robots, and large-scale vision or language
tasks are untested; the arm sets carry no approach trajectories, so backlash signatures are only
proxy-tested. The FR3 arm result comes from a single unit, and the socket FR3 unit is at parity by
construction, so unit-to-unit variability of the deformation is not measured. Signal-level denoising subnets (e.g., TinyGC-Net~\citep{cui2024tinygc}) are
complementary and out of scope.

\noindent\textbf{Evidence caveats.} One claim remains synthetic (rigid-body closure); gains trace to
structure--task alignment, not overly complex arithmetic. Opponents use default hyperparameters (the
registered four-configuration tune leaves the headline gap); the two-slot's weight-decay dose follows a
size-based rule selected on these splits (X26c), and transfer of the rule to a new protocol was not
measured; beyond the flat MLPs ($h{=}8$, $h{=}32$), the
IMU probe adds only a GRU and a residual MLP. On the real arm the learned model's edge over the
same-class Gauss--Newton fit is the data-scarce regime, and the shared loss is conservative for a
position metric. The dense IMU edge is seed-bimodal at $5\,$s (one seed reaches $73\degunit$), and
capacity does not cure it: a weight-decay dose leaves the outlier ($73.9\to76.8$), while the pruned
$18$-parameter edge stays stable (spread $0.017$) --- the deployment claim applies to the pruned
edge. Predictions are an authorial record: fourteen soft verdicts, one inconclusive proxy and one
post-hoc fix are weaker than clean confirmations, and the per-joint corrections do not prune on
hardware.

\section{Future work}
\label{sec:future}

\looseness=-1
The measured envelope points to the next steps. The recipe fixes the known structure in advance;
its natural generalisation is to \emph{grow} it: distil, during training, whatever is expressible
as a known law into the exact node, then learn only the remainder. Where several simple symmetries
coexist in the data, extract them separately and route between them:
\begin{itemize}\setlength{\itemsep}{0pt}\setlength{\parskip}{0pt}\setlength{\topsep}{1pt}
\item (i) carrying the layer into a real pipeline where the increment is latent and the surrounding
model is large;
\item (ii) state- and trajectory-conditioned channels on real data --- the mechanism probes motivate
velocity-conditioned edges and trajectory-instrumented collection (X19--X21);
\item (iii) broader manipulator coverage --- payloads, wear states, closed-loop tracking;
\item (iv) \emph{growing the known structure}: not learning a law we already know (vacuous by our
own principle), but distilling the law-like part of a learned solution into known form and learning
only the remainder;
\item (v) charts and group laws beyond the three;
\item (vi) composition laws beyond groups: monoid and semigroup products (strings, processes,
typed chains): node, edge and deformation carry over without inverses; the group assumption buys
the converse of the span criterion;
\item (vii) \emph{several laws at once} --- extracting several simple symmetries from the data and
routing between them;
\item (viii) a direct head-to-head with configuration-dependent PoE \citep{lu2025configuration} on a
shared dataset --- the closest learned competitor, not yet measured here.
\end{itemize}
\looseness=-1
A controlled synthetic sweep of deformation regimes (smooth, local, rough, mixed) against the three
heads would turn the applied map into a predictive rule; we leave it as a follow-up.

\section{Conclusion}
\label{sec:conclusion}

\looseness=-2
LawKAN embeds the group law directly in the node and delegates the coordinate mapping to the chart
edges; composition is representable exactly when the edge features span the chart's logarithm, and
better features then improve results. Where the hardware deviates from the law, a zero-initialized
isolated correction removes a floor that no edge-side structure removes, and doubles as a test that
localizes the deviation.

\looseness=-1
The value is architectural and conditional: on the composition tasks we study the exact composer
wins with a small fraction of the parameters and saves data on a real arm at small training sets,
while it matches classical fits under the same law and loses where the imposed law is not the
task's. This measured envelope --- when embedding a known law pays off, when it gives parity, and
when it loses --- is the paper's main object (Table~\ref{tab:applicability}).

\begin{table}[H]
\centering
\small
\setlength{\tabcolsep}{4pt}
\caption{When the group-law node pays off: the task-type map, the paper's main object (read a row by
task type; the verdict, its structural
reason and example tasks follow). The
structural reason is the analytic optimum (parity), identifiability (win/match), or the absence of
an analytic component. The last column names example tasks, measured here or in the literature.}
\label{tab:applicability}
\begin{tabular}{@{}>{\raggedright\arraybackslash}p{0.21\columnwidth}>{\raggedright\arraybackslash}p{0.10\columnwidth}>{\raggedright\arraybackslash}p{0.31\columnwidth}>{\raggedright\arraybackslash}p{0.31\columnwidth}@{}}
\toprule
Task type & Verdict & Reason & Example tasks \\
\midrule
Analytic estimator exists (gyro, strapdown; FK/arm calibration) & parity, tiny model &
the exact analytic optimum is already reachable; classical strapdown and PoE calibration are the constant-edge
limits of the same recipe (Section~\ref{sec:imu-app}, Section~\ref{sec:fk}, Section~\ref{sec:armcal}) &
strapdown integration; PoE arm calibration $\mathrm{SE}(3)$ \\
Latent but identifiable increment (robot calibration) & win / match & identifiable increments; the
gain is in the data-scarce regime, and a relaxed-node correction beats the classical fit at large
$N$ when the hardware deviates from the law (App.~\ref{app:probes}; Section~\ref{sec:armcal}) &
arm kinematic calibration $\mathrm{SE}(3)$; monocular SLAM alignment $\mathrm{sim}(3)$; inertial
odometry $\mathrm{SE}(3)$ \\
Latent increment with evolving state (rigid-body ODE) & only with the state-conditioned edge &
needs a recurrence / ODE edge (Section~\ref{sec:rigid}) &
rigid-body dynamics (Baxter, Section~\ref{sec:rigid}); trajectory-conditioned odometry \\
The imposed law is not the task's (null task) & loses & we impose the wrong law (Section~\ref{sec:scope}) &
generic regression targets (Section~\ref{sec:scope}) \\
\bottomrule
\end{tabular}
\end{table}

\looseness=-1
On the deformed arm, the training spikes are a property of the optimisation path, not of the learned
map: averaging the weights over training removes them and leaves the final error unchanged
(Section~\ref{sec:convergence}). A branch-safe quaternion variant showed no measured cost at the tested
budget, although equivalence at our margin was not established.

\looseness=-1
\paragraph{Verification.} This paper pre-specified $130$ predictions in its internal experiment
log before their runs ($119$ in the main log and $11$ branch probes); $29$ were refuted and are
kept and reported in Table~\ref{tab:verdicts-main}. Structure that is written down
can be \emph{checked}; structure that is learned must be \emph{audited} --- which is what the
isolation protocol and the measured envelope of Section~\ref{sec:applied}--Section~\ref{sec:limitations} do.

\paragraph{Data and code availability.}
The code is available \ifquatkananon at \url{https://anonymous.4open.science/r/lawkan-6A4A/}
(anonymous mirror; the public URL is withheld for double-blind review)\else at
\url{https://github.com/occamolab/lawkan}\fi{} under AGPL-3.0-or-later.
Experiments use BROAD \citep{laidig2021broad,broad2021dataset} (CC BY 4.0), MUKCa
\citep{franzese2025mukca,mukca2024dataset} (MIT), FR3/PhilNad
\citep{philnad2024robotkinecal} (MIT), IPA Series A
\citep{landgraf2021hybrid,ipa2021dataset} (MIT), pauljac UR5
\citep{yao2026graph,pauljac2025ur5} (MIT), the GTPN UR5 dataset
\citep{zenodo18954650} (CC BY 4.0), and the Baxter dataset
\citep{baxterzenodo} (CC BY 4.0). The ABB RobotCali dataset
\citep{robotcali2021dataset} was used under the authors' stated permission for
calibration and verification; it has no explicit license, and we do not redistribute
the raw file.

\phantomsection
\pdfbookmark[1]{References}{references}%
\bibliographystyle{tmlr}
\bibliography{quatkan}

\appendix
\appendixdivider

\section*{Appendix guide}
\noindent What each appendix contains (main-text anchors in parentheses):
\begin{description}\itemsep2pt
\item[A] Reproducibility, statistics protocol, experiment map (all).
\item[B] Notation and conventions.
\item[C] Exact representability: the span criterion and the floors (Section~\ref{sec:principle}).
\item[D] Extended per-seed tables and the cross-group characterization
(Section~\ref{sec:crossgroup}).
\item[E] Compact-instance details, opponent matrix, isolation and pruning
(Section~\ref{sec:main}, Section~\ref{sec:minimal}).
\item[F] Applied-probe details: p-values, CIs, three-platform tables (Section~\ref{sec:applied}).
\item[G] Isolated probes, learned-law controls, identifiability and diagnostics
(Section~\ref{sec:almost-theory}, Section~\ref{sec:almost}).
\item[H] Compact-instance construction: $K{=}1$, pruning.
\item[I] Mechanism details: variable angles, dead zones, order rule.
\item[J] The exp parameterization as a retraction.
\item[K] Registered predictions: verdict summary.
\item[L--M] Use of AI tools and broader impact.
\end{description}

\noindent\emph{First read: A, C, G are required; D--F are the number checks.}

\setcounter{figure}{0}
\renewcommand{\thefigure}{A\arabic{figure}}

\section{Reproducibility}
\label{app:repro}

\looseness=-1
Every claim was pre-specified in our internal experiment log, with its control, as a falsifiable
prediction before its run, and refuted predictions are kept. Paired tests, $95\%$ CIs and the multiple-comparison bookkeeping with
two one-sided tests (TOST) are in Section~\ref{app:seeds}; experiment indices, environment, seeding and
accounting conventions are in this appendix and Appendix~\ref{app:notation}. The registered suite
includes two blind necessity tests --- the feature-group ablation (a positive floor when a group is
removed: $6.1\mathrm{e}{-4}$ vs $0.387$) and the order-rule replication (linear floor scaling, ratio
$4.10$, with the $K{=}2$ numerical bound \emph{refuted} and reported) --- with per-seed numbers in
Appendices~\ref{app:crossgroup} and~\ref{app:mech}.

\subsection{Code}

The code, the canonical JSON dumps and the reproduction instructions are available under
AGPL-3.0 (research use is free): \ifquatkananon
\url{https://anonymous.4open.science/r/lawkan-6A4A/} (anonymous mirror; the public URL is
withheld for double-blind review)\else
\url{https://github.com/occamolab/lawkan}\fi{}. The repository is now \texttt{lawkan}; the
supplementary JSON dump keys and code identifiers keep the pre-rename spelling. Every displayed
number resolves through the \path{repro/canonical.json} manifest and can be re-derived with the
repository's check script; \path{tools/span_check.py} implements the span criterion as a linear
probe. Table~\ref{app:blockmap} maps the block groups to the paper locations.

{\small
\begin{longtable}{@{}l >{\raggedright\arraybackslash}p{0.86\textwidth}@{}}
\caption{Map of experiment blocks: each block and the paper location of its numbers.}
\label{app:blockmap}\\
\toprule
Block & Paper location / contents \\
\midrule
\endfirsthead
\toprule
Block & Paper location / contents \\
\midrule
\endhead
\bottomrule
\endfoot
E1--E14 & Section~\ref{sec:main}--Section~\ref{sec:stress}: the synthetic core (opponent matrix, null, boundaries) \\
E15--E37 & Section~\ref{sec:crossgroup}--Section~\ref{sec:minimal}: the group instances and edge/head variants \\
X5--X10 & Section~\ref{sec:applied}: the applied probes (IMU, gyro, arm, FK) \\
X37--X52 & Section~\ref{sec:socket}--Section~\ref{sec:parity}: socket heads, ABB range-only, matched-optimizer and capacity/locality probes \\
X11--X36 & Section~\ref{sec:almost}--Section~\ref{sec:parity}: the phase-2 probes and controls \\
X53--X61 & Section~\ref{sec:convergence}: the pole-fix branch (chart singularity, EMA, group-metric kernel, T1/T2); scripts \path{X53_socket_convergence.py}--\path{X61_confirm.py} \\
S1--S8 & Section~\ref{sec:deformed}, Section~\ref{sec:applied}: the synthetic and relaxed controls \\
\end{longtable}
}

\emph{The embedded-component experiments of the first version are excluded from this version by the
repositioning to a stand-alone micromodel.}

\emph{Synthetic numbers are verbatim from the reference run; real-data numbers are taken from the dumps listed in Table~\ref{tab:figprov}. Error is geodesic, degrees, ``$\pm$'' std across seeds.}

\subsection{Task and data}

\textbf{Composition $n{=}2$} (\texttt{make\_compn\_data}): axes $a_1, a_2$ are normalized Gaussian
vectors (uniform on $S^2$); rotation angles fixed: $\theta_1 = 0.7$, $\theta_2 = 1.1$ rad (for
$n{=}3$: 0.7, 1.1, 0.9). Input $X = (\alpha_1, \beta_1, \alpha_2, \beta_2) + \mathcal{N}(0,
\sigma^2)$, $\sigma = 0.03$ rad; target $Y = (\alpha, \beta)$ of the rotated probe vector
$(q_2q_1)\cdot \hat{e}_z + \mathcal{N}(0, \sigma^2)$. Train: 800 examples, test: 4000 (generated
with a different seed).

\textbf{Extrapolation} (\texttt{make\_comp\_banded}): axes are sampled area-uniformly within a
polar band: $\cos\beta \sim U[\cos\beta_{\max}, \cos\beta_{\min}]$, $\alpha \sim U[-\pi, \pi]$.
Train: ``tropics'' $\beta \in [0.5, 2.6]$ for both axes, 1000 examples. Tests of 2000 examples
each: inside / axis-1 pole ($\beta \in [0, 0.3]$) / axis-2 pole / both poles. Area uniformity
matters: uniform sampling in $\beta$ would over-densely cover the poles and distort the metric.

\textbf{$\mathbb{Z}_n$} (\texttt{make\_zn}): input phases are random group elements plus Gaussian
noise $\sigma = 0.30$ rad; target is $(a{+}b) \bmod n$. Train 400 examples, test 4000. The Bayes
ceiling is computed by Monte-Carlo over 200{,}000 examples (seed 7).

\textbf{Later blocks.} The applied, full-rotation and phase-2 isolation blocks are specified in
their sections; the generator above is the canonical one for the core synthetic battery
(E1--E14).

\subsection{Optimization}

Core protocol (E1--E14; deviations below): full-batch Adam, 3000 epochs (2500 in E3.2--E3.4 and E4), lr = 1e-2
for LawKAN-SU(2) and 3e-3 for all opponents, gradient clipping 1.0, geodesic loss on $S^2$:
$\mathrm{mean}(1 - \cos\Delta)$. Opponent learning rates were not tuned; they receive the
standard 3e-3 for such networks; LawKAN-SU(2) is stable at 1e-2 (itself a consequence of the
parameterization, see Appendix~\ref{app:retraction}). One exception: the $\mathbb{Z}_n$ models of
E5 (GroupSumKAN and the additive baseline, both with $K{=}7$ Fourier features) train at lr = 3e-2,
a regime chosen for convergence speed on the classification-style readout, applied identically
to both compared models.

Later blocks keep the same optimizer and registration discipline, with the budgets, learning rates
and splits stated in their sections; Table~\ref{tab:seeds} lists the seed counts.

\subsection{Seed protocol}
\label{app:seeds}

A reported $p$ is the two-sided $p$-value of a paired $t$-test over the seeds shared by the two
models; intuitively it is the \emph{significance of the result}: the probability of seeing a gap at
least this large if the two methods were equally good, so a small $p$ means the gap is unlikely to be
chance (it does not by itself say \emph{which} method wins --- that is the sign of the difference and
the table entries).

For each repeat $s$: training data \texttt{seed=s}, test data \texttt{seed=100+s}, network init
\texttt{torch.manual\_seed(100+s)}. Repeat counts (Table~\ref{tab:seeds}):

\begin{table}[H]
\centering
\small
\caption{Seed counts for the core and applied blocks (E1--X10; E-numbers of Table~\ref{app:blockmap}). Later blocks state seeds in their captions and in \texttt{repro/registry.json}.}
\label{tab:seeds}
\begin{tabular}{ll}
\toprule
Experiment & Seeds \\
\midrule
E1: LawKAN-SU(2), MLP-Quat $h{=}32$, MLP-Quat $h{=}32$+RPMG, MLP-6D $h{=}16$, MLP $h{=}4$ & 5 \\
E1: remaining 13 rows & 3 \\
E3.1 ($n{=}3$), E7 (scope), E8 (chains), E9 (LR sweep) & 3 \\
E3.2 (noise), E3.3 (sample eff.), E4, E12 (paired t), E13 ($K$ ablation) & 5 \\
E3.4 (extrapolation), E5 ($\mathbb{Z}_n$) & 2 \\
E6 (dead zones), E11 (Lie baseline), E14 ($n{=}6$) & 10 \\
E15/E17 and the FK probe (applied) & 3 \\
E16v2, E20, E21, E22 & 5 \\
E23, E24 (blind, held-out 20--24) & 5 \\
E25, E25b (held-out 30--34) & 5 \\
E26b, E26c (squared/random 70/30, transfer circle) & 3 \\
S1 (synthetic exact/deformed law, $N{=}20/80/160$) & 3 \\
S2 (UR5-GTPN relaxed) & 5 \\
X12/X12b/X13/X14/S3 (FR3 calibration, X7 splits) & 5 \\
E29 (held-out 50--54) & 5 \\
E30 & 5 \\
E19b (random equal-size masks; 5 masks per seed) & 5 \\
E31, E31b, E31c (node/edge isolation family) & 5 \\
E31d (pruning curve, E21 recipe) & 5 \\
E32 (opponent tuning grid) & 3 \\
E33 (floor recompute), E33b/E33c (metric-loss runs) & 5 \\
X5--X9 (real data, trial-level splits) & 3 (X7: 5 splits) \\
X10 (real calibration, 3 platforms, held-out 50--54) & 5 \\
\bottomrule
\end{tabular}
\end{table}

Documented exceptions: E5 uses init \texttt{torch.manual\_seed(50+s)} (the initialization seeds
$50$--$51$ overlap the held-out data-seed range $50$--$54$, but the data draws are disjoint); E6 varies only the init
(\texttt{seed=s}) with the data draw fixed (seed=0) across all 10 runs (a dead-zone ablation needs
identical data so only the trajectory differs); E3.3/E4 share one fixed test draw (\texttt{seed=100})
across all repeats and $N$ (strengthening the pairing); E3.4 varies only the init with the training
draw fixed (seed=0). Otherwise seeds are fully fixed and shared
across models within each cell (the phase-2 controls reuse the main range $0$--$4$; E32 uses $0$--$2$),
pre-specified before their runs; the paired-statistics protocol (Section~\ref{sec:main},
Appendix~\ref{app:tables}), reproducible up to cuDNN nondeterminism. The held-out blocks use disjoint
seed ranges (20--24, 30--34, 40--44, 50--54) that were not used for tuning. The script logs mean$\pm$std and raw
per-seed values per cell and ends with a JSON dump; the tables are regenerated from it mechanically.

\paragraph{Multiple comparisons and equivalence tests.} Because the battery
is large, individual $p$-values are \emph{per-comparison} (each is the two-sided $p$-value of a paired
$t$-test over the seeds shared by the two models, on identical splits --- the \emph{significance of the
result}, i.e.\ the probability of a gap at least this large if the two methods were equally good, so a
small $p$ means the gap is unlikely to be chance, and it does not by itself say which method wins; the
test's sample size is the number of seeds); a Benjamini--Hochberg correction over the
headline set (main matrix, sparse/dense, order rule, IMU, arm) leaves every reported significant gap
significant (largest adjusted $p\approx2.6\times10^{-3}$), and extended to \emph{all} $87$ recorded
per-comparison tests, $72$ survive at $q{=}0.05$ with every gap we describe as significant remaining.
Parity cells are tested for equivalence rather than inferred from non-significance: TOST at a $10\%$
margin where seed variance exists (the arm and IMU fits), and exact deterministic differences where
it does not (the gyroscope fits). Wall-clock is the only noisy axis
(Section~\ref{app:wallclock}).

\subsection{Wall-clock protocol (E2)}
\label{app:wallclock}

Batch 1024, 300 measured iterations after 30 warmup; train step = forward + backward +
optimizer.step, inference = forward under \texttt{no\_grad}. Measured on one machine in one
process, without parallel load. The result follows CUDA kernel-launch overhead rather than FLOPs
(see Section~\ref{sec:accounting}): at this model scale (54--1542 parameters) this is expected. Reference
numbers: LawKAN-SU(2) $4.57$ ms/it train and $1.08$ ms/batch inference vs $2.36$--$3.20$ ms/it and
$0.19$--$0.40$ ms/batch for the opponents. Wall-clock is the only noisy
axis: reruns on the same GPU vary by tens of percent (LawKAN-SU(2) $4.6$--$8.5$ ms/it across runs),
and near-identical timings for different opponents reflect measurement noise, not copied values.
Accuracy, parameter counts and FLOPs are exact.

\subsection{Figure provenance}

All vector figures are generated directly from the reference run numbers (not from
intermediate CPU versions) by the scripts in \texttt{tools/figures/} of the code repository. File names are stable
identifiers and do not track figure numbers (the mapping is in Table~\ref{tab:figprov}); the
\textsc{tikz} figures (recipe, wiring, support) are drawn inline in the paper source
and have no separate file. All plots share one style module, \texttt{figstyle.py} (single
font, size and Okabe--Ito colour palette), so the figures are visually consistent
(Table~\ref{tab:figprov}).

\begin{table}[H]
\centering
\small
\caption{Figure provenance: each vector figure, its script and data source (E-numbers of
Table~\ref{app:blockmap}).}
\label{tab:figprov}
\begin{tabular}{@{}l >{\raggedright\arraybackslash}p{0.17\textwidth} >{\raggedright\arraybackslash}p{0.17\textwidth} >{\raggedright\arraybackslash}p{0.42\textwidth}@{}}
\toprule
Figure & File & Script & Data source \\
\midrule
Figure~\ref{fig:crossgroup} & \path{fig02_crossgroup.pdf} & \path{fig02_crossgroup.py} & E16 v2 ($GL^{+}(2,\mathbb{R})$), E22 (SE(3)) \\
Figure~\ref{fig:nodecapacity} & \path{fig09_node_capacity.pdf} & \path{fig09_node_capacity.py} & E31/E31b/E31c, E13; controls E1/E11/E32; floor E33 \\
Figure~\ref{fig:pareto} & \path{fig04_matrix_pareto.pdf} & \path{fig04_matrix_pareto.py} & E1 (opponent matrix, error vs parameters; main text) \\
Figure~\ref{fig:sample} & \path{fig05_sample_efficiency.pdf} & \path{fig05_sample_efficiency.py} & E4 \\
Figure~\ref{fig:noise} & \path{fig12_noise_extrap.pdf} & \path{fig12_noise_extrap.py} & E3.2 (noise), E3.4 (pole extrapolation) \\
Figure~\ref{fig:imu} & \path{fig08_imu.pdf} & \path{fig08_imu.py} & X5 (BROAD, error vs horizon) \\
Figure~\ref{fig:x10} & \path{fig10_x10_platforms.pdf} & \path{fig10_x10_platforms.py} & X10 (three real platforms; P48) \\
Figure~\ref{fig:learnedlaw} & \path{fig13_learned_law.pdf} & \path{fig13_learned_law.py} & (a) X16; (b) X14/X34; (c) X7bench/X35/X22 \\
Figure~\ref{fig:kinovaseeds} & \path{fig16_kinova_seeds.pdf} & \path{fig16_kinova_seeds.py} & X32 (Kinova seed stability) \\
Figure~\ref{fig:lawdiag} & \path{fig14_law_diagnostics.pdf} & \path{fig14_law_diagnostics.py} & (a) S1; (b) X11/S2 \\
Figure~\ref{fig:convergence} & \path{fig17_convergence.pdf} & \path{fig17_convergence.py} & X46 A1 (checkpoint grid $10$--$4000$, $5$ seeds) \\
Figure~\ref{fig:newplat} & \path{fig15_new_platforms.pdf} & \path{fig15_new_platforms.py} & RC1 (ABB draw-wire), MK1 (MUKCa socket, KUKA) \\
Figure~\ref{fig:recipe} & \path{(TikZ, inline)} & \path{quatkan_body_v2.tex} & recipe diagram (no dump) \\
Figure~\ref{fig:wiring} & \path{(TikZ, inline)} & \path{quatkan_appendix_v2.tex} & construction wiring (no dump) \\
Figure~\ref{fig:support} & \path{(TikZ, inline)} & \path{quatkan_appendix_v2.tex} & analytic support (no dump) \\
\bottomrule
\end{tabular}
\end{table}

\section{Notation and conventions}
\label{app:notation}

\subsection{Groups and charts}
A task composes elements of a group $G$ (operation $\circ$; for $\mathrm{SU}(2)$ it is the Hamilton product, written by juxtaposition, and for $\mathrm{SE}(3)$ and $GL^{+}(2,\mathbb{R})$ the matrix product) in a fixed order. A chart
$\chi:U\to G$ maps coordinates to group elements ($U$: the chart's domain; $x_i\in U$ the
per-factor inputs in composition order); $R=\chi(x_n)\circ\cdots\circ\chi(x_1)$ is the composed
element and $\pi$ the readout evaluated on it; $\exp$ is the exponential of the group's Lie
algebra and $\log$ its inverse chart ($\log\chi$ the chart's logarithm); $F$ is the edge's
feature map with coefficient matrix $W$; $K$ is the Fourier order and $n$ the chain length.

Throughout, $q_2q_1$ denotes the Hamilton product, $\circ$ the generic group operation, and
$AB$ the matrix product; each group uses one convention throughout.

\subsection{Phase encoding}

A unit vector $v = (x, y, z)$ is encoded by a phase pair $(\alpha, \beta) \in S^1 \times [0, \pi]$:
$\alpha = \mathrm{atan2}(y, x)$: azimuth; $\beta = \arccos(z)$, polar angle. The encoding is
degenerate at the poles ($\beta \in \{0, \pi\}$: $\alpha$ undefined), which is exactly why
extrapolation to the poles is a stress test (Section~\ref{sec:extrap}), not a routine regime.

\subsection{Increments}

The edge output before the node is the \emph{increment}: a Lie-algebra element, written $\xi$,
in the compact/exp case (a vector $v\in\mathbb{R}^3$); in the no-exp variants it is the group
element itself.
Throughout, $\xi$ and $v$ denote the same object in their respective sections.

\subsection{Quaternions}

$q = (w, x, y, z) \in S^3 \subset \mathbb{H}$, scalar-first convention. Hamilton product (juxtaposition, as usual):
$q_1q_2 = (w_1 w_2 - x_1 x_2 - y_1 y_2 - z_1 z_2,\; w_1 x_2 + x_1 w_2 + y_1 z_2 - z_1 y_2,\;
w_1 y_2 - x_1 z_2 + y_1 w_2 + z_1 x_2,\; w_1 z_2 + x_1 y_2 - y_1 x_2 + z_1 w_2)$.
Vector rotation: $v \mapsto q(0, v)q^*$. Axis-angle: $q = (\cos(\theta/2),\,
a \sin(\theta/2))$. Double cover: $q$ and $-q$ are the same rotation; the loss is defined on the
rotated vector and is therefore sign-invariant.

\subsection{Composition order}

Data are generated as $q_2q_1$ (and $q_3q_2q_1$ for $n{=}3$); the network
computes the product in the same order: the node multiplies groups from last to first.

\subsection{Parameter accounting}

Convention and the LawKAN-SU(2) count are in Section~\ref{sec:method}. Extras: a quaternion weight counts
as 4 scalars and an exp-edge output as 3 degrees of freedom (conservative for LawKAN-SU(2));
opponent counts are standard per-layer (all linear layers including biases).

\subsection{Model components}

\begin{table}[!htbp]
\centering
\small
\setlength{\tabcolsep}{5pt}
\caption{Edge types studied here: the feature map into the node's target.}
\label{tab:edges}
\begin{tabular}{@{}>{\raggedright\arraybackslash}p{0.13\columnwidth}>{\raggedright\arraybackslash}p{0.83\columnwidth}@{}}
\toprule
edge & Form \\
\midrule
\emph{Fourier} & order $K$; dense or pruned to top-$k$ \\
\emph{MLP} & width $h$ \\
\emph{per-joint} & one constant $\mathrm{SE}(3)$ screw per joint, in screw coordinates \\
\emph{state} & sees the evolving state (the ODE/dynamics task; Section~\ref{app:probes}) \\
\emph{analytic} & a fixed, non-learned support \\
\bottomrule
\end{tabular}
\end{table}

\begin{table}[!htbp]
\centering
\small
\setlength{\tabcolsep}{3pt}
\caption{Node variants studied here. The relaxed node adds the learned correction $\psi$, whose head is
a global readout, a local kernel, or their two-slot sum.}
\label{tab:nodes}
\begin{tabular}{@{}>{\raggedright\arraybackslash}p{0.10\columnwidth}>{\raggedright\arraybackslash}p{0.13\columnwidth}>{\raggedright\arraybackslash}p{0.73\columnwidth}@{}}
\toprule
node & $\psi$ head & Form / Use \\
\midrule
\emph{exact} & --- & the known group law exactly (the default) \\
\emph{relaxed} & global & a linear readout on the pair features (Fourier on the arm; a small MLP on
the socket/range protocols); smooth, data-rich deformations \\
\emph{relaxed} & local kernel & an RBF kernel over the pair features; localized deformations and the
data-starved corner; the collapse-safe choice \\
\emph{relaxed} & two-slot & their sum; the default \\
\bottomrule
\end{tabular}
\end{table}

\noindent\textbf{Ledger} (compact instance, $n{=}2$, $K{=}1$): $54 = 2\times3\times9$ edge
scalars; node and readout $0$; the analytic edge keeps $18$.

\subsection{FLOP accounting}
\label{app:flops}

Convention in Section~\ref{sec:method}; extra detail: a linear layer $i \to o$ counts as $2io$ (multiply +
add). LawKAN-SU(2) ($K{=}1$, $n{=}2$) breakdown: \textbf{228 FLOPs/example} = 26 (Fourier features) + 102 (einsum
over groups) + 30 (exp-maps) + 28 (Hamilton product) + 30 (probe rotation, via the unit-quaternion
identity $v + 2w(\mathbf{u}{\times}v) + 2\mathbf{u}{\times}(\mathbf{u}{\times}v)$) + 12 (output head). A MACs-only
alternative is reported alongside (Section~\ref{sec:accounting}): LawKAN-SU(2) 102 MACs vs 480--1472 for the
standard heads' linear layers.

\subsection{Error metric}

The core matrix uses the geodesic angle on $S^2$ (Section~\ref{sec:method}): the output is the direction of
the rotated probe, not a full element of $\mathrm{SO}(3)$, so those numbers carry one fewer
orientational degree of freedom and are \textbf{not directly comparable} to full-rotation benchmarks
(e.g.\ the ModelNet/SO(3) tasks in \citet{zhou2019continuity}). The probe readout is the minimal task in
which the group structure shows up; the variable-angle experiment (E29) and the full-rotation opponent
matrix (E30) use the standard full-$\mathrm{SO}(3)$ geodesic instead, with the ordering unchanged
(Section~\ref{sec:main}).

\section{Exact representability of composition at $K{=}1$}
\label{app:exact}

The headline $K{=}1$ result is not a numerical coincidence: at this order the architecture
\emph{contains} the exact composer as a single point in weight space, and no lower order can.
Throughout, the rotation angles $\theta_1,\dots,\theta_n$ are fixed constants (true of the headline
fixed-angle battery E1--E14) and only the axes vary. Write $w_i := \theta_i/2$,
\begin{equation}
a_i := (\sin\beta_i\cos\alpha_i,\ \sin\beta_i\sin\alpha_i,\ \cos\beta_i), \qquad
q_i := \exp(w_i a_i) = (\cos w_i,\ a_i \sin w_i), \qquad Q := q_n\cdots q_1
\end{equation}
(Hamilton product in data order, Section~\ref{app:notation}); the network's answer is the direction
$\pi(\widehat Q)$, defined by the rotated probe vector $\widehat Q\,\hat e_z$.

\textbf{Non-degeneracy:} $\exists j:\ \theta_j \notin 2\pi\mathbb{Z}$; otherwise the composition
map reduces to the identity operator on the probe vector, rendering the representability question
vacuous.

Lemma~\ref{lem:kan-exact} is stated in Section~\ref{sec:construct}; here is its full proof.

\begin{proof}
At $K{=}1$ the nine features $\{1,\cos\alpha,\sin\alpha\}\otimes\{1,\cos\beta,\sin\beta\}$ include
the three axis components as single features: $a_i=(\cos\alpha\sin\beta,\ \sin\alpha\sin\beta,\
\cos\beta)$. Setting the only nonzero edge weights to $\theta_i/2$ on those three (zero on the other
$24$ scalars per edge) gives $v_i=(\theta_i/2)\,a_i=w_i a_i$, whose exponential is exactly $q_i$; the
node composes by the Hamilton product, $\widehat Q=q_n\cdots q_1=Q$.
\end{proof}

\begin{corollary}[poles are not special]
\label{cor:poles}
At $\beta_i \in \{0,\pi\}$ the features $f_6, f_9$ vanish and $f_2 = \cos\beta_i = \pm 1$; the
construction yields $v_i = (\theta_i/2)(0,0,\pm1) = w_i a_i$ exactly, although the azimuth
$\alpha_i$ is undefined there. The \emph{readout} is global by construction even though the
\emph{input encoding} is singular at the poles; this is why the pole-extrapolation stress test
(Section~\ref{sec:extrap}) does not break the architecture.
\end{corollary}

\begin{lemma}[$K{=}1$ is minimal]
\label{lem:kan-minimal}
Under non-degeneracy, no $K{=}0$ network can represent the composition exactly: at $K{=}0$ the
edge output is a constant, while the true output is not. $K{=}1$ is the minimal Fourier order.
\end{lemma}

\begin{proof}
At $K{=}0$ the edge has the single feature $f = 1$, so $v_i$ (and therefore
$\pi(\widehat Q)$) is independent of the input. We show $\pi(Q)$ is not. Choose $j$ with
$\theta_j \notin 2\pi\mathbb{Z}$, fix all axes except $a_j$, and let $u$ be the probe vector
after the first $j{-}1$ rotations; $u \neq 0$ does not depend on $a_j$. Two admissible axis
choices:
\begin{itemize}
\item $a_j = u$: a rotation about its own axis leaves $u$ fixed, so $R(q_j)u = u$;
\item $a_j \perp u$: the only fixed directions of a rotation by $\theta_j \notin
2\pi\mathbb{Z}$ are $\pm a_j$, and $u \neq \pm a_j$, so $R(q_j)u \neq u$.
\end{itemize}
The remaining rotations $q_{j+1},\dots,q_n$ are fixed and invertible, hence the two final vectors
$Q\hat e_z$ differ. The true output takes two distinct values; a $K{=}0$ output is one constant.
\end{proof}

\begin{remark}[the exact weights are not unique]
\label{rem:nonunique}
The output is read through the rotated vector, so the antipodal symmetry of the quaternion
representation is invisible to the readout:
$v_i = (\theta_i/2 + \pi m_i)a_i$, $m_i \in \mathbb{Z}$, are all exact solutions (even $m_i$:
$2\pi$-periodicity of $\exp$ along a fixed axis; odd $m_i$: $\hat q_i = -q_i$, the same rotation).
Consequently, convergence of the \emph{metrics} to the Bayes floor does not by itself imply
convergence of the \emph{weights} to the constructed point. Lemma~\ref{lem:kan-exact} is a
statement about the function class, not about optimization.
\end{remark}

\begin{remark}[relation to the Bayes floor and to noise]
With the constructed weights the approximation error is identically zero, so on noiseless data the
output equals the truth for every input. With noise on the \emph{inputs}, the network's response
is exactly the sensitivity of the true composition map: the architecture adds no error of its own,
and the residual is the propagated input-noise floor, not a representational artifact; the
remaining $0.05\degunit$ is the gap between the exact composer and the Bayes-optimal denoiser.
This is the mechanism behind LawKAN-SU(2) lying $0.05\degunit$ above the floor while the strongest
standard opponent is $0.63\degunit$ above it (Section~\ref{sec:main}).
\end{remark}

\begin{remark}[$K \ge 2$ and the E13 ablation]
The $K{=}2$ feature set is a superset of the $K{=}1$ set, so Lemma~\ref{lem:kan-exact} applies
verbatim and the representation error stays exactly zero; the measured $K{=}2$ regression
($2.87\pm0.03$ vs $2.79\pm0.01\degunit$ on $10$ paired seeds, paired $p{=}1.7\times10^{-5}$;
Appendix~\ref{app:tables}) is an optimization effect (extra harmonics fit noise), not a
representational one. The function-class result
(Lemma~\ref{lem:kan-exact}) and the optimization behavior (E13) are thus cleanly separated.
\end{remark}

\begin{remark}[phase pairing is necessary; formalizes P4]
\label{rem:pairing}
The construction uses \emph{products} of functions of $\alpha$ and of $\beta$ ($f_6 =
\cos\alpha\sin\beta$, etc.); it cannot be carried out by phase-split edges
$v_i = g(\alpha_i) + h(\beta_i)$. Indeed, $\sin\beta\cos\alpha$ would have to be such a sum, but
its mixed derivative $\partial_\alpha\partial_\beta(\sin\beta\cos\alpha) = -\cos\beta\sin\alpha$
is not identically zero, while every sum $g(\alpha) + h(\beta)$ has identically zero mixed
derivative. Paired phase edges are therefore a representational \emph{necessity} for exact
composition: the measured ``phase pairing'' ablation (P4, Section~\ref{sec:ablation}) is not a tuning
choice.
\end{remark}

\begin{remark}[angles that vary per example]
\label{rem:var-angles}
Lemma~\ref{lem:kan-exact} assumes fixed $\theta_i$, true of the headline battery (E1--E14).
If $\theta_i$ varies per example and is supplied as an input, a \emph{linear}
$\theta_i$-feature is not enough (the coefficient of $\theta_i$ in the target depends on
$(\alpha_i,\beta_i)$, which a constant coefficient cannot match); exactness is restored by the
tensor-product extension
\begin{equation}
\{1,\theta_i\}\otimes\{1,\cos\alpha,\sin\alpha\}\otimes\{1,\cos\beta,\sin\beta\}
\quad\text{--- }18\text{ features per edge},
\end{equation}
with weight $\tfrac12$ on the three $\theta_i$-multiplied axis features and zeros elsewhere. If
varying $\theta_i$ are \emph{not} given as input, the target is not a function of
$(\alpha_i,\beta_i)$ at all and the task is stochastic, exactly the P16b regime of
Section~\ref{sec:scope}, in which every architecture degrades by an order of magnitude.
\end{remark}

\subsection{Proof of Proposition~\ref{prop:span} and the floor}
\label{app:spanproof}

\emph{Hypotheses.} (i) the chart $\chi:U\to G$ is injective on the input domain; (ii) the edge is
linear in the features $F$ with weights $W$; (iii) the node exponentiates and applies the exact law of
$G$; (iv) $e\in U$, and $\chi(U)$ generates $G$ (products of chart elements cover the tested
compositions).

\emph{Constructive direction.} If $\log\chi$ lies in the row space of $F$, choose $W$ as the coefficients of
$\log\chi$ in the features; the edge then emits $\log\chi(x)$ on the domain, the node returns
$\chi(x)$, and every finite product of exact factors is exact. Lemma~\ref{lem:kan-exact} makes this
explicit for $\mathrm{SU}(2)$, and Section~\ref{app:support} for $GL^{+}(2,\mathbb{R})$ and $\mathrm{SE}(3)$.

\emph{Converse.} The chart-spanning principle (Appendix~\ref{app:crossgroup}) proves the converse: if the
architecture is exact on $U\times U$, then $h=\chi\,c^{-1}$ with $c$ a central involution;
substituting forces $c^{2}=e$, so up to the double cover $h\equiv\chi$, i.e.\
$\log\chi$ lies in the row space of $F$.

\emph{Floor.} If $\log\chi$ does not lie in the row space of $F$, write $\delta(x)$ for the residual of the
projection of $\log\chi(x)$ onto the span. The node receives $\log\chi-\delta$; since the exponential
is a retraction at the identity (Appendix~\ref{app:retraction}), the induced error is first order in
$\Vert\delta(x)\Vert$ and strictly positive whenever $\delta\ne0$, so the mean error has a floor independent of
training. The registered blind tests exhibit it: removing feature groups (E23) and the $K{=}1$
harmonic floor (E24, Appendix~\ref{app:mech}).

\subsection{Analytic support for the three instances}
\label{app:support}

The constructive direction written out for the instances.

\emph{$\mathrm{SU}(2)$.} With axis $a(\alpha,\beta)=(\sin\beta\cos\alpha,\ \sin\beta\sin\alpha,\
\cos\beta)$ and fixed $\theta$, the increment is $\log q=(\theta/2)\,a$: three of the nine $K{=}1$
features (Lemma~\ref{lem:kan-exact}); constructed weights exact to $7.6\times10^{-7}$ degrees.

\emph{$GL^{+}(2,\mathbb{R})$.} For $g=R(\theta)D(s_1,s_2)H(t)$ in Iwasawa coordinates, the entries are
$g_{11}=e^{s_1}c$, $g_{21}=e^{s_1}s$, $g_{12}=t e^{s_1}c-e^{s_2}s$, $g_{22}=t e^{s_1}s+e^{s_2}c$
($c,s\equiv\cos,\sin\theta$); each entry is a linear combination of the six features
$\{e^{s_1}c,\ e^{s_1}s,\ e^{s_2}c,\ e^{s_2}s,\ t e^{s_1}c,\ t e^{s_1}s\}$; constructed weights exact to
$1.1\times10^{-7}$ (relative Frobenius).

\emph{$\mathrm{SE}(3)$.} For $g=(R(a,\theta),p)$ the logarithm is $\xi=(\omega,v)$ with
$\omega=\theta a$ and $v=V^{-1}(\omega)\,p$, where
$V^{-1}=I-\tfrac12\omega^\wedge+d(\theta)(\omega^\wedge)^2$ and
$d(\theta)=\theta^{-2}-(1+\cos\theta)/(2\theta\sin\theta)$. Expanding,
\begin{equation}
v=(1-d\theta^2)\,p-(\theta/2)\,(a\times p)+d\theta^2\,a(a\cdot p),
\end{equation}
so the increment is linear in $\{a,\ p,\ a\times p,\ a(a\cdot p)\}$; constructed weights exact to
$1.6\times10^{-7}$.

\subsection{Truncated BCH is never exact}
\label{app:bch}
For non-commuting increments the logarithm of a product is the Baker--Campbell--Hausdorff series
\begin{equation}
\log(e^{v_1}e^{v_2})=v_1+v_2+\tfrac12[v_1,v_2]
+\tfrac1{12}\bigl([v_1,[v_1,v_2]]+[v_2,[v_2,v_1]]\bigr)+\cdots,
\end{equation}
which does not terminate in general; a finite truncation therefore leaves a positive floor for generic
inputs (E18), while the exact group law is a finite operation (the Hamilton product costs $28$ scalar
operations). Matched-parameter truncations thus trade exactness for no savings, which is why the node
uses the exact law.

\section{Extended per-seed tables and cross-group proof}
\label{app:tables}

\noindent The three-instance summary is in the main text (Table~\ref{tab:instances}); the per-seed
numbers follow.

\begin{figure}[t]
\centering
\caption{Cross-group characterization (log scale) for the two non-compact instances: the analytic
edge against the dense/linear, no-exp and MLP alternatives; the analytic edge
is orders of magnitude below all of them. Each group is evaluated in its own metric
($GL^{+}(2,\mathbb{R})$: relative Frobenius; $\mathrm{SE}(3)$: per-coordinate probe), so heights are
comparable within a group, not across the two. ($\mathrm{SU}(2)$ is exact at $K{=}1$
(Lemma~\ref{lem:kan-exact}) and is therefore not part of this log-scale comparison.)}
\label{fig:crossgroup}
\includegraphics[width=0.615\columnwidth]{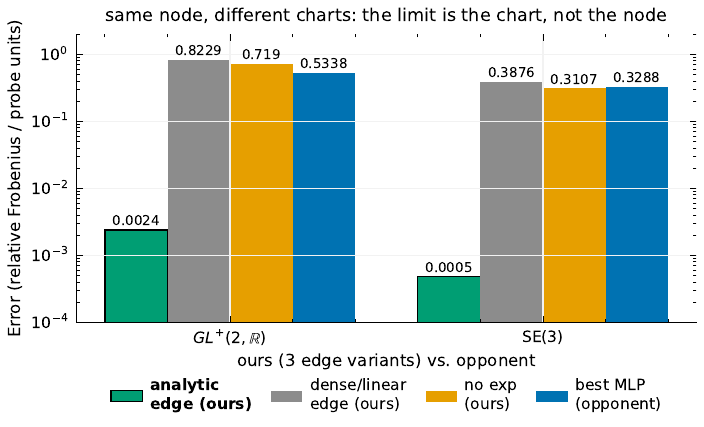}

\end{figure}

\subsection{The non-compact instance: $GL^{+}(2,\mathbb{R})$}
\label{sec:gl2}

$GL^{+}(2,\mathbb{R})$ (orientation-preserving plane deformations, in Iwasawa coordinates) is
composed by matrix product. Here the exponential is \emph{not} global, and the compact-instance
recipe (edge $\to\mathfrak{g}\to\exp$) does not transfer: sending the product features through a
further exponential \emph{hurts}. The LawKAN-GL$^{+}$ model, using the exact node law with the six
analytic Iwasawa features
$\{e^{s_1}c,\ e^{s_1}s,\ e^{s_2}c,\ e^{s_2}s,\ t e^{s_1}c,\ t e^{s_1}s\}$
($c,s\equiv\cos,\sin\theta$), achieves a relative Frobenius error of $0.0024$ and beats an MLP
with $100\times$ the parameters ($0.53$); a dense edge over raw coordinates does not beat the MLP. The
limit is the chart and the feature span (Table~\ref{tab:gl2}, Section~\ref{app:crossgroup}).

\subsection{The mixed instance: $\mathrm{SE}(3)$}
\label{sec:se3}

$\mathrm{SE}(3)$, the rigid motions of space (a rotation with a translation; robot poses), is
non-compact but has a surjective exponential (screw theorem), so the full design (edge $\to$ Lie
algebra $\to$ exp $\to$ product) can be exact. On two composed rigid transforms the analytic edge
edge is near-exact (constructed weights exact to $1.6\times10^{-7}$; trained edge at
$4.8\times10^{-4}$ relative Frobenius) and orders of magnitude below
a dense linear edge and an MLP $h{=}64$ (paired $p{=}3\mathrm{e}{-8}$). A blind feature-group test
(held-out seeds) confirms both non-linear groups are necessary and that the floor scales linearly
in the translation noise (Table~\ref{tab:instances}; per-seed numbers in Section~\ref{app:crossgroup}). One exact node, different charts: the limit is the chart and the
feature span.

Full per-seed and extended tables are regenerated
mechanically from the released JSON dumps (\texttt{repro/}, Appendix~\ref{app:repro}); the compact
summaries are in the text, and the JSON dumps are the mechanical source of every number. Below we
keep the additional opponent/robustness checks and the cross-group proof with its per-seed
results.

\subsection{Additional opponent and robustness checks (E18, X2, X4)}
\label{app:extra}

\textbf{BCH ladder (E18).} On the E1 task ($5$ seeds), replacing the Hamilton node by a truncated
Baker--Campbell--Hausdorff series leaves a floor: exact Hamilton $2.78\pm0.02\degunit$, zeroth-order sum
$5.91\pm0.06\degunit$, $+\frac12$ commutator $4.00\pm0.04\degunit$, $+$ triple commutators
$3.88\pm0.06\degunit$ ($p\le4.5\mathrm{e}{-6}$ against Hamilton); at order $2$ the truncation already
costs no fewer scalar operations than the exact product. The flop-lean rotation identity agrees with the
quaternion sandwich to $9.5\mathrm{e}{-7}$.

\textbf{Non-Gaussian noise (X2).} With uniform and Laplace input/target noise at matched root-mean-square (RMS) noise, the
ordering is unchanged at every $\sigma\in\{0.05,0.10,0.20\}$: e.g.\ Laplace $\sigma{=}0.10$:
LawKAN-SU(2) $8.90\pm0.05\degunit$ vs MLP-Quat $h{=}32$ $10.20\pm0.15\degunit$
(paired $p{=}5.7\mathrm{e}{-5}$); uniform $\sigma{=}0.10$: $9.36\pm0.03$ vs $10.59\pm0.22\degunit$
(paired $p{=}4.9\mathrm{e}{-4}$). RPMG does not change the ordering.

\textbf{Vector Neurons (X4).} A minimal $\mathrm{SO}(3)$-equivariant VN-MLP ($h{=}8$) reaches
$12.12\pm1.28\degunit$ against LawKAN-SU(2) $2.78\pm0.02\degunit$ ($p{=}1.3\mathrm{e}{-4}$) and
MLP-Quat$\,+$RPMG $3.42\pm0.15\degunit$: equivariant transport of the input axes is not a substitute
for a composition law.

\subsection{Cross-group characterization: proof and per-seed results}
\label{app:crossgroup}

\textbf{Principle (chart spanning).} Let $\chi:U\to G$ be a chart, $h(x)=\exp(WF(x))$, and suppose
the architecture is exact for two-fold composition on $U\times U$:
$h(x_2)h(x_1)=\chi(x_2)\chi(x_1)$ for all $x_1,x_2\in U$. With $s(x_1):=h(x_1)\chi(x_1)^{-1}$ this
gives $h(x_2)s(x_1)=\chi(x_2)$; the left side depends on $x_1$ only through $s$, the right only on
$x_2$, so both equal a constant $c$, i.e.\ $h(x)=\chi(x)c^{-1}$. Substituting back gives
$c^{-1}\chi(x_1)c^{-1}=\chi(x_1)$ for all $x_1\in U$; at the identity this forces $c^2=e$, so $c$ is a
central involution ($c=e$, or $c=-I$ in the double-cover case). Hence $h\equiv\chi$ up to a central
involution, and since the readout and every reported metric are invariant under the
$\mathrm{SU}(2)\to\mathrm{SO}(3)$ cover (Remark~\ref{rem:doublecover}), exactness on the product set
forces $\log\chi$ to lie in the row space of $F$ up to this double cover, and conversely. For the no-exp variants, replace $\log\chi$ by $\chi$.

\begin{table}[H]
\centering
\small
\caption{Transfer to the non-compact group $GL^{+}(2,\mathbb{R})$ (v2): relative
Frobenius error, five seeds (per-seed values rounded to three decimals). \emph{Structured}
carries the six analytic Iwasawa features; the retracted v1 run is reported below.}
\label{tab:gl2}
\setlength{\tabcolsep}{4pt}
\begin{tabular}{@{}>{\raggedright\arraybackslash}p{0.28\columnwidth}rr>{\raggedright\arraybackslash}p{0.30\columnwidth}@{}}
\toprule
Model & Params & Rel.\ Frobenius & Per-seed \\
\midrule
LawKAN-GL$^{+}$ (exp, all inputs) & 48 & 0.823 $\pm$ 0.005 & [0.818, 0.828, 0.819, 0.829, 0.820] \\
LawKAN-GL$^{+}$ (no exp: edge emits the group element) & 48 & 0.719 $\pm$ 0.004 & [0.723, 0.718, 0.714, 0.723, 0.717] \\
\textbf{LawKAN-GL$^{+}$ (analytic edge)} & \textbf{48} & \textbf{0.0024 $\pm$ 0.0005} & [0.003, 0.002, 0.002, 0.003, 0.003] \\
MLP $h{=}3$ (matched) & 55 & 0.806 $\pm$ 0.094 & [0.729, 0.686, 0.945, 0.873, 0.796] \\
MLP $h{=}64$ & 4996 & 0.534 $\pm$ 0.016 & [0.543, 0.504, 0.553, 0.534, 0.535] \\
\bottomrule
\end{tabular}
\end{table}

\medskip
\noindent\textbf{E22/E23 (SE(3), full design and feature ablation).} Numbers are in
Section~\ref{sec:crossgroup}; per-seed values and the blind-test details are in the prediction log
(E22, E23; Appendix~\ref{app:predictions}).

\medskip
\noindent\textbf{E16 v1 (retracted).} The early $GL^{+}(2,\mathbb{R})$ run used only $2$ of the $4$
Iwasawa inputs (relative Frobenius $0.900\pm0.014$, $72$ params); it is superseded by the full-input
runs of Table~\ref{tab:gl2} and kept here only for provenance.

\medskip
\noindent\textbf{E21 (analytic edge is discoverable).} Starting from a generic $24$-feature
dictionary and pruning to the top-$m$ features with retraining selects exactly the six analytic
Iwasawa features (top-$6$ $=$ analytic on the recorded seed) and reaches the exact level at $m{=}6$, beating the unpruned
dictionary (replication seeds $10$--$14$):
\begin{center}\small
\begin{tabular}{lcccccc}
\toprule
features kept $m$ & $2$ & $4$ & $5$ & $6$ & $12$ & $24$ (dense) \\
\midrule
error & $0.81$ & $0.48$ & $0.29$ & \best{0.0026} & $0.0045$ & $0.0069$ \\
\bottomrule
\end{tabular}
\end{center}

\section{Compact-instance details: full SO(3) readout and plots}
\label{app:compact}

\begin{table}[H]
\centering
\small
\caption{Full SO(3) readout (E30): geodesic error (deg), mean $\pm$ std; target-noise-only
bound $5.48\degunit$, Monte-Carlo floor with input noise $6.05\degunit$ (E36). Paired vs
LawKAN-SU(2): Lie head and MLP-Quat $p{=}4.8\mathrm{e}{-5}$. The 6D and 9D-SVD rows use
the corrected Shepperd conversion; they are tied within noise, and the LawKAN/Lie/MLP
ordering is unchanged.}
\label{tab:e30}
\begin{tabular}{lr}
\toprule
Model & Error (deg) \\
\midrule
\textbf{LawKAN-SU(2)} & \textbf{6.12 $\pm$ 0.05} \\
Lie head (strongest opponent) & 6.86 $\pm$ 0.10 \\
MLP-Quat $h{=}32$ ($+$RPMG) & 7.05 $\pm$ 0.08 \\
MLP-6D $h{=}32$ & 7.73 $\pm$ 0.17 \\
MLP-9D-SVD $h{=}16$ & 7.73 $\pm$ 0.08 \\
\bottomrule
\end{tabular}
\end{table}

\begin{figure}[H]
\centering
\caption{Capacity curves under the fixed $\exp{+}$Hamilton node (E31/E31b/E31c,
E13): geodesic error against parameters for MLP and Fourier edges sharing the node, with the no-node
control band (grey; the full opponent matrix is Figure~\ref{fig:pareto}) and the Bayes floor
(dashed). The $54$-parameter Fourier edge matches a
$278$-parameter MLP edge ($228$ vs $588$ forward FLOPs/example, Appendix~\ref{app:flops}); points are
mean $\pm$ std over seeds.}
\label{fig:nodecapacity}
\includegraphics[width=0.675\columnwidth]{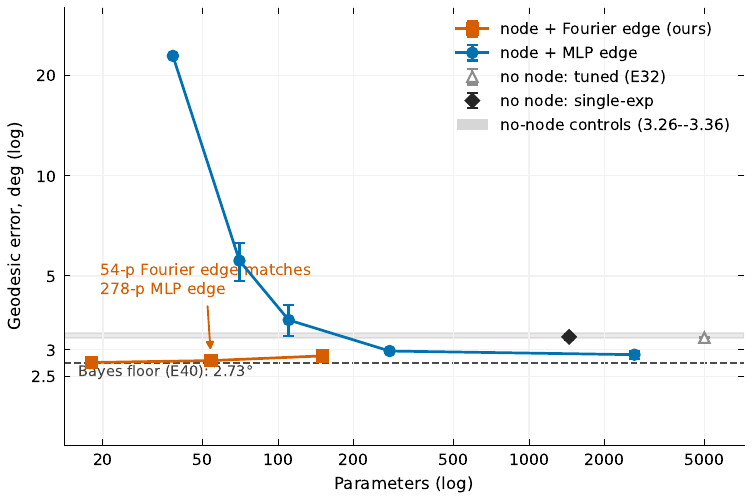}

\end{figure}

\begin{figure}[H]
\centering
\caption{Sample-efficiency curves (E4): LawKAN-SU(2)@$N{=}50$ $\approx$ 6D@$N{=}1600$ (within noise);
against the fixed-protocol opponent structure saves $\approx32\times$ data, against the strongest
tuned opponent $\approx7\times$ (E39).}
\label{fig:sample}
\includegraphics[width=0.54\columnwidth]{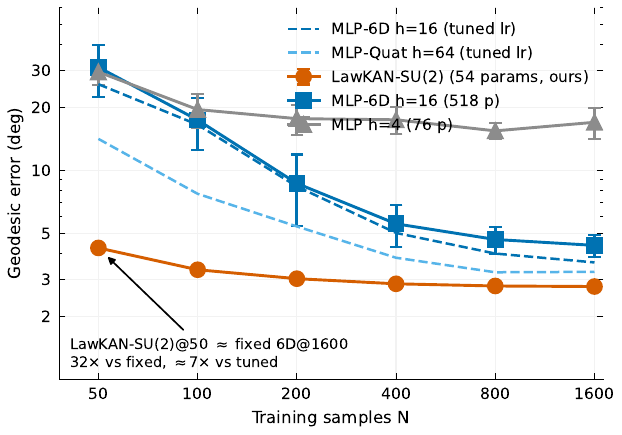}

\end{figure}

\begin{table}[t]
\centering
\small
\caption{Full opponent matrix: geodesic error (deg), mean $\pm$ std,
input/output noise $\sigma{=}0.03$, target-noise-only floor $\approx1.9\degunit$; the 9D-SVD family is
shown to $h{=}16$; the tuned rows are at each opponent's best learning rate (E38).}
\label{tab:matrix-full}
\begin{tabular}{lrr}
\toprule
Model & Params & Error (deg) \\
\midrule
\textbf{LawKAN-SU(2), edge: Fourier ($K{=}1$), node: exact} & \textbf{54} & \textbf{2.78 $\pm$ 0.02} (5 seeds) \\
\midrule
\multicolumn{3}{@{}l}{\emph{Opponents}} \\
MLP + quaternion head $h{=}32$ & 1476 & 3.36 $\pm$ 0.06 (5 seeds) \\
MLP + quaternion head $h{=}32$ + RPMG \citep{chen2022rpmg} & 1476 & 3.42 $\pm$ 0.15 (5 seeds) \\
MLP + quaternion head $h{=}16$ & 484 & 3.77 $\pm$ 0.07 \\
MLP + 6D $h{=}32$ \citep{zhou2019continuity} & 1542 & 3.91 $\pm$ 0.11 \\
MLP + 6D $h{=}16$ \citep{zhou2019continuity} & 518 & 4.49 $\pm$ 0.65 (5 seeds) \\
MLP + 9D-SVD $h{=}16$ \citep{levinson2020svd} & 569 & 4.32 $\pm$ 0.46 \\
RealKAN + 6D head $h{=}16$ & 550 & 4.90 $\pm$ 0.26 \\
MLP + quaternion head $h{=}8$ & 180 & 5.84 $\pm$ 0.24 \\
MLP + 9D-SVD $h{=}8$ & 225 & 7.11 $\pm$ 0.34 \\
MLP + 6D $h{=}8$ \citep{zhou2019continuity} & 198 & 7.56 $\pm$ 0.27 \\
RealKAN + 6D head $h{=}8$ & 278 & 7.83 $\pm$ 0.36 \\
MLP + quaternion head $h{=}4$ & 76 & 11.08 $\pm$ 0.30 \\
MLP + 9D-SVD $h{=}4$ & 101 & 12.06 $\pm$ 0.53 \\
MLP + 6D $h{=}4$ \citep{zhou2019continuity} & 86 & 13.81 $\pm$ 0.93 \\
MLP $h{=}4$ (atan2 head) & 76 & 15.14 $\pm$ 1.44 (5 seeds) \\
RealKAN 2-layer K=3 & 168 & 19.97 $\pm$ 1.16 \\
RealKAN K=3 & 56 & 32.50 $\pm$ 0.22 \\
\midrule
\multicolumn{3}{@{}l}{\emph{Tuned learning rate (E38)}} \\
MLP + quaternion head $h{=}64$, tuned & 4996 & 3.25 $\pm$ 0.02 (5 seeds) \\
MLP + 6D $h{=}16$, tuned & 518 & 4.00 $\pm$ 0.54 (5 seeds) \\
MLP $h{=}4$ (atan2), tuned & 76 & 14.41 $\pm$ 1.05 (5 seeds) \\
\bottomrule
\end{tabular}
\end{table}

\begin{table}[H]
\centering
\small
\caption{Synthetic IMU probe: geodesic error ($\degunit$), $20$-step
trajectories; lower rows: low-data run ($50$ trajectories).}
\label{tab:imu-syn}
\begin{tabular}{lrr}
\toprule
Model & Params & Error, $\degunit$ \\
\midrule
LawKAN-SU(2) & $27$ & \best{0.082 \pm 0.000} \\
MLP $h{=}8$ & $596$ & $18.1 \pm 1.0$ \\
MLP $h{=}32$ & $3{,}140$ & $9.15 \pm 0.11$ \\
MLP-Quat $h{=}32$ $+$ RPMG & $3{,}140$ & $9.21 \pm 0.07$ \\
\emph{low-data:} LawKAN-SU(2) & $27$ & $0.089 \pm 0.006$ \\
\emph{low-data:} MLP $h{=}32$ & $3{,}140$ & $42.6 \pm 12.3$ \\
\bottomrule
\end{tabular}
\end{table}

\begin{table}[t]
\centering
\small
\caption{Bayes floor (the reported-metric minimizer gives the same floor)
and the gap to it; unrounded $n{=}2$ floor $2.733\degunit$ (matched to the E1 test draws, E40) (Appendix~\ref{app:tables}).}
\label{tab:bayes-main}
\begin{tabular}{lrrr}
\toprule
Task & Bayes floor & LawKAN-SU(2) & Top opponent \\
\midrule
$n{=}2$ (fixed angles) & $2.73\degunit$ & $2.78\degunit$ ($+0.05\degunit$) & $3.36\degunit$ ($+0.63\degunit$) \\
$n{=}3$ (fixed angles) & $3.06\degunit$ & $3.19\degunit$ ($+0.13\degunit$) & $6.60\degunit$ ($+3.54\degunit$) \\
$n{=}2$ random & $37.7\degunit$ & $38.21\degunit$ ($+0.5\degunit$) & $46.48\degunit$ ($+8.8\degunit$) \\
\bottomrule
\end{tabular}
\end{table}

\subsection{Battery, sample efficiency and accounting}
\label{sec:accounting}
\label{sec:battery}
\label{sec:sample}

\noindent The margin holds across the battery (Appendix~\ref{app:tables}): at $n{=}2$, at $n{=}3$ and
at the poles LawKAN-SU(2) leads the strongest head by a factor that grows with depth and
extrapolation, and RPMG~\citep{chen2022rpmg} on the same head is statistically indistinguishable from
it and still loses. Sample efficiency: LawKAN-SU(2) at $N{=}50$ matches the fixed-protocol 6D
baseline at $N{=}1600$ within noise (about $32\times$ fewer samples at matched error; against the
strongest tuned opponent the ratio is about $7\times$, E39), with no crossover up to $N{=}3200$ and
no advantage beyond the plateau. Accounting: under one convention
(Appendix~\ref{app:flops}) LawKAN-SU(2) needs $228$ FLOPs/example against $2848$ for the strongest
head --- $12\times$ fewer FLOPs and $27\times$ fewer parameters at matched accuracy ($92\times$ against the
strongest tuned head) --- at a raw cost
$\sim1.8\times$ the tiny MLP, the price of built-in algebra; the figure counts every transcendental as
one FLOP, so it \emph{understates} $\exp/\mathrm{atan2}$ and on wall-clock LawKAN-SU(2) is the slower
of the two. The Hamilton node is the flop-cheapest \emph{exact} composer
(Appendix~\ref{app:extra}). Full tables and the sample-efficiency and noise curves are
Figures~\ref{fig:sample} and~\ref{fig:noise}; the full SO(3) readout is
Table~\ref{tab:e30}.

\subsection{Justification notes for the benchmark}
\label{app:main-notes}
\begin{table}[t]
\centering
\small
\caption{Node/edge isolation on the composition task (5 seeds): one fixed
$\exp{+}$Hamilton node, different edges (upper block); no-node controls (lower block). Mean $\pm$ std,
geodesic degrees; bold marks the best row. The $54$-parameter Fourier edge is the default used in the
main text and is within noise of the $18$-parameter analytic edge.}
\label{tab:isolation}
\begin{tabular}{lrr}
\toprule
Edge (same node unless noted) & Params & Error (deg) \\
\midrule
Fourier edge ($K{=}1$) & 54 & $2.78 \pm 0.02$ \\
analytic edge & 18 & \best{2.75 \pm 0.02} \\
MLP edge ($h{=}8$) & 278 & $2.97 \pm 0.04$ \\
MLP edge ($h{=}32$) & 2630 & $2.90 \pm 0.08$ \\
MLP edge ($h{=}32$, best lr) & 2630 & $2.86 \pm 0.03$ \\
\midrule
MLP edge ($h{=}4$, 2 layers) & 110 & $3.69 \pm 0.39$ \\
MLP edge ($h{=}4$, 1 layer) & 70 & $5.56 \pm 0.74$ \\
MLP edge ($h{=}2$, 1 layer) & 38 & $22.9 \pm 0.4$ \\
linear edge & 30 & $13.3 \pm 0.1$ \\
\midrule
no node: plain MLP-Quat $h{=}32$ & 1476 & $3.36 \pm 0.06$ \\
no node: single-exp MLP-Lie $h{=}32$ & 1443 & $3.29 \pm 0.10$ \\
\bottomrule
\end{tabular}
\end{table}

Here ``strongest'' refers to the standard rotation-head matrix, whose strongest row is
MLP-Quat $h{=}32$ ($3.36$); the no-node single-exp MLP-Lie control reaches $3.29$ (Table~\ref{tab:isolation})
but is a different readout class and is reported as a control, not as a matrix opponent.
The gap to the strongest fixed-protocol opponent grows with depth; the Lie-algebra baseline
(MLP $\to\so(3)\to\exp$) does not close it, so the parameterization alone does not explain the win.
Tuning the opponent does not close it either (the best of nine $h\times\mathrm{lr}$ cells stays above
the model; at its own best rate the $h{=}32$ edge moves from $2.90$ to $2.86\degunit$, E38), and
training directly on the reported metric does not move it; the parameter margin should
therefore be read as bounding the whole structured design, not the node alone. Seeds are shared across
rows, so the gap over MLP-Quat $h{=}32$ is significant (disjoint $95\%$ CIs, Appendix~\ref{app:seeds});
the ranking is robust to the opponent's learning rate (a registered three-rate sweep), and enlarging
the Fourier order does not help (a Fourier-order ablation; Table~\ref{tab:sparse-ablation}).

\paragraph{Node/edge isolation details.} With the same
$\exp{+}$Hamilton node, Fourier and MLP edges $h{\ge}8$ beat both no-node controls (paired
$p<0.01$; capacity curves in Figure~\ref{fig:nodecapacity}), and the linear-edge failure is what the
span criterion predicts: $\log\chi$ is not in the span of four raw phase features. The
$18$-parameter analytic edge is the cheapest point; the Fourier edge matches a $278$-parameter MLP
edge at a fraction of its forward FLOPs (Appendix~\ref{app:flops}). Full-$\mathrm{SO}(3)$ readout:
LawKAN-SU(2) stays the most accurate at $27\times$ fewer parameters, now against the Lie head
($6.12$ vs $6.86\degunit$), and is $+0.07\degunit$ above the Monte-Carlo floor with input noise
(strongest opponent $+0.81\degunit$; Table~\ref{tab:e30}). The margin holds across the battery and the stress tests
(Section~\ref{sec:stress-tests}); sample efficiency and the unified FLOP/parameter accounting are in
Section~\ref{sec:accounting}.

\paragraph{IMU details.} LawKAN-SU(2) computes the magnitude through a
linear channel and the direction through the $K{=}1$ Fourier phase pair; the pruned edge reduces to the
analytic edge; the network maps the whole sequence to the total rotation in one pass and does not
iterate the ODE. For the real probe, the dataset is split into $800$ training and $2000$ testing
examples with no overlap between the splits; we evaluate $3$ seeds with the true $\mathrm{SO}(3)$
geodesic (an earlier rotvec-chart metric, discontinuous at $\pi$, is retracted).
Synthetically the shared $27$-parameter edge is at the metric floor while MLPs two orders of
magnitude larger stay far above it (Table~\ref{tab:imu-syn}).

\paragraph{Arm-calibration details.} The per-joint SE(3) correction is
ahead of PoE at every size: paired $p{=}0.019$ at $N{=}20$, $p{=}0.06$--$0.84$ at larger sizes;
it beats the residual MLP at $N{=}20$ ($p{=}0.005$) and $N{=}40$ ($p{=}0.049$), tied at $N{=}60/80$
($N{=}80$: $1.42$ vs $1.45\,$mm, $p{=}0.85$). TOST (margin $10\%$ of the classical baseline,
$\alpha{=}0.05$) certifies parity at $N{\ge}40$: paired differences $-0.007/-0.031/-0.062\,$mm,
$90\%$ CIs inside the margin (TOST $p{=}0.004/0.0003/0.025$).

\subsection{Boundaries: stress tests}
\label{sec:stress-tests}

All stress tests use the \emph{dense} edge ($54$ parameters, $K{=}1$), the architecture-agnostic
default: the pruned edge is a derived construction (valid only where the composition angles are fixed
and known) and cannot relax these boundaries; where it applies it shifts them in our favor
(Section~\ref{sec:pruning}, Section~\ref{sec:imu}).

\begin{figure}[t]
\centering
\caption{(a) Error vs noise $\sigma$: LawKAN-SU(2) dominates the whole curve; the
$2.2\times$ gap at $\sigma{=}0.01$ is the opponent's approximation residual. (b) Pole extrapolation: the standard head degrades $1.9\times\to4.0\times$ against $1.24\times$ for LawKAN-SU(2).}
\label{fig:noise}
\includegraphics[width=0.95\columnwidth]{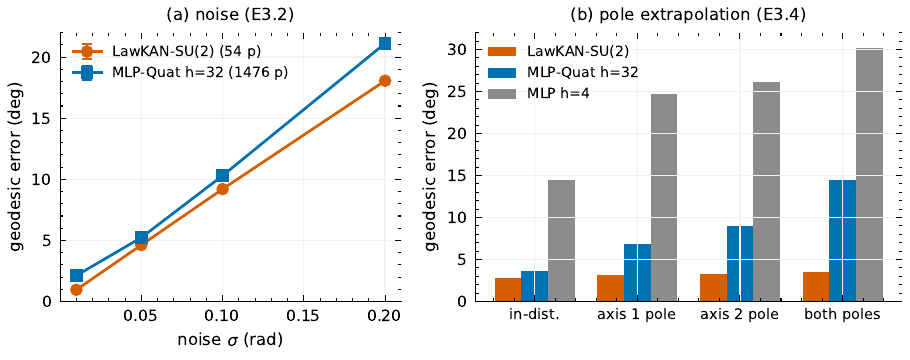}

\end{figure}

\paragraph{Noise, extrapolation and depth.}
\label{sec:noise}
\label{sec:extrap}
\label{sec:n3}
\label{sec:chain}
Across input/target noise LawKAN-SU(2) degrades linearly and dominates the whole curve (the gap at low
noise is the opponent's approximation residual, not noise); out of band the Fourier edge and the group
law are global by construction, so the poles are not special (the standard head degrades
$1.9\times\to4.0\times$ against $1.24\times$ for LawKAN-SU(2); Figure~\ref{fig:noise}). Depth helps
within the learnable regime: at $n{=}3$ the gap over the strongest head is ${\sim}2\times$, and along
chains up to eight factors the advantage widens as the opponent degrades superlinearly, until every
architecture leaves the learnable regime (around $n{=}6$ the compact instance turns seed-bimodal, and
by $n{=}8$ all models are at chance; Appendix~\ref{app:tables}).

\paragraph{Mismatch: the imposed law is not the task's.}
\label{sec:scope}
\label{sec:zn}
With inputs, noise and budget held fixed but a target with \emph{no} group law (a mildly warped sum of
phases), the Hamilton node is a mismatch: $23.87\pm0.41\degunit$ against $12.64\pm3.00$ for an MLP
$h{=}4$ and $3.45\pm0.35$ for MLP-Quat $h{=}32$. With random \emph{hidden} angles the target is not a
function of the input, every architecture faces irreducible uncertainty, and the margin narrows to the
Bayes floor; when the angle \emph{is} supplied (IMU and kinematic probes) varying angles are handled
well. The advantage is therefore structure--task alignment, not a universal parameterization win, and
Table~\ref{tab:applicability} maps the task types. The registered $\mathbb{Z}_n$/BKT analogy did not
appear and remains motivation only.

Two further isolated probes, the synthetic forward-kinematics calibration and the rigid-body ODE with
a state-conditioned edge, are deferred to Appendix~\ref{app:probes}: they close their synthetic
targets (a $4\times$ margin over an MLP, and the numerical floor of the analytic integrator,
respectively) without changing the envelope.

\section{Applied-probe details}
\label{app:applied}

\subsection{IMU orientation (controlled and real BROAD)}
\label{sec:imu}

The first applied block reconstructs orientation from angular velocity ($20$ integration steps),
evaluated on the real BROAD dataset \citep{laidig2021broad,broad2021dataset} (consumer gyroscopes at $286\,$Hz,
structured noise, optical ground truth) and on a controlled synthetic benchmark. The map
$\omega_s\mapsto\exp(\omega_s dt)$ is the same at every step, so one shared edge ($27$ parameters,
pruned to $18$) learns that single step while the exact node composes all $20$ in one pass; baselines
are the training-free strapdown integrator \citep{titterton2004strapdown}, flat MLPs, a GRU and a
residual MLP, and RPMG \citep{chen2022rpmg}. The real probe uses $16$ disturbed trials with a
trial-level split ($3$ seeds, true $\mathrm{SO}(3)$ geodesic); details, the synthetic block and the early retracted metric are below.

\begin{table}[!ht]
\centering
\small
\caption{Real BROAD IMU orientation at three horizons, geodesic error ($\degunit$):
$27$-p dense edge up to $0.07\,$s, $36$-p at longer horizons where a bias feature is added, $18$-p
sparse; the dense edge is seed-bimodal at long horizons; per-seed values are listed and the failing
seed is environment-dependent. The synthetic block is Table~\ref{tab:imu-syn}. $3$ seeds; the classical same-class comparator is the
training-free strapdown. \textsuperscript{\dag}Classical AHRS fusion filter
\citep{laidig2023vqf}; it uses the accelerometer (different input class) and is shown as a scope
reference, not as a gyro-only comparator.}
\label{tab:imu}
\begin{tabular}{lrrr}
\toprule
Model & $0.07\,$s & $1.0\,$s & $5.0\,$s \\
\midrule
VQF 6D\textsuperscript{\dag} fusion (gyro+accel) & $0.201$ & $0.536$ & $0.680$ \\
Strapdown (training-free) & $0.394$ & $1.148$ & $1.191$ \\
LawKAN-SU(2) dense ($27/36$p) & $0.423$ & $1.147$ / $1.089$ / $1.186$ & $1.142$ / $1.185$ / $73.1$ \\
LawKAN-SU(2) sparse ($18$p) & --- & \best{1.057{\pm}0.000} & \best{1.189{\pm}0.008} \\
MLP $h{=}8$ & $0.534$ & $47.2$ & $88.2$ \\
MLP $h{=}32$ & $0.552$ & $58.8$ & $102.2$ \\
GRU $h{=}32$ ($3{,}939$p) & --- & $14.9$ ($28.3$t) & $78.7$ ($83.0$t) \\
residual MLP & --- & $1.55$ & $2.58$ \\
\bottomrule
\end{tabular}
\end{table}

\begin{figure}[!ht]
\centering
\caption{BROAD real-IMU (log scale): the training-free integrator and the pruned $18$-p edge
stay on the manifold; the dense edge is seed-bimodal (dots $=$ seeds; one diverges at $5\,$s, so its
mean is far above the median); the flat MLP leaves the manifold.}
\label{fig:imu}
\includegraphics[width=0.465\columnwidth]{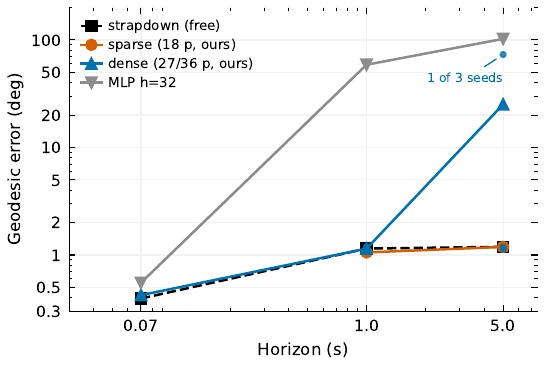}
\end{figure}

On the real sensor the same structure stays on the rotation manifold at every horizon while the
flat MLPs leave it, and the pruned $18$-parameter edge is seed-stable and is the only learned model
to achieve parity with the training-free integrator at $5\,$s; synthetically the shared edge is at the metric
floor while much larger MLPs stay far above it (Table~\ref{tab:imu-syn}). The win
here is not competing with the group law but suppressing structured gyro noise without seed
selection. For scope, the classical AHRS fusion filter VQF~\citep{laidig2023vqf} reaches
$0.20/0.54/0.68\degunit$ on the same windows with the gyroscope \emph{and} accelerometer --- a
different input class, listed as a reference; the like-for-like gyro-only classical bound remains
the training-free integrator.

\subsection{Gyroscope calibration (same sensor)}
\label{sec:gyrocal}

The second block calibrates the angular-velocity sensor itself: fit
$\omega_{\mathrm{true}} = M(\omega_{\mathrm{raw}}-b)$ (scale/misalignment $M$, bias $b$) on a $70/30$
split of the (non-overlapping) windows pooled from the $23$ undisturbed BROAD trials, and evaluate strapdown integration with the corrected
measurements on the held-out $30\%$. The learned $12$-parameter $(M,b)$ \emph{matches} the classical
Levenberg--Marquardt (LM) fit of the same model at every horizon (differences $\le0.009\degunit$;
deterministic fits, so the parity is exact rather than statistical) and gains $0.044\degunit$ over
per-trial bias subtraction, from physically sane parameters (Table~\ref{tab:gyrocal}). This is a parity cell, not a win: the group-law structure is not where
the gain lies on this task, and here parity is deterministic rather than statistical --- the learned
$(M,b)$ reaches the LM optimum on every window (Table~\ref{tab:gyrocal}).

\subsection{Deployment profile}
\label{sec:deploy}

Deployment is cheap and reproducible: one forward pass per window, no recurrent state or sampling
variance at inference, training without ensembles or reseeding; a fit takes minutes on one CPU, and
the full arm protocol (Section~\ref{sec:armcal}) --- all sizes, seeds and models --- takes hours. Energy and
embedded latency are not measured (Limitations).

\begin{table}[H]
\centering
\small
\caption{Gyroscope calibration (X6): geodesic error ($\degunit$) of strapdown
integration with corrected measurements on the held-out split; $\Delta=$ gain over bias-only
subtraction. The TinyGC-style row is measured under this protocol: its subnet reduces to the same
affine class~\citep{cui2024tinygc} and matches the LM fit within $0.002\degunit$ at every horizon;
$3$ seeds.}
\label{tab:gyrocal}
\begin{tabular}{lrrrr}
\toprule
Correction & $0.07\,$s & $0.35\,$s & $1.0\,$s & $\Delta_{\max}$ \\
\midrule
none (raw integration) & $0.594$ & $1.161$ & $1.453$ & --- \\
bias-only, per-trial & $0.593$ & $1.144$ & $1.374$ & --- \\
classical LM $(M,b)$ ($12$p) & $0.588$ & \best{1.100} & \best{1.362} & $0.044\degunit$ \\
LawKAN-SU(2), learned $(M, b)$ ($12$p) & \best{0.588} & $1.100$ & $1.371$ & $0.044\degunit$ \\
TinyGC-style subnet ($27$p) & $0.589$ & $1.101$ & $1.363$ & $0.043\degunit$ \\
\bottomrule
\end{tabular}
\end{table}

\begin{table}[H]
\centering
\small
\caption{Multi-platform FK calibration (X10, P48): held-out position error (mm,
mean $\pm$ std); our screw-coordinate model ($55/48$ parameters) vs the constant base/tool fit
($12$ parameters). FR3 data: \citep{philnad2024robotkinecal}; UR10: IPA Series A
\citep{landgraf2021hybrid,ipa2021dataset}; UR5: pauljac, 1020 poses, position-only
\citep{yao2026graph,pauljac2025ur5}. UR10 at $N{=}20$ is seed-unstable.}
\label{tab:x10}
\begin{tabular}{lccccc}
\toprule
Platform (LawKAN-SE(3)) & const.~fit, $N{=}80$ & $N{=}20$ & $N{=}40$ & $N{=}80$ & gain, $N{=}80$ \\
\midrule
FR3 ($55$\,p) & $3.34\pm0.09$ & $1.78\pm0.18$ & $1.43\pm0.10$ & \best{1.32\pm0.13} & $2.5\times$ \\
UR10 ($48$\,p) & $0.76\pm0.01$ & $1.16\pm1.26$ & $0.47\pm0.03$ & \best{0.43\pm0.01} & $1.8\times$ \\
UR5 ($48$\,p, pos.) & $0.51\pm0.02$ & $0.25\pm0.04$ & $0.16\pm0.02$ & \best{0.14\pm0.00} & $3.7\times$ \\
\bottomrule
\end{tabular}
\end{table}

\begin{figure}[H]
\centering
\caption{Three-platform FK calibration (X10): held-out error against training size per
platform; constant fit (grey), dense screw model, rotation-only sparse; gains at $N{=}80$:
$2.5\times$/$1.8\times$/$3.7\times$; UR10 at $N{=}20$ is seed-unstable.}
\label{fig:x10}
\includegraphics[width=0.95\columnwidth]{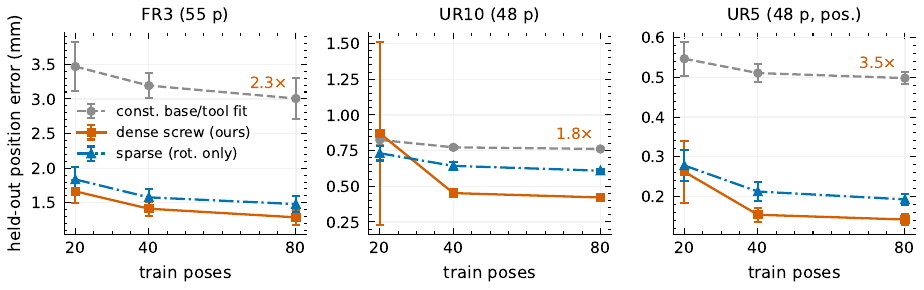}

\end{figure}

\begin{figure}[H]
\centering
\caption{Seed stability of the relaxation on Kinova (X32): per-seed socket RMS for the reduced
edge, the global head and the local kernel. The global head collapses on the two outlying seeds
($8$, $9$) of the ten-seed probe, above the reduced edge, while the local kernel stays below it on
every seed --- the instability is a small-data property of the global head, not of the correction.}
\label{fig:kinovaseeds}
\includegraphics[width=0.55\columnwidth]{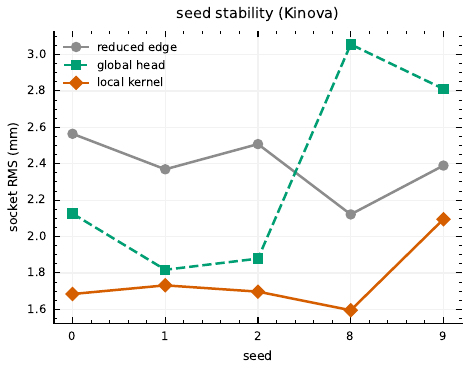}
\end{figure}

\begin{figure}[H]
\centering
\caption{Two further platforms with different measurement modalities. (a) MUKCa socket consistency
(KUKA iiwa14): held-out scatter for the nominal model, the reduced edge, the dense screw model, the
isolated correction and the joint control ($10$ seeds); the isolated correction removes the floor
below the edge. (b) RobotCali draw-wire range-only (ABB IRB 120): held-out mean $|\Delta L|$ against
training size; the isolated correction beats the exact screw and, for $N\ge160$, the same-class
Gauss--Newton fit (tying it at $N{=}80$), and the local kernel adds nothing here.}
\label{fig:newplat}
\includegraphics[width=0.98\columnwidth]{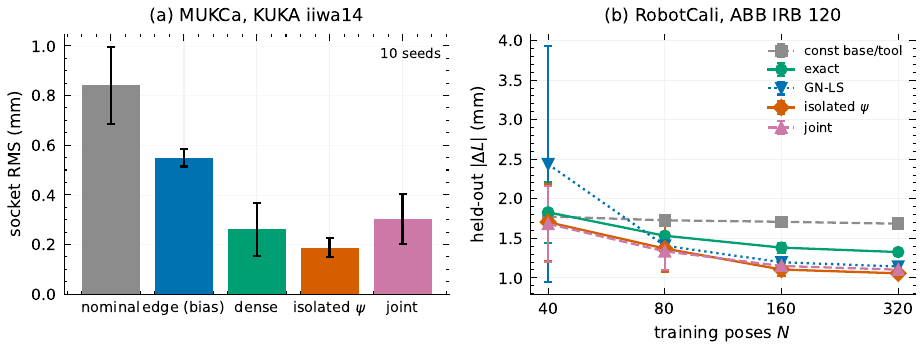}
\end{figure}

The isolated correction on the ABB range-only protocol is already ahead of the same-class exact
screw fit at $N{=}40$ ($1.70$ vs $1.83\,$mm, $p{=}0.039$; $10$ seeds) and below it at $N\ge80$; it is
below the GN fit numerically at $N\ge80$ and significantly from $N\ge160$ (Table~\ref{tab:rangeonly}); a
Gauss--Newton fit of the same class reaches $1.41/1.20/1.14\,$mm at $N{=}80/160/320$, numerically
between the exact and isolated results. On the socket protocol the dense screw ($55$p) control is in the
table and the joint control in Figure~\ref{fig:newplat}; on the FR3 unit the dense per-joint screw wins (a base-form boundary), consistent with its
large base mismatch. On the Kinova platform, the reduced edge underfits the large base
mismatch, so $\psi$ must absorb a large residual. The socket objective scores the scatter of each
socket around its \emph{own} predicted mean (\texttt{socket\_loss}), so a map that sends a whole
configuration class to one point has near-zero scatter --- a degenerate minimum. On $2/10$ seeds the
optimizer reaches this collapse basin instead of the correct calibration (iso $3.06/2.81$ vs edge
$2.12/2.39$; the paired test over all $10$ seeds gives $p{=}0.26$; TOST fails;
Figure~\ref{fig:kinovaseeds}); this is a property of the unregularized \emph{global}
head --- local-kernel and two-slot heads remove it (Section~\ref{sec:almost},
Appendix~\ref{app:probes}) --- whereas the shuffled-label control below shows the
metric is robust against degenerate minima when the constraint is real. In the shuffled-label control (\texttt{rng}(2000)
permutation of the concatenated sockets, all else identical) the isolated correction stays at
$60.5/59.3\,$mm on KUKA/Panda after $3000$ steps versus $0.19/0.36\,$mm under the real constraint,
so workspace collapse does not artificially minimize the socket metric. The registered pooled criterion (Fisher
over the ABB $N{=}80$ and KUKA front paired tests, $10$ seeds) gives $\chi^2{=}58.5$,
$p{\approx}6{\cdot}10^{-12}$. The extension rule admits Kinova to the pooled claim if its own gate
passes; under the recommended local kernel it does (kernel vs reduced edge $-0.55\,$mm,
$p{<}10^{-4}$; the kernel matches the dense screw, $p{=}0.10$, non-significant), so the extended claim is made: Fisher over
\{ABB $N{=}80$, KUKA front, Kinova front\} gives $\chi^2{=}78.1$, $p{\approx}9{\cdot}10^{-15}$ ---
across three platforms. Within-robot replication: on the two further KUKA
tool sessions isolated $\psi$ wins both ($0.231/0.381$ vs edge $0.611/0.747$,
$p{=}2{\cdot}10^{-4}/10^{-4}$). On both applied protocols the correction is fed the canonicalized
quaternion pair --- $q$ (with $w\ge0$) and $t$ per factor, $14$ numbers --- so the axis--angle chart
never enters it; the only residual is the antipodal seam at $w{=}0$, a bounded jump that costs a
peak ($13.2\to9.6\,$mm) but not the final error (X57).

\noindent\textbf{Known-deformation probes and the shape of the deformation.} Four registered probes
ask where the deformation comes from (all isolated on the frozen exact edge).
(i)~\emph{Gravity torque} (X19): a physics correction $\delta q_j=\alpha_j\tau_j+h(\tau_j)$
using the nominal gravity torque computed from the encoders is at parity with the exact model and
\emph{worse} than the agnostic $\psi$ at $N\ge60$ ($+0.23/+0.27\,$mm; generic $1.17/1.14$),
though it cures the agnostic small-$N$ overfit ($1.54$ vs $2.49$ at $N{=}20$): the deformation is
not gravity-shaped. (ii)~\emph{Workspace shift} (cluster split): the pose-$\psi$ loses
$1.6$--$1.9\times$ to the exact model when whole configuration clusters are held out
($p\le0.009$) --- a local fit, not a global law. (iii)~\emph{Backlash, synthetic} (X20): with
matched approaches the sign of the last joint motion predicts the residual direction ($1.000$
alignment, permutation $p<10^{-3}$), and the constrained correction removes $98\%$ of the
hysteresis floor and is inert without friction. (iv)~\emph{Backlash proxy on real data} (X21, collected
from an IPA Series A manipulator with real timestamps): inconclusive --- the recording-order signature does not rise above
the $0.44\,$mm residuals ($3/6$ joints positive, $p{=}0.92$); arm datasets carry no approach
trajectories. For the gyroscope, a coning probe (X18) finds the first-order floor at $286\,$Hz to
be quadrature-dominated ($0.167^\circ$; the BCH coning term $1.9^\circ$ partially cancels it),
removed by midpoint sampling ($0.005^\circ$); an unconstrained correction destroys it and a
constrained coning-statistic map removes only $3\%$.

\begin{table}[!htbp]
\centering
\small
\setlength{\tabcolsep}{4pt}
\caption{Which correction head to apply, by setting. The two-slot (the default) is trained jointly
(sequential-on-residual matches it, Section~\ref{sec:almost-theory}); the local kernel is the collapse-safe
choice for a small, fixed, seed-unstable configuration set; a plain global head suffices where the
deformation is smooth.}
\label{tab:whichhead}
\begin{tabular}{@{}p{0.31\columnwidth}p{0.19\columnwidth}p{0.42\columnwidth}@{}}
\toprule
Setting & Head & Why \\
\midrule
Deformed composition, data not scarce (KUKA, Panda) & two-slot on KUKA; global on Panda & KUKA: two-slot lowest; Panda: global lowest, two-slot within noise \\
Deformed arm deep-dive (FR3) & two-slot & ties the best pure head; kernel-safe \\
Small fixed set, seed-unstable (Kinova) & local kernel & the global slot re-collapses; the kernel stays below the edge \\
Range-only, smooth (ABB) & global (MLP) & the kernel adds nothing \\
\bottomrule
\end{tabular}
\end{table}

\noindent\textbf{Why the baselines differ.} Each protocol is compared with its classical
\emph{same-class} estimator. For arm calibration the classical estimator is the product-of-exponentials
least-squares (\`a la \citet{philnad2024robotkinecal}), which on the range-only ABB protocol is the
exact screw model and its Gauss--Newton fit --- so the correction is compared with the Gauss--Newton
(its strongest classical form). The MUKCa socket protocol has no Gauss--Newton: its classical method
is the consistency-and-distortion optimization of \citet{franzese2025mukca,mukca2024dataset}, whose full per-joint
screw fit is our \emph{dense screw} (the same-class reference); the LawKAN base is the
uncorrected \emph{reduced edge} (the correction's base).

\section{Two isolated probes: FK calibration and ODE dynamics}
\label{app:probes}

\subsection{Forward-kinematics calibration (5-DOF arm, synthetic)}
\label{sec:fk}

On a synthetic $5$-DOF arm with a fixed per-joint miscalibration, LawKAN-SU(2) ($150$
parameters; joint-wise edges $v_j=\theta_j a_j+\mathrm{Fourier}_j(\theta_j)$, composed along the
chain) reaches $0.159\pm0.095\degunit$ against $0.661\pm0.267$ for an MLP, a $4\times$ margin that is
stable over $3$ seeds (Table~\ref{tab:fk}); per-joint identifiability is limited by the calibration
gauge. What matters here is the linear-in-$\theta$ channel (a pure Fourier edge saturates near
$18\degunit$); the real-hardware counterpart is Section~\ref{sec:armcal}.

\begin{table}[H]
\centering
\small
\caption{Forward-kinematics calibration (synthetic 5-DOF arm, 3 seeds): endpoint-orientation error.}
\label{tab:fk}
\begin{tabular}{lrr}
\toprule
Model & Params & Error ($\degunit$) \\
\midrule
LawKAN-SU(2) & $150$ & \best{0.159 \pm 0.095} \\
MLP $h{=}32$ & $1380$ & $0.661 \pm 0.267$ \\
MLP $h{=}32$ $+$ RPMG & $1380$ & $0.672 \pm 0.293$ \\
\bottomrule
\end{tabular}
\end{table}

\subsection{Rigid-body dynamics: a state-conditioned quadratic edge solves the ODE case (E25/E25b)}
\label{sec:rigid}

Regressing the non-chaotic solution map of the Euler equations ($20$-step rollouts from $\omega_0$
only), a \emph{state-conditioned} quadratic edge ($39$ parameters) reproduces the analytic integrator
to $0.0818\pm0.0001\degunit$ (its numerical floor), while the linear edge stalls at $7.13$ and
the cubic diverges ($4/5$, the order rule). Feed-forward one-pass controls (E25b) do not unroll
(best $0.84\degunit$), and MLPs need $34$--$450\times$ more parameters for $3.4$--$6.1\times$ worse
error (Table~\ref{tab:e25}). The ODE case is thus \emph{solved}, not merely
bounded: the \emph{state}, not capacity, closes the gap. With an estimated state the edge
order is no longer necessary (reported in the prediction log).

\begin{table}[H]
\centering
\small
\caption{Rigid-body dynamics: state-conditioned (E25) vs feed-forward (E25b) edges,
degrees. State edges see the evolving $\omega$ (the model unrolls the ODE); feed-forward edges see
only $\omega_0$ in one pass; the analytic integrator is at the numerical floor.}
\label{tab:e25}
\begin{tabular}{lrr}
\toprule
Model & Params & Error ($\degunit$) \\
\midrule
Analytic integrator (metric floor) & $0$ & $0.0816$ \\
\midrule
\multicolumn{3}{@{}l}{\emph{State-conditioned (LawKAN-SE(3), E25; \emph{state} edge, sees the evolving $\omega$):}} \\
\quad quadratic & $39$ & \best{0.0818 \pm 0.0001} \\
\quad linear & $12$ & $7.13 \pm 0.02$ \\
\quad cubic & $120$ & diverged ($4/5$) \\
\midrule
\multicolumn{3}{@{}l}{\emph{Feed-forward (LawKAN-SE(3), E25b; edge sees only $\omega_0$, one pass):}} \\
\quad quadratic & $39$ & $1.58 \pm 0.02$ \\
\quad cubic & $120$ & $0.84 \pm 0.03$ \\
\quad linear & $12$ & $5.21 \pm 0.04$ \\
\quad 20-step fixed input & $99$ & $7.02 \pm 5.45$ \\
\quad Fourier $K{=}1$ (E17) & $54$ & $11.31 \pm 4.59$ \\
\midrule
MLP $h{=}32$ (baseline) & $1316$ & $0.2765 \pm 0.0085$ \\
MLP $h{=}128$ (baseline) & $17540$ & $0.4991 \pm 0.5519$ \\
\bottomrule
\end{tabular}
\end{table}

\subsection{Real-data audit: Baxter arm dynamics (E26b)}
\label{sec:baxter}

The real-data counterpart of the ODE probe: on recorded Baxter arm trajectories
\citep{baxterzenodo} a state-conditioned edge learns $\ddot q = f(q,\dot q,\tau)$, and the exact
integrator (semi-implicit Euler) composes the $14$-dimensional state over $H{=}50$ steps; the node
carries no parameters. Training uses the squared and random trajectories ($70/30$), the test is the
held-out segment and the transfer target is the disjoint circle trajectory ($3$ seeds; metric: root-mean-square error (RMSE)
of the state increment; the effective sample step from the data is $2.4\,$ms against the journal's
$1\,$ms assumption, so both rates are reported). On transfer the linear edge with exact integration
is the only learned model below the constant-velocity baseline and below the flat MLP, while the
higher-capacity edges win in-distribution and do not transfer; the two predictions were registered
in the project plan before the run, and we report that one of them (the MLP-edge variant) is
refuted. The exact-integrator component therefore provides a modest, budget-sensitive advantage on this real
task, not an asymptotic win, as Table~\ref{tab:baxter} shows.

\begin{table}[H]
\centering
\small
\caption{Real Baxter arm dynamics (E26b): RMSE of the integrated state increment ($14$-dim, mixed
units) over $H{=}50$ steps, three seeds. ``In-dist'' is the held-out segment of squared/random;
``circle'' is the disjoint transfer trajectory. The constant-velocity baseline is training-free;
the flat MLP maps to the increment in one pass with no integration (the $2.0\,$ms row is
rounded from the effective $2.4\,$ms sample step). The edge rows are
LawKAN-SE(3) variants (a \emph{state} edge plus the exact integrator, or a flat edge).}
\label{tab:baxter}
\begin{tabular}{llrr}
\toprule
Edge (learned) & Params & In-dist & Circle \\
\midrule
\multicolumn{4}{@{}l}{\emph{dt $=1.0\,$ms}}\\
zero (constant velocity, no learning) & $0$ & $0.063$ & $0.126$ \\
linear $+$ exact integrator & $154$ & $0.053$ & \best{0.114} \\
physics features $+$ exact integrator & $504$ & $0.051$ & $0.161$ \\
MLP $+$ exact integrator & $6023$ & \best{0.040} & $0.129$ \\
flat MLP (no integration) & $6478$ & $0.040$ & $0.138$ \\
\midrule
\multicolumn{4}{@{}l}{\emph{dt $=2.0\,$ms}}\\
zero (constant velocity, no learning) & $0$ & $0.063$ & $0.125$ \\
linear $+$ exact integrator & $154$ & $0.051$ & \best{0.110} \\
physics features $+$ exact integrator & $504$ & $0.050$ & $0.159$ \\
MLP $+$ exact integrator & $6023$ & \best{0.037} & $0.139$ \\
flat MLP (no integration) & $6478$ & $0.040$ & $0.138$ \\
\bottomrule
\end{tabular}
\end{table}

\subsection{Almost laws: details, controls and diagnostics (X13/X14/S1/S2/S3/X16)}
\label{sec:learnednode}

The Corollary bounds the \emph{node} from the other side: if the composition of a real system is not
exactly the chosen group law (compliance, thermal deformation, backlash), the mismatch leaves a
training-independent floor that no edge can remove. We relax the law with a small learned correction
while keeping everything else fixed:
$R(A,B)=\mathrm{compose}(A,B)\exp(\psi(f(A),f(B)))$, where $\psi$ has a \emph{zero} last layer
(at initialization the relaxed model is exactly the exact model) and is shared across the chain's
compositions. Early $\psi$ variants were a linear Fourier head over the $30$ pair
features (per factor: four chart angles $(\theta,a_x,a_y,a_z)$ in the basis
$\{1,\sin x,\cos x\}$, $12$, plus three translations, $15$) read out to the six
$\mathfrak{se}(3)$ components ($30\cdot6+6=186$ parameters; with the dense per-joint edge $240$,
$\psi$-only $198$) and an MLP ($14{\to}16{\to}6$, $342$; totals $396/354$). These change only the
functional class inside the same slot (Table~\ref{tab:learnednode}); the \emph{final} default is the two-slot head -- a global
linear readout ($186$) plus a local kernel ($120$), $306$ in all ($360$ with the dense edge,
Section~\ref{sec:almost-theory}) -- which is what Table~\ref{tab:armcal} reports.

On the real FR3 the relaxation removes the floor. Under the two-slot head this holds at every
training size, significantly from $N{\ge}40$; at $N{=}20$ it is within noise ($1.51$ vs $1.53\,$mm,
$p{=}0.84$; isolated $1.12$ at $N{=}80\,$mm,
Table~\ref{tab:armcal}); the earlier dense-edge Fourier realization ($1.20/1.18\,$mm at $N{=}60/80$, Table~\ref{tab:learnednode})
overfit at $N{=}20$ ($2.49\,$mm), which the local slot repairs.
With the exact edge frozen the isolated variant keeps the win,
while joint training is the compensation control.
The Fourier variant matches the MLP within noise at $N{\ge}40$ while being
smaller and interpretable (it also improves the $\psi$-only form, $-0.074\,$mm at $N{=}80$,
$p{=}0.023$). Training the two slots together is as good as isolating them: the sequential-on-residual
variant (kernel first, then the global slot on its residual) matches joint training ($1.64$ vs
$1.51\,$mm at $N{=}20$; $1.14$ vs $1.12\,$mm at $N{=}80$), and the global-slot weight grows with $N$
($0.04\to1.37$); the slots are not orthogonal (a gauge, Section~\ref{sec:almost-theory}).

\noindent The ladder of exclusions on the correction head is summarized in
Table~\ref{tab:slot} (Section~\ref{sec:almost-theory}); the per-platform detail follows.

\begin{table}[H]
\centering
\small
\caption{The correction slot: one support axis, measured scopes. Each entry was
registered before its run (App.~\ref{app:predictions}); the winding and magnitude
hypotheses were refuted and are reported. The two-slot sum is the default: it
dominates both pure heads at every training size (App.~\ref{app:probes}).}
\label{tab:slot}
\begin{tabular}{@{}p{0.17\columnwidth}p{0.12\columnwidth}p{0.30\columnwidth}p{0.33\columnwidth}@{}}
\toprule
Head & Support & Verdict on our battery & Scope (when it wins) \\
\midrule
global linear & global & no collapse protection; overfits at small $N$ &
smooth distributed deformations, data abundant; statistical niche at large $N$ \\
bounded chart (rotation) & global & collapse not cured ($p{=}0.68$) &
\emph{refuted as our mechanism}; retained for large-angle deformations
($\mathrm{sat}_{\max}\!\to\!1$ signals activation) \\
magnitude bound & global & never engaged ($\mathrm{sat}_{\max}{<}1$) &
\emph{refuted as our mechanism}; a safety cap where extrapolation is possible \\
$+$ weight decay & global, capped & cures; dose $\propto$ size, narrow window &
second layer on the global slot; dose from the size table (X26c) \\
local kernel & local & cures; no dose; improves healthy fits &
localized mechanisms (backlash, friction, compliance); data-starved corner \\
\textbf{two-slot} (default) & sum & dominates both pure heads at every $N$;
$|g_{\mathrm{slot}}|$ grows with $N$ & no per-task choice required \\
\bottomrule
\end{tabular}
\end{table}

Three controls separate the placement from capacity and from identifiability.
\emph{(i) Placement (S3):} a parameter-matched residual MLP ($377$ parameters, end-effector frame)
does not improve over the exact model ($1.85$ vs $1.49\,$mm at $N{=}40$; paired vs the relaxed node
$+0.50$, $p{=}0.008$), so the gain is where the correction is placed, not its budget.
\emph{(ii) No gain on exact-law data (S1a/S1b):} on synthetic data generated by an \emph{exact}
serial-chain law the relaxation never wins and overfits at small $N$ (paired $+2.61\,$mm worse at
$N{=}20$, $p{=}0.026$; $+0.18\,$mm at $N{=}80$, $p{=}0.022$; $+0.07\,$mm at $N{=}160$, $p{=}0.059$),
while on data deformed by a known pair-dependent function it recovers the noise floor
($1.40$ vs $11.5\,$mm at $N{=}160$, $p{=}0.014$; exact stuck at its floor).
\emph{(iii) A second real dataset (S2):} on the independent UR5 set \citep{zenodo18954650} (position-only) the relaxation
(old MLP head) gives no gain on held-out data ($+0.02\,$mm at $N{=}80$, $p{=}0.23$; worse at small $N$), but the
joint optimum is compensated rather than flat: with the correction switched off at test the same
model's error rises from $0.40$ to $2.1\,$mm, i.e.\ the edge drifted and $\psi$ absorbed the drift
(identifiability, next paragraph).

\emph{Identifiability of the correction, and isolation.} The per-joint edge (constant screws) and
the composition correction (a pair-dependent term) are not orthogonal: a retuned edge reproduces
part of any correction, so joint training is non-identifiable and may end in a compensated solution
that fits but does not transfer. The third real platform (UR10; IPA Series A \citep{landgraf2021hybrid,ipa2021dataset}) shows the failure mode cleanly: with
no deformation to find, the jointly trained model (S8) is worse than its own exact model at every
size (test $0.736\pm0.169$ vs $0.429\pm0.006\,$mm at $N{=}80$; $2.63$ vs $1.16\,$mm at $N{=}20$),
its train error is not better ($0.43$ vs $0.39\,$mm), and with the correction switched off at test
its error rises to $11.7\,$mm (seed~$52$, $N{=}80$) --- the edge itself moved and $\psi$
compensates it. On UR5 the same
compensation occurs (mean per-composition correction norm $2\cdot10^{-2}$, correction-off error
$2.1\,$mm) but the compensated solution still generalizes ($0.404$ vs $0.410\,$mm). Freezing the
exact edge and training only the correction (isolation) removes the drift: on UR10 (five seeds) the
isolated model is at parity at $N{=}20/80$ but loses $+0.036\,$mm at $N{=}40$ ($p{=}0.044$)
($1.146/0.502/0.433$ vs $1.164/0.465/0.429\,$mm; paired $-0.018$/$+0.036$/$+0.004$) where joint training was $+1.47$/$+0.27$/$+0.31\,$mm worse, and the
correction norm shrinks below $10^{-3}$. On FR3 (five seeds) the \emph{Fourier}-isolated variant keeps the large-$N$ win
($-0.209$ at $N{=}60$, $p{=}0.011$; $-0.260$ at $N{=}80$, $p{=}0.005$) but retains the small-$N$
overfit ($+0.956$ at $N{=}20$, $p{=}0.039$); the final two-slot head removes it ($1.51$ vs
$1.53\,$mm at $N{=}20$, $p{=}0.84$; $-0.23/-0.28$ at $N{=}60/80$). On UR5 the isolated correction is harmless from
$N{\ge}60$ ($+0.002$, $p{=}0.64$) with a small residual cost at $N{=}20/40$ ($+0.10/+0.07\,$mm).
Decomposing the trained correction's effect field against the edge tangent space (finite differences
over configurations, G3) confirms the mechanism: on FR3 the edge maps absorb $98\%$ of the jointly trained correction
(correction-off error $10.2$ vs $1.38\,$mm on) while the isolated correction is
$88\%$ orthogonal; on UR10 the split is $99\%$/$85\%$; on UR5 the edge field can still absorb the isolated correction, yet the correction remains
harmless ($0.402$ vs $0.410\,$mm) --- the discriminator is the correction-off
error, not the share. Capacity and gauge are complementary levers: restricting the correction to the
top-$6$ chart atoms softens the small-$N$ overfit ($1.81$ vs $2.49\,$mm at $N{=}20$) while keeping
the win ($1.28$ vs $1.41\,$mm at $N{=}60$, $p{=}0.034$), and projecting it to the edge-orthogonal
complement lowers the absorbable share to $0.10$--$0.14$ at $N{\ge}60$ without losing the win (G4).
No capacity/gauge variant removes the Fourier small-$N$ overfit; a validation gate remains the
lever there.

\emph{What the learned correction sees (X16, isolated, Fourier head).} Trained at $N{=}80$ with the exact edge frozen,
the Fourier weights split as accumulated pose $A$ vs step $B$: angle groups $0.30\pm0.05$ and
$0.38\pm0.04$, translation groups $0.14\pm0.01$ and $0.65\pm0.10$ (mean$\pm$std over five seeds). The
input ablation (held-out error with a channel group zeroed) is $1.30\,$mm full, $4.71$ without $A$
angles, $3.75$ without $A$ translations, $4.12$ without $B$ angles, $3.95$ without $B$ translations:
both sides of the pair matter ($2.9$--$3.6\times$ degradation), i.e.\ the isolated correction is
\emph{pair-dependent} and, being $88\%$ orthogonal to the edge (G3), the pair-dependence is a
property of the deformation rather than of a joint-training artifact. The top univariate
contributions are the $y$-axis component of $A$, the $x$-axis component of $B$ and $\theta_B$
(Figure~\ref{fig:learnedlaw}a). Raw $\psi$ weights are stored in the diagnostic dumps. With the final two-slot head the same
split holds: full $1.248\,$mm, without $A_{\rm ang}$ $4.376$, without $B_{\rm ang}$ $3.683$,
without $A_{\rm tr}$ $3.372$, without $B_{\rm tr}$ $3.345$ (X16b); the isolated two-slot
correction is $88\%$ orthogonal to the edge (G3b, orthogonal share $0.880$).

\emph{The same structure on the edge does not transfer.} Shared error modes (X12) lose to free
per-joint corrections ($1.96$ vs $1.49\,$mm at $N{=}40$; the full-coupling variant is still $2.47$
at $N{=}20$); adding a chain coupling on top of per-joint corrections is exactly neutral
($\beta\to0$; X12b), and a shared position-dependent source helps only by $0.07\,$mm at
$N{=}40/60$; and a deterministic error-accumulation law on the edge improves in-distribution fit
but not transfer (E26c, real Baxter dynamics). The gain comes specifically from relaxing the
\emph{node}.

\begin{table}[H]
\centering
\footnotesize
\setlength{\tabcolsep}{2pt}
\caption{Almost laws on the real FR3: held-out position error (mm), 5 seeds, same splits as
Table~\ref{tab:armcal}. Extends Table~\ref{tab:armcal} with the remaining variants ($\psi$-only,
MLP-$\psi$, matched residual MLP); the exact, GN and two-slot rows are in that table (the two-slot
rows are repeated here for comparison). Relaxed
variants add a zero-initialized shared correction $\psi$ to the law; $\psi$-only counterparts have no
per-joint corrections. The single-head variants show that the win does not depend on one head.}
\label{tab:learnednode}
\begin{tabular}{lrrrr}
\toprule
Model (params) & $N{=}20$ & $N{=}40$ & $N{=}60$ & $N{=}80$ \\
\midrule
LawKAN-SE(3), node: relaxed (two-slot), isolated ($306$p) & \best{1.51\pm0.24} & $1.24\pm0.08$ & $1.14\pm0.07$ & $1.12\pm0.11$ \\
LawKAN-SE(3), node: relaxed (two-slot), joint control ($306$p) & $1.52\pm0.31$ & \best{1.19\pm0.04} & \best{1.08\pm0.06} & \best{1.09\pm0.17} \\
LawKAN-SE(3), node: relaxed (Fourier $\psi$), $\psi$ only ($198$p) & $2.55\pm0.70$ & $1.53\pm0.16$ & $1.20\pm0.09$ & $1.18\pm0.14$ \\
LawKAN-SE(3), node: relaxed (MLP $\psi$) ($396$p) & $2.56\pm0.93$ & $1.35\pm0.08$ & $1.15\pm0.09$ & $1.15\pm0.15$ \\
LawKAN-SE(3), node: relaxed (MLP $\psi$), $\psi$ only ($354$p) & $2.40\pm0.45$ & $1.46\pm0.11$ & $1.25\pm0.07$ & $1.26\pm0.15$ \\
Matched residual MLP ($377$, opponent) & $2.57\pm0.56$ & $1.85\pm0.23$ & $1.48\pm0.31$ & $1.33\pm0.23$ \\
\bottomrule
\end{tabular}
\end{table}

\begin{figure}[H]
\centering
\caption{(a) S1 synthetic controls: on the exact law the relaxed and exact models coincide at large
$N$ (at $N{=}20$ the relaxed model overfits), on the deformed law the relaxed model reaches the noise
floor while the exact model stalls. (b) S2: on the independent UR5 dataset the relaxation gives no
gain --- the correction yields no gain where the data obey the law.}
\label{fig:lawdiag}
\includegraphics[width=0.78\columnwidth]{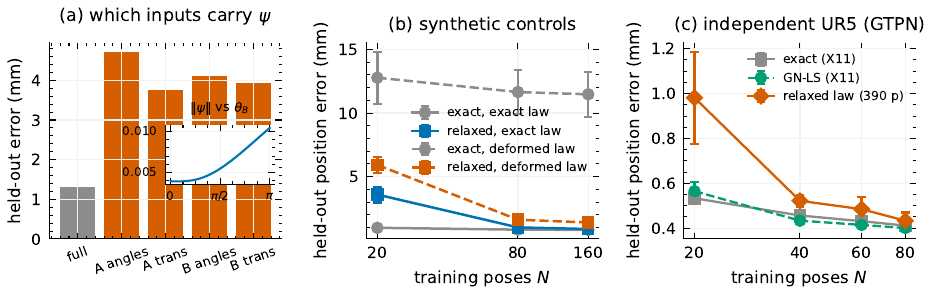}
\end{figure}

\section{Compact-instance construction and mechanism}
\label{app:construction}

\subsection{The LawKAN-SU(2) construction}

\noindent\textbf{Setup.} Inputs are rotations given by axes $a_i \in S^2$, encoded as phase pairs
(azimuth $\alpha_i$, polar angle $\beta_i$) $\in S^1 \times [0,\pi]$, the longitude/latitude of the
axis; the target is the composition $q_n\cdots q_1$ applied to a probe vector, read out as two
angles. Each edge answers ``how should this axis be rotated?'' and the node answers ``in which order
do the rotations multiply?''; the exp-map keeps every intermediate a valid rotation, so no
normalization step can divide by a near-zero norm.

\noindent\textbf{Edge: Fourier function of a phase pair $\to \so(3)$.} Each group $i$ has an edge that
sees both phases,
\begin{equation}
\varphi_i(\alpha,\beta) = \sum_{m,n \in \{-K..K\}} w_{mn}\, e_m(\alpha)\, e_n(\beta) \in \mathbb{R}^3,
\end{equation}
with $\{e_k\}$ the real Fourier basis ($2K{+}1$ per phase), so $(2K{+}1)^2 \times 3$ scalars per edge
($9\times3$ at $K{=}1$); the output $v \in \mathbb{R}^3$ is read as a Lie-algebra element.

\noindent\textbf{Node: group law in fixed order.} $q_i = \exp(v_i) = \cos|v_i| + (v_i/|v_i|)\sin|v_i|$
(always unit; no singularity at $v{=}0$ up to a removable $0/0$ handled by the series). The node
computes the Hamilton product in the data's composition order, $Q = q_n\cdots q_1$, with the operand
order fixed at the architecture level rather than learned. Quaternions make composition a single
constraint-free multiplication and are the cheapest \emph{exact} encoding of the group
(Section~\ref{sec:accounting}).

\noindent\textbf{Output.} $Q$ rotates the probe; direction is read as (azimuth, polar). Loss: geodesic
on $S^2$ (the angle between directions), $1 - \cos\Delta$ (avoids acos boundary NaNs); the reported
metric is that angle in degrees.

\noindent\textbf{Wiring.} The parameter-annotated wiring, with the learned tensors marked, is
Figure~\ref{fig:wiring} below.

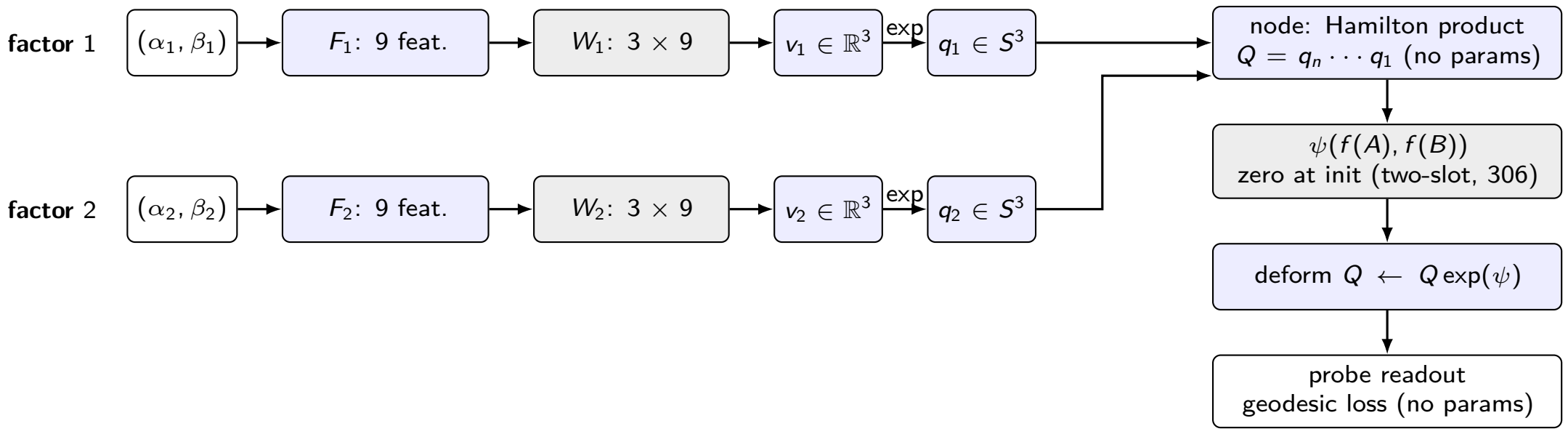
\begin{figure}[!htbp]
\centering
\caption{LawKAN-SU(2) wiring for the $n{=}2$, $K{=}1$ instance (the general case has one such edge
per factor). Inputs are the axis phase pairs $(\alpha_i,\beta_i)$ --- azimuth and polar angle of
factor $i$'s rotation axis (Appendix~\ref{app:notation}) --- mapped to the nine fixed Fourier
features $F_i$ (the edge's design inputs); the learned tensor is the edge matrix $W_i$ (its
coefficients): $n\times3\times9$ scalars in total
($54$ here), and its output $v_i$ is exponentiated to the unit quaternion $q_i$. The node multiplies
in the fixed order ($Q=q_n\cdots q_1$, no parameters); the optional zero-initialized
correction deforms the composition, $Q\leftarrow Q\exp(\psi(f(A),f(B)))$, and is the only
other learned tensor (two-slot, $306$); the readout carries no parameters.}
\label{fig:wiring}
\begin{tikzpicture}[font=\figfont,
  box/.style={draw, rounded corners=2pt, align=center, inner sep=2.5pt, minimum height=6mm},
  learned/.style={box, fill=black!7}, exact/.style={box, fill=blue!7}, io/.style={box, fill=white},
  ar/.style={-{Latex[length=1.4mm]}, semithick}, ln/.style={semithick}]
\node[io] (a1) {$(\alpha_1,\beta_1)$};
\node[exact, right=4mm of a1, text width=17mm] (f1) {$F_1$: $9$ feat.};
\node[learned, right=4mm of f1, text width=16mm] (w1) {$W_1$: $3\times9$};
\node[exact, right=4mm of w1] (v1) {$v_1\in\mathbb{R}^3$};
\node[exact, right=4mm of v1] (q1) {$q_1\in S^3$};
\draw[ar] (a1)--(f1); \draw[ar] (f1)--(w1); \draw[ar] (w1)--(v1);
\draw[ar] (v1)-- node[above, font=\figfont\itshape, inner sep=0.5pt]{$\exp$} (q1);
\node[left=1.5mm of a1, font=\figfont\bfseries]{factor $1$};
\node[io, below=9mm of a1] (a2) {$(\alpha_2,\beta_2)$};
\node[exact, right=4mm of a2, text width=17mm] (f2) {$F_2$: $9$ feat.};
\node[learned, right=4mm of f2, text width=16mm] (w2) {$W_2$: $3\times9$};
\node[exact, right=4mm of w2] (v2) {$v_2\in\mathbb{R}^3$};
\node[exact, right=4mm of v2] (q2) {$q_2\in S^3$};
\draw[ar] (a2)--(f2); \draw[ar] (f2)--(w2); \draw[ar] (w2)--(v2);
\draw[ar] (v2)-- node[above, font=\figfont\itshape, inner sep=0.5pt]{$\exp$} (q2);
\node[left=1.5mm of a2, font=\figfont\bfseries]{factor $2$};
\node[exact, right=16mm of q1, text width=30mm] (n) {node: Hamilton product\\ $Q=q_n\cdots q_1$ (no params)};
\node[learned, below=4mm of n, text width=30mm] (p) {$\psi(f(A),f(B))$\\ zero at init (two-slot, $306$)};
\node[exact, below=4mm of p, text width=30mm] (e) {deform $Q\leftarrow Q\exp(\psi)$};
\node[io, below=4mm of e, text width=30mm] (rr) {probe readout\\ geodesic loss (no params)};
\draw[ar] (q1.east) -- (n.west);
\draw[ar] (q2.east) -- ++(6mm,0) |- ([yshift=-3mm]n.west);
\draw[ar] (n)--(p); \draw[ar] (p)--(e); \draw[ar] (e)--(rr);
\end{tikzpicture}
\end{figure}

\subsection{Minimal-order design ($K{=}1$)}
\label{sec:k1}

For LawKAN-SU(2), the \emph{minimal-order construction} is Lemma~\ref{lem:kan-exact}: at $K{=}1$ the
nine edge features $\{1,\cos\alpha,\sin\alpha\}\otimes\{1,\cos\beta,\sin\beta\}$ already contain the
three axis components, so weight $\theta_i/2$ on them makes the edge output $\theta_i a_i/2$ whose
exponential is the exact per-factor rotation, and no lower order suffices
(Lemma~\ref{lem:kan-minimal}). The edge only has to learn the constant ``half the rotation
amount'', and the Hamilton node returns the exact composition for every input. This explains the
data efficiency and extrapolation, motivates the pruning below, and identifies the parameterization
that avoids the singular $q/\Vert q\Vert$ regimes.

\subsection{Pruning the edge to the analytic edge (E19; predictions P23--P27)}
\label{sec:pruning}

In exponential coordinates the target is $(\theta/2)\,\mathbf{a}$, whose three axis components are
themselves three of the nine $K{=}1$ Fourier features. We can therefore freeze the other eighteen
weights to zero (Figure~\ref{fig:support}) and prune the edge to this analytic edge.

The constructed weights are exact. More interestingly, the trained sparse edge ($18$ parameters) is
statistically \emph{more} accurate than the dense edge ($54$ parameters) and is closer to the Bayes
floor (Table~\ref{tab:sparse-ablation}; paired test, $5/5$ seeds). A random equal-size control
($18$ p drawn from the same dictionary, $5$ masks $\times$ $5$ seeds) gives $32.9 \pm 9.0\degunit$
(E19b/P63): the identity of the support, not the parameter count, is the operative factor.
Optimization recovers the constructed point itself: after training, the analytic
$\theta/2$ weights are matched to $\max_i|w_i-\theta_i/2|=6$--$8\times10^{-3}$ (relative
Frobenius $0.7$--$0.9\%$ on seeds $0$--$1$; E19.7), up to the sign ambiguity of
Remark~\ref{rem:nonunique} --- so the optimizer finds the construction, not merely a function
close to it. The advantage persists on the noise curve, at $n{=}3$ and along chains to $n{=}4$.

\begin{figure}[H]
\centering
\begin{minipage}[t]{0.54\columnwidth}
\centering
\captionof{figure}{Analytic-support pruning. The nine $K{=}1$ features (rows: $\beta$-terms;
columns: $\alpha$-terms); the exact solution uses only the three axis components (shaded), so $18$ of
the $54$ parameters at $n{=}2$ suffice.}
\label{fig:support}
\begin{tikzpicture}[font=\figfont, xscale=0.85, yscale=0.85]
\node at (0.5,3.35) {$1$};
\node at (1.5,3.35) {$\cos\alpha$};
\node at (2.5,3.35) {$\sin\alpha$};
\node at (-0.45,0.5) {$1$};
\node at (-0.45,1.5) {$\cos\beta$};
\node at (-0.45,2.5) {$\sin\beta$};
\fill[blue!16] (0,1) rectangle (1,2);
\fill[blue!16] (1,2) rectangle (2,3);
\fill[blue!16] (2,2) rectangle (3,3);
\foreach \i in {0,1,2} \foreach \j in {0,1,2} \draw (\i,\j) rectangle (\i+1,\j+1);
\node at (0.5,1.5) {$v_3$};
\node at (1.5,2.5) {$v_1$};
\node at (2.5,2.5) {$v_2$};
\end{tikzpicture}
\end{minipage}\hfill
\begin{minipage}[t]{0.42\columnwidth}
\centering
\captionof{table}{Fourier-order and feature sparsity ablation of LawKAN-SU(2) (E19.2, comp2): the exact solution
needs three of the nine features; E19.2 uses 5 seeds, and dense $K{=}2$ is an independent
10-seed rerun of E13.}
\label{tab:sparse-ablation}
\begin{tabular}{lcc}
\toprule
Edge & Params & Error, $^\circ$ \\
\midrule
dense $K{=}2$ & 150 & $2.87 \pm 0.03$ \\
dense $K{=}1$ & 54 & $2.79 \pm 0.02$ \\
sparse (3 of 9) & 18 & \best{2.75 \pm 0.02} \\
\bottomrule
\end{tabular}
\end{minipage}
\end{figure}

The pruned edge is meaningful only where the analytic support is right (fixed known angles, or the
IMU increment, Section~\ref{sec:imu}); elsewhere the dense edge is the architecture-agnostic default and is
required (random angles, rigid-body, the null task), because pruning would discard needed degrees of
freedom. What matters is \emph{where the exact solution lives}, not capacity.

\section{Mechanism details: variable angles, dead zones and the order rule}
\label{app:mech}

\subsection{Variable angles: the tensor extension (E29)}
\label{sec:varangles}

Lemma~\ref{lem:kan-exact} assumes fixed angles, which holds for the fixed-angle composition tasks but
not for sensors or encoders, whose rotation amount changes at every step. If we supply $\theta$ as an
input and extend the edge to the tensor
product $\{1,\theta\}\otimes\{1,\cos\alpha,\sin\alpha\}\otimes\{1,\cos\beta,\sin\beta\}$
($18$ features), the three $\theta$-scaled axis components become single features; putting weight
$1/2$ on them gives $v=(\theta/2)a$, so the architecture is exact for
\emph{arbitrary, variable} angles (the variable-angle remark of Section~\ref{sec:principle}). The IMU probe uses the factorized magnitude+phase edge ($27$ parameters, pruned to $18$); the
tensor construction above is the variable-angle edge.

Table~\ref{tab:varangles} confirms it on a composition of $n$ elements with variable angles and a
full-$\mathrm{SO}(3)$ readout: the constructed tensor weights are exact, and the trained tensor edge
stays near the floor. By contrast, the fixed-angle edge lacks the $\theta$ channel and leaves a large
positive floor that \emph{grows with the chain length}, exactly as the feature-span prediction says;
dropping $\alpha$ or $\beta$ is worse still. Exactness, then, is not special to fixed rotations: it is
a property of the feature span.

\begin{table}[H]
\centering
\small
\caption{Variable-angle exactness (E29; $\theta\sim U[0.2,2.5]$, full SO(3)
readout): constructed tensor weights are exact to $9.7\times10^{-7}$; edges missing the $\theta$
channel (or another feature group) leave positive floors.}
\label{tab:varangles}
\begin{tabular}{lrrr}
\toprule
Edge & $n{=}2$ & $n{=}3$ & $n{=}4$ \\
\midrule
LawKAN-SU(2), tensor edge ($18$ features, trained) & \best{0.28} & \best{0.14} & \best{0.35} \\
fixed, $9$ features (no $\theta$) & $24.4$ & $30.2$ & $35.0$ \\
no $\beta$ & --- & $38.8$ & --- \\
no $\alpha$ & --- & $53.2$ & --- \\
\bottomrule
\end{tabular}
\end{table}

\subsection{Dead zones and the exp-map (quantified ablation)}
\label{sec:dead}

Here we compare two forward parameterizations of the node output under identical training,
normalizing the raw quaternion ($q/\Vert q\Vert$) versus the exp-map (a retraction), on $n{=}2$,
$\sigma{=}0.03$, $10$ runs. The issue is training stability: a well-behaved parameterization should
not let the optimizer fall into a dead configuration and stay there.

Normalization fails on $3$ of $10$ runs (up to $32.3\degunit$; the raw norm $\Vert\qtilde\Vert$ passes through a mid-training minimum of
$0.30$--$0.50$ (final $0.65$--$1.18$), while the exp-map fails none ($2.78$--$2.80\degunit$).
All ten normalization runs pass through a mid-training norm minimum, and three never recover. The
\emph{dead-zone mechanism} (Appendix~\ref{app:norm}) is the normalization Jacobian
$1/\Vert\qtilde\Vert$, so survival is decided by noise near the norm minimum. Because RPMG
\citep{chen2022rpmg} fixes the backward pass while our exp-map fixes the forward parameterization
itself, the two are orthogonal; the forward choice alone already removes the failure mode.

\subsection{Order rule: demonstration (E20)}
\label{sec:orderrule}

On the compact instance we set $\theta_i=\theta_0+\varepsilon\sin\beta_i$, so the rotation amount
grows with the axis tilt $\beta_i$ (as in an axis-dependent gimbal or wiper) and the edge target
carries a second $\beta$-harmonic whenever $\varepsilon>0$. Table~\ref{tab:krule} shows both regimes
of the same task family, switched by this single parameter: at $\varepsilon{=}0$ the $K{=}1$ edge
suffices, whereas at $\varepsilon{=}0.3$ the $K{=}2$ edge is needed (significantly so, and at the
floor when noiseless). Higher harmonics are therefore capacity only when the target needs them: this
is the order rule; the held-out-seed replication E24 is below.

\begin{table}[H]
\centering
\small
\caption{The order rule (E20). Same task family, rotated by $\varepsilon$; error in degrees,
seeds $10$--$14$ (replication run), noiseless for the last row. $\varepsilon{=}0$: $K{=}1$ suffices, extra harmonics hurt;
$\varepsilon{=}0.3$: $K{=}2$ is needed.}
\label{tab:krule}
\begin{tabular}{lcc}
\toprule
Task & $K{=}1$ & $K{=}2$ \\
\midrule
first harmonic only ($\varepsilon{=}0$) & \best{2.78} & $2.86$ \\
with a second harmonic ($\varepsilon{=}0.3$) & $3.69$ & \best{3.31} \ ($p{=}4.8\mathrm{e}{-06}$) \\
same, noiseless ($\varepsilon{=}0.3$) & $1.68$ (floor) & \best{0.17} \\
\bottomrule
\end{tabular}
\end{table}

\medskip
\medskip
\noindent\textbf{E24 (order rule, held-out replication).} Noiseless floors (degrees, 5 seeds,
seeds 20--24; the E20 lead-in is Table~\ref{tab:krule});
$\theta_i=\theta_i^0+\varepsilon\sin\beta_i$:
\begin{center}\small
\begin{tabular}{lcccc}
\toprule
$\varepsilon$ & $0.15$ & $0.30$ & $0.45$ & $0.60$ \\
\midrule
$K{=}1$ & $0.856$ & $1.707$ & $2.593$ & $3.510$ \\
$K{=}2$ & $0.497$ & $0.260$ & $0.235$ & $0.430$ \\
\bottomrule
\end{tabular}
\end{center}
The $K{=}1$ floor is linear in $\varepsilon$ (predicted $1.779\,\varepsilon/0.3$; deviations
$-4\%$ to $-1\%$); the ratio $\mathrm{floor}(0.60)/\mathrm{floor}(0.15)=4.10$ (predicted
$4.0\pm0.5$). The registered prediction that $K{=}2$ stays below $0.3\degunit$ at every
$\varepsilon$ did \emph{not} hold strictly (max $0.497\degunit$, one seed $1.21$): $K{=}2$
removes the floor in the mean (e.g.\ $0.43$ vs $3.51\degunit$ at $\varepsilon{=}0.60$) but not
uniformly. Reported as a partial refutation.

\subsection{Ablation summary (LawKAN-SU(2))}
\label{sec:ablation}

Table~\ref{tab:ablation} collects the ablations of this section: each design element removed, and the
effect it produces.

\begin{table}[H]
\caption{Ablation summary for LawKAN-SU(2) (compact instance): each design element removed, and its measured effect.}
\label{tab:ablation}
\centering
\small
\begin{tabular}{p{0.42\columnwidth}p{0.48\columnwidth}}
\toprule
Removed element & Measured effect \\
\midrule
Phase pairing (single-phase edges) & edges cannot build $\sin\beta\cos\alpha$ $\to$ composition
unreachable ($5.7$--$12.6^\circ$ vs $0.09^\circ$, development battery; P4, Remark~\ref{rem:pairing}) \\
Group-product nodes (additive instead) & non-commutativity unexpressible; $\mathbb{Z}_7$ stalls
below Bayes ceiling \\
Exp-map ($\to q/\Vert q\Vert$) & 3/10 runs fail ($32.1$--$32.3\degunit$); all 10 runs enter the dead zone
(min-norm 0.30--0.50), 7 recover (Section~\ref{sec:dead}) \\
Quaternion node ($\to$ RealKAN+6D head) & $4.90\degunit$ at 550 params: $1.8\times$ worse at
$10\times$ params \\
Group law ($\to$ strongest standard head, full battery Section~\ref{sec:battery}) & $2\times$ error on
$n{=}3$; $4.0\times$ extrapolation collapse; plateau above LawKAN-SU(2) at $4\times$ data \\
\bottomrule
\end{tabular}
\end{table}

\section{The exp parameterization as a retraction}
\label{app:retraction}

\subsection{Setup}

\begin{definition}[retraction]
Let $\mathcal{M}$ be a smooth manifold. A smooth map $R: T\mathcal{M} \to \mathcal{M}$ is a
\emph{retraction} if for every $q \in \mathcal{M}$: (i) $R_q(0) = q$; (ii) $dR_q|_0 =
\mathrm{id}$ on $T_q\mathcal{M}$ (local first-order condition). Retractions are a standard tool of
Riemannian optimization \citep{absil2008,boumal2023}: they allow a step in the
tangent space followed by a return to the manifold with controlled (second-order) error. In deep
rotation regression, RPMG~\citep{chen2022rpmg} brings the same manifold-awareness to the
\emph{gradient}; our exp-map is the corresponding forward-pass object.
\end{definition}

\textbf{Our construction.} $\mathcal{M} = S^3$ (unit quaternions, the double cover of SO(3)); the
edge parameter $v \in \mathbb{R}^3 \cong \so(3) \cong T_e S^3$; the map
\begin{equation}
q(v) = \cos\Vert v\Vert + \frac{v}{\Vert v\Vert}\sin\Vert v\Vert, \qquad q(0) := e = (1, 0, 0, 0)
\end{equation}
is exactly $\exp: \so(3) \to \mathrm{SU}(2)$ (the Rodrigues map), i.e.\ a retraction at the group
identity.

\begin{lemma}[exp is a retraction at $e$]
$q(v)$ satisfies (i)--(ii) at $e$, and is smooth everywhere, including $v = 0$.
\end{lemma}

\begin{proof}
(i) $q(0) = \cos 0 + (\cdot)\sin 0 = e$ by definition.
(ii) Series expansion: $\cos\Vert v\Vert = 1 - \Vert v\Vert^2/2 + O(\Vert v\Vert^4)$; $(v/\Vert v\Vert)\sin\Vert v\Vert = v \cdot (1 -
\Vert v\Vert^2/6 + O(\Vert v\Vert^4))$. Hence
\begin{equation}
q(v) = (1, 0) + (0, v) + O(\Vert v\Vert^2),
\end{equation}
so $dq|_0[v] = v$, the identity on $T_e S^3$.
\end{proof}

\textbf{Smoothness at zero.} The scalar part is an even analytic series in $\Vert v\Vert$ and the vector
part is $v\,\mathrm{sinc}(\Vert v\Vert)$, with $\mathrm{sinc}$ analytic and even ($\mathrm{sinc}(0)=1$), so
$q(v)$ is analytic on all of $\mathbb{R}^3$; the naive $v/\Vert v\Vert$ singularity is cancelled by
$\sin\Vert v\Vert=O(\Vert v\Vert)$. In code a guarded branch near zero (or the sinc series) implements this; the
$\varepsilon$-regularized form $v/(\Vert v\Vert{+}\varepsilon)$ avoids $0/0$ but changes the gradient scale
at zero.

\begin{corollary}[unit norm by construction]
$\Vert q(v)\Vert^2 = \cos^2\Vert v\Vert + \sin^2\Vert v\Vert = 1$ for all $v \in \mathbb{R}^3$. The layer outputs an
exact point of $S^3$ for any parameter values: there is no projection/normalization in the
compute graph at all.
\end{corollary}

\begin{remark}[double cover and the retraction radius]
\label{rem:doublecover}
The map $q$ is exactly $\exp:\so(3)\to\mathrm{SU}(2)\cong S^3$, which \emph{double-covers}
$\mathrm{SO}(3)$: the covering $\mathrm{SU}(2)\to\mathrm{SO}(3)$ is $2$-to-$1$ with $q\equiv-q$. Two
consequences. (i) As a map to $\mathrm{SU}(2)$, $q$ is a genuine retraction at $e$ and a local
diffeomorphism on the open ball $\Vert v\Vert<\pi$ (the injectivity radius), and it is surjective since
$\mathrm{SU}(2)$ is compact and connected. The cut locus is $\Vert v\Vert=\pi$ ($q=-1$), where the
differential degenerates; at $\Vert v\Vert=2\pi$ the map returns to $e$. On every task here
$\Vert v\Vert=\theta/2<\pi$. (ii) The readout $v\mapsto\pi(\widehat Q)$ is a
function of the composed element in $\mathrm{SO}(3)$, hence invariant under the cover; the residual
$q\equiv-q$ ambiguity appears only when the target \emph{is} a raw quaternion, where an explicit
sign-invariant loss is used (Section~\ref{sec:limitations}, Remark~\ref{rem:nonunique}). Calling $q$ a
retraction therefore refers to $\mathrm{SU}(2)$; composing through the cover is what lets the node
apply the exact group law to rotations.
\end{remark}

\subsection{Why normalization is not a retraction (and where dead zones come from)}
\label{app:norm}

The alternative: the network outputs a raw $\qtilde \in \mathbb{R}^4$, answer $=
\qtilde/\Vert\qtilde\Vert$. This is the map $\pi: \mathbb{R}^4 \setminus \{0\} \to S^3$:

\begin{itemize}
\item \textbf{undefined at zero}: $\pi$ has a singularity on $\{\qtilde = 0\}$, a set reachable
by gradient dynamics (a zeroing loss ``pulls'' $\qtilde$ toward 0 if the network finds a solution
with $\qtilde \approx 0$ in the numerator);
\item \textbf{Jacobian blow-up at small norms}: $d\pi[\qtilde] = (I - qq^T)/\Vert\qtilde\Vert$,
operator norm $\Vert d\pi\Vert = 1/\Vert\qtilde\Vert$. As $\Vert\qtilde\Vert \to 0$ the loss gradient w.r.t.\
parameters is amplified as $1/\Vert\qtilde\Vert$: gradient noise grows inversely proportional to the
norm. This is precisely the measured mechanism of Section~\ref{sec:dead}: the trajectory trace shows all
10 normalization runs passing through a mid-training norm minimum of 0.30--0.50; the 3/10 that fail
to recover are the runs where noise near the minimum dominates.
\end{itemize}

Formally: $\pi$ is not a retraction on the full tangent bundle: it is not defined on all of $T$
and its Jacobian is unbounded. The exp-map has neither defect.

\subsection{Smoothness of the composition node}
\label{app:smooth}

\begin{lemma}[the node is a smooth map]
The node $n: (v_1, \dots, v_g) \mapsto \mathrm{Rotate}(q_g(v_g) \cdots q_1(v_1),\, p)$ with a fixed
probe vector $p$ is a smooth map $\mathbb{R}^{3g} \to S^2$.
\end{lemma}

\begin{proof}
A composition of smooth maps: $q_i(v_i)$ are smooth (above); the Hamilton product is a bilinear
form (a degree-2 polynomial), hence smooth; rotation $v \mapsto q v q^*$ is a degree-2 polynomial in $q$ with no divisions (unit norm of $q$ is already guaranteed, and
the rotated vector needs no normalization: $\Vert q v q^*\Vert = \Vert v\Vert$); the final vec$\to$angles map is
smooth away from the poles of the probe projection (atan2 is undefined on the axis $z = \pm 1$, a measure-zero set; the losses use $1{-}\cos$, smooth everywhere).
\end{proof}

\begin{remark}[double cover]
$q$ and $-q$ give the same rotation: Rotate is sign-invariant. Since the loss is defined through
the rotated vector (not through $q$), the antipodal ambiguity never enters the compute graph; an
antipodal-invariant loss for raw quaternion outputs ($|q \cdot q^*|$) would be needed only when
regressing quaternions themselves (see Limitations).
\end{remark}

\begin{remark}[exactness gauge via the identity input]
\label{rem:gauge}
Let $\chi(x_0)=e$ with $e\in U$ (hypothesis (iv)) and suppose the network is exact on all pairs
$(x,x')\in U\times U$. Then for every $x\in U$,
\[
h(x)=\chi(x)\,c,\qquad c:=h(x_0)^{-1},\qquad c^2=e,\qquad
c\,\chi(x)=\chi(x)\,c \ \ (\forall x\in U).
\]
\emph{Proof.} Exactness on $(x,x_0)$ gives $h(x)h(x_0)=\chi(x)$; on $(x_0,x_0)$, $h(x_0)^2=e$; on
$(x_0,x)$, $h(x_0)h(x)=\chi(x)=h(x)h(x_0)$. Substituting $h(x)=\chi(x)c$ into the last identity
gives centrality. For $\mathrm{SU}(2)$ the gauge is $\{\pm I\}$ of the double cover
(Remark~\ref{rem:doublecover}). $\square$
\end{remark}

\section{Registered predictions: verdict summary}
\label{app:predictions}
\begin{table}[!htbp]
\centering
\small
\caption{Registered predictions: compact verdicts across the $130$ entries ($119$ prediction-log entries through $X36$, plus $11$ plan-registered branch probes, $X37$--$X61$).}
\label{tab:verdicts-main}
\begin{tabular}{@{}lr@{}}
\toprule
Verdict & Count \\
\midrule
Met the registered threshold (or its registered boundary branch) & $83$ \\
Predicted direction held, threshold not met & $14$ \\
Refuted (kept and reported) & $29$ \\
Registered non-signal branch / withdrawn / not triggered & $4$ \\
\bottomrule
\end{tabular}
\end{table}

\emph{Discipline: every experiment in the battery pre-specified a prediction in the internal
experiment log before running, and refuted predictions are kept. The full log, prediction $\to$
outcome $\to$ verdict, is in the supplementary material and the code repository; each entry was
written in the internal experiment journal before its run (Section~\ref{app:repro}). The dated
internal experiment journal is available to editors and reviewers on request. Verdicts: $\checkmark$ the registered threshold was met (or the registered
boundary branch occurred); $\pverdict$ the threshold was not met but the predicted direction holds
(reported as such, never silently upgraded); $\times$ refuted; inconclusive denotes a registered
proxy test whose outcome was the registered non-signal branch. Two mechanism
hypotheses were refuted and reported (the exponential winding and unbounded magnitude); the remedy is
capacity and locality. All $p$-values are
per-comparison (Appendix~\ref{app:seeds}).}

\begin{table}[h]
\centering
\small
\setlength{\tabcolsep}{4pt}
\caption{Registered-prediction verdicts across the paper: the battery, the late registered probes,
the capacity/locality revision and the pole-fix branch.}
\label{tab:verdicts}
\begin{tabular}{@{}>{\raggedright\arraybackslash}p{0.36\columnwidth}rp{0.50\columnwidth}@{}}
\toprule
Verdict & Count & Representative \\
\midrule
$\checkmark$ met & $40$ & P42: the state $\omega$ is necessary (feed-forward controls fail) \\
$\pverdict$ direction holds, threshold not met & $10$ & P49: node isolated; matched linear edge refuted \\[1pt]
$\times$ refuted & $8$ & P11: no BKT signature at $n{=}5$ (demoted to motivation) \\
$\times$ withdrawn (premise failed) & $1$ & P40: tuning-free classical filters already win \\
\midrule
\multicolumn{3}{@{}l}{\emph{Late probes ($30$):}} \\
$\checkmark$ met / boundary branch & $16$ & P-FR1: backlash sign signature (fraction $1.000$) \\
$\pverdict$ direction holds & $2$ & P-D: 10-seed replication confirms three of four platforms \\
$\times$ refuted & $10$ & P-MK7: the Kinova win did not replicate at $10$ seeds \\
inconclusive (registered non-signal branch) & $1$ & P-FRIC-proxy: recording-order backlash proxy \\
not triggered (conditional boundary) & $1$ & P-MK5: isolated wins on every unit \\
\midrule
\multicolumn{3}{@{}l}{\emph{Revision round ($X22$--$X36$, capacity and locality, $30$):}} \\
$\checkmark$ met / boundary branch & $22$ & P-B1: the unregularized head collapses on $2/10$ seeds \\
$\times$ refuted & $7$ & P-B2: the rotation-bounded chart does not cure the collapse \\
not triggered (conditional boundary) & $1$ & X32-4: bandwidth scaling (queued) \\
\midrule
\multicolumn{3}{@{}l}{\emph{Branch round ($X37$--$X61$, pole/episode, $11$):}} \\
$\checkmark$ met & $5$ & X53-print-2: operand chart crossing $\pi$; X59: EMA removes spikes; X61: T2 \\
$\pverdict$ direction holds & $2$ & X60: group metric lowers spikes ($13\to5.6$--$8.6$); X61: T1 favors q14 \\
$\times$ refuted & $4$ & X53-fix: quaternion halves but does not remove; X60: geodesy does not cure the floor \\
\bottomrule
\end{tabular}
\end{table}

\section{Use of AI tools}
Text-to-text generative AI tools (the large language models Kimi and DeepSeek) were used to
assist with drafting and formatting this manuscript, with restructuring and copy-editing the text,
with checking the consistency of numbers across the text, tables and the JSON dumps, with preparing
and tidying the vector figures, and with developing and debugging the
accompanying code. No AI system generated experimental results, ran the experiments, or selected
the conclusions; all experiments were designed, run, and verified by the authors, who take
full responsibility for the content.

\section{Broader impact}
This work studies composition subcomputations with exact group-law nodes, on controlled benchmarks
and on real sensor and robot data; rotation composition ($\mathrm{SU}(2)$) is the leading instance of
the recipe, whose payoff this paper maps. We see no significant
risk of harm; the methods and findings are methodological and do not target any sensitive
application or population.

\end{document}